\pdfoutput=1
\documentclass{article}

\usepackage{microtype}
\usepackage{graphicx}
\usepackage{subfigure}
\usepackage{booktabs}%

\usepackage{enumitem} %
\usepackage{multicol}  %

\usepackage{hyperref}

\usepackage[accepted]{arxiv}

\usepackage{amsmath}
\usepackage{amssymb}
\usepackage{mathtools}
\usepackage{amsthm}
\usepackage{algorithm}
\usepackage{algorithmic}
\usepackage{xspace}
\usepackage{xcolor}
\usepackage{dashrule}
\usepackage{xhfill}
\usepackage{titlesec}

\titlespacing*{\paragraph}
{0pt}%
{0em}%
{1em}%

\usepackage[capitalize,noabbrev]{cleveref}

\theoremstyle{plain}
\newtheorem{theorem}{Theorem}[section]
\newtheorem{proposition}[theorem]{Proposition}
\newtheorem{lemma}[theorem]{Lemma}

\theoremstyle{definition}
\newtheorem{definition}[theorem]{Definition}
\newtheorem{assumption}[theorem]{Assumption}
\theoremstyle{remark}

\Crefname{assumption}{Assumption}{Assumptions} %

\usepackage[textsize=tiny]{todonotes}

\usepackage{pgfplots}
\pgfplotsset{compat=newest}
\usepackage{tikz}
\usetikzlibrary{backgrounds,calc,shadings,shapes.arrows,shapes.symbols,shadows,fit,positioning,topaths,hobby}

\icmltitlerunning{Byzantine-Resilient Zero-Order Optimization for Communication-Efficient Heterogeneous Federated Learning}

\newcommand{\hide}[1]{}
\newcommand{\seed}{\ensuremath{r}}
\newcommand{\lepoch}{\ensuremath{\ell}}
\newcommand{\lepochtmp}{\ensuremath{m}}
\newcommand{\nlepochs}{\ensuremath{K}}
\newcommand{\gepoch}{\ensuremath{t}}
\newcommand{\ngepochs}{\ensuremath{T}}
\newcommand{\client}{\ensuremath{i}}
\newcommand{\clientdata}{\ensuremath{\mathcal{D}_\client}}
\newcommand{\globaldata}{\ensuremath{\mathcal{D}}}
\newcommand{\clienttmp}{\ensuremath{j}}
\newcommand{\nclients}{\ensuremath{n}}
\newcommand{\nseeds}{\ensuremath{\nu}}
\newcommand{\byzantine}{\ensuremath{\mathcal{B}}}
\newcommand{\nbyz}{\ensuremath{b}}
\newcommand{\model}{\ensuremath{\mathbf{w}}}
\newcommand{\modeli}{\ensuremath{\mathbf{w}_\client}}
\newcommand{\modelavg}{\ensuremath{\mathbf{\bar{w}}}}
\newcommand{\modelb}{\ensuremath{\mathbf{\omega}}}
\newcommand{\modelt}[1][\gepoch]{\ensuremath{\model_{#1}}}
\newcommand{\modeltl}[1][\gepoch]{\ensuremath{\model_{#1, \lepoch}^\client}}
\newcommand{\modeltll}[1][\lepoch]{\ensuremath{\model_{\gepoch, #1}^\client}}
\newcommand{\modeltm}[1][\gepoch]{\ensuremath{\model_{#1, \lepochtmp}^\client}}
\newcommand{\modeltmj}[1][\gepoch]{\ensuremath{\model_{#1, \lepochtmp}^\clienttmp}}
\newcommand{\modeltlj}[1][\gepoch]{\ensuremath{\model_{#1, \lepoch}^\clienttmp}}
\newcommand{\modeltlavg}[1][\gepoch]{\ensuremath{\hat{\model}_{#1, \lepoch}}}
\newcommand{\modeltmavg}[1][\lepoch]{\ensuremath{\hat{\model}_{\gepoch, #1}}}
\newcommand{\dimension}{\ensuremath{d}}
\newcommand{\pscale}{\ensuremath{\mu}}
\newcommand{\perturb}{\ensuremath{\mathbf{Z}}}
\newcommand{\perturbvec}{\ensuremath{\mathbf{z}}}
\newcommand{\perturbtr}{\ensuremath{\mathbf{z}^\seed_\gepoch}}
\newcommand{\perturbtl}[1][\lepoch]{\ensuremath{\mathbf{Z}_{\gepoch, #1}}}
\newcommand{\perturbttx}[1][\lepoch]{\ensuremath{\mathbf{Z}_{\gepoch}}}
\newcommand{\perturbtm}[1][\lepochtmp]{\ensuremath{\mathbf{Z}_{\gepoch, #1}}}
\newcommand{\perturbtlr}[1][\seed]{\ensuremath{\perturbvec^{#1}_{\gepoch, \lepoch}}}
\newcommand{\perturbtoner}[1][\seed]{\ensuremath{\perturbvec^{#1}_{\gepoch, 1}}}

\newcommand{\gradi}[1][\client]{\ensuremath{\mathbf{g}_{#1}}}

\newcommand{\graditl}{\ensuremath{\mathbf{g}_\client(\modeltl)}}

\newcommand{\projtmp}{\ensuremath{g(\model, \perturbvec, \pscale, \globaldata)}}
\newcommand{\projtmpargs}[3]{\ensuremath{g(#1, #2, \pscale, #3)}}
\newcommand{\projitr}{\ensuremath{g_\client(\modelt, \perturbtr)}}
\newcommand{\projitoner}[1][\seed]{\ensuremath{g_\client(\modeltl, \perturbtoner[#1])}}
\newcommand{\projitlr}[1][\seed]{\ensuremath{g_\client(\modeltl, \perturbtlr[#1])}}
\newcommand{\projitone}[1][\lepoch]{\ensuremath{\mathbf{g}_\client(\modeltl, \perturbtl[1])}}
\newcommand{\projitl}[1][\lepoch]{\ensuremath{\mathbf{g}_\client(\modeltl, \perturbtl)}}
\newcommand{\projiitl}[1][\client]{\ensuremath{\mathbf{g}_{#1}(\modeltl, \perturbtl)}}
\newcommand{\projitll}[1][\lepoch]{\ensuremath{\mathbf{g}_\client(\modeltll[#1], \perturbtl[#1])}}
\newcommand{\projitm}[1][\seed]{\ensuremath{\mathbf{g}_\client(\modeltm, \perturbtm)}}
\newcommand{\projjtl}[1][\seed]{\ensuremath{\mathbf{g}_j(\modeltlj, \perturbtl)}}
\newcommand{\projjtm}[1][\seed]{\ensuremath{\mathbf{g}_j(\modeltmj, \perturbtm)}}

\newcommand{\projitlravg}[1][\seed]{\ensuremath{\bar{g}(\modeltl, \perturbtlr[#1])}}
\newcommand{\projitlavgh}[1][\seed]{\ensuremath{\bar{\mathbf{g}}_\honest(\modeltl, \perturbtl)}}

\newcommand{\loss}{\ensuremath{F}}
\newcommand{\honest}{\ensuremath{\mathcal{H}}}

\newcommand{\optlossh}{\ensuremath{\loss_\honest^\star}}
\newcommand{\lossh}{\ensuremath{\loss_\honest}}
\newcommand{\lossi}{\ensuremath{\loss_\client}}
\newcommand{\nlossh}{\ensuremath{\nabla \loss_\honest}}
\newcommand{\nlossi}{\ensuremath{\nabla \loss_\client}}

\newcommand{\nlossj}{\ensuremath{\nabla \loss_\clienttmp}}
\newcommand{\nlossiu}{\ensuremath{\nabla \loss_\client^\mu}}
\newcommand{\nlossju}{\ensuremath{\nabla \loss_\clienttmp^\mu}}
\newcommand{\nlossg}{\ensuremath{\nabla \lossh}}
\newcommand{\E}[1]{\ensuremath{\mathbb{E}\left[#1 \right]}}

\newcommand{\Raggtr}[1]{\ensuremath{R\left(\{\projitr\}_{\client=1}^\nclients \right)}}
\newcommand{\raggtr}[1]{\ensuremath{R\left(\{\projitr\}_{\client=1}^\nclients \right)}}
\newcommand{\raggt}{\ensuremath{R_{\gepoch}}}
\newcommand{\raggtl}{\ensuremath{R_{\gepoch, \lepoch}}}

\newcommand{\Raggtl}{\ensuremath{R\left(\{\projitl\}_{\client=1}^\nclients \right)}}

\newcommand{\unitball}{\ensuremath{\mathbb{S}^{\dimension}}}
\newcommand{\uniform}[1][\unitball]{\ensuremath{\mathcal{U}(#1)}}

\newcommand{\gradvar}{\ensuremath{\sigma^2}}
\newcommand{\graddiv}{\ensuremath{\zeta^2}}
\newcommand{\lipschitz}{\ensuremath{L}}
\newcommand{\glipschitz}{\ensuremath{G}}
\newcommand{\numconst}{\ensuremath{\varphi}}
\newcommand{\hlipschitz}{\ensuremath{D}}
\newcommand{\robustness}{\ensuremath{\kappa}}
\newcommand{\robustaggop}{\ensuremath{R}}
\newcommand{\lr}{\ensuremath{\eta}}
\newcommand{\tradetmp}{\ensuremath{\tau}}

\newcommand{\sqnorm}[1]{\ensuremath{\left\Vert#1 \right\Vert^2}}
\newcommand{\sqnormbig}[1]{\ensuremath{\begin{aligned}[t] \bigg\Vert #1 \bigg\Vert^2 \end{aligned}}}
\newcommand{\sqnormbignal}[1]{\ensuremath{\bigg\Vert #1 \bigg\Vert^2}}
\newcommand{\al}[1]{\ensuremath{\begin{aligned}[t] #1 \end{aligned}}}
\newcommand{\norm}[1]{\ensuremath{\left\Vert#1 \right\Vert}}
\newcommand{\card}[1]{\ensuremath{\left\vert#1 \right\vert}}
\newcommand{\innerprod}[2]{\ensuremath{\left\langle #1, #2 \right\rangle}}
\newcommand{\e}[1]{\ensuremath{\mathbb{E}\left[#1\right]}}
\newcommand{\ec}[2]{\ensuremath{\mathbb{E}\left[#2\right]}}
\newcommand{\ecbig}[2]{\ensuremath{\begin{aligned}[t] \mathbb{E}\bigg[ #2 \bigg] \end{aligned}}}

\newcommand{\sumc}{\ensuremath{\sum_{\client=1}^\nclients}}
\newcommand{\sumhc}{\ensuremath{\sum_{\client \in \honest}}}
\newcommand{\sumhcj}{\ensuremath{\sum_{j \in \honest}}}
\newcommand{\sumle}{\ensuremath{\sum_{\lepoch=1}^\nlepochs}}

\newcommand{\hist}{\ensuremath{\phi_\gepoch}}
\newcommand{\histl}[1][\lepoch]{\ensuremath{\phi_\gepoch^{#1}}}
\newcommand{\randl}[1][\lepoch]{\ensuremath{\kappa_\gepoch^{#1}}}

\newcommand{\consta}[1][\lepoch]{\ensuremath{c_{13}}}
\newcommand{\constb}[1][\lepochtmp]{\ensuremath{c_{14}}}
\newcommand{\constc}[1][\lepochtmp]{\ensuremath{c_{15}}}
\newcommand{\constd}[1][\lepochtmp]{\ensuremath{c_{16}}}

\newcommand{\conste}[1][\lepoch]{\ensuremath{c_6}}
\newcommand{\constf}[1][\lepoch]{\ensuremath{c_7}}
\newcommand{\constg}[1][\lepoch]{\ensuremath{c_8}}

\newcommand{\constl}{\ensuremath{c_{1}}}
\newcommand{\constm}{\ensuremath{c_{2}}}

\newcommand{\consth}[1][\lepoch]{\ensuremath{c_9}}
\newcommand{\consti}[1][\lepoch]{\ensuremath{c_{10}}}
\newcommand{\constj}[1][\lepoch]{\ensuremath{c_{11}}}
\newcommand{\constk}[1][\lepoch]{\ensuremath{c_{12}}}

\newcommand{\constaprime}[1][\lepoch]{\ensuremath{c_{13}^\prime(#1)}}
\newcommand{\constbprime}[1][\lepochtmp]{\ensuremath{c_{14}^\prime}}
\newcommand{\constcprime}[1][\lepochtmp]{\ensuremath{c_{15}^\prime}}
\newcommand{\constdprime}[1][\lepochtmp]{\ensuremath{c_{16}^\prime}}

\newcommand{\consteprime}[1][\lepoch]{\ensuremath{c_6^\prime(#1)}}
\newcommand{\constfprime}[1][\lepoch]{\ensuremath{c_7^\prime(#1)}}
\newcommand{\constgprime}[1][\lepoch]{\ensuremath{c_8^\prime}}

\newcommand{\constlprime}{\ensuremath{c_{1}^\prime}}
\newcommand{\constmprime}{\ensuremath{c_{2}^\prime}}

\newcommand{\consthprime}[1][\lepoch]{\ensuremath{c_9^\prime}}
\newcommand{\constiprime}[1][\lepoch]{\ensuremath{c_{10}^\prime}}
\newcommand{\constjprime}[1][\lepoch]{\ensuremath{c_{11}^\prime}}
\newcommand{\constkprime}[1][\lepoch]{\ensuremath{c_{12}^\prime}}

\newcommand{\constn}[1][\lepoch]{\ensuremath{c_{3}}}
\newcommand{\consto}[1][\lepoch]{\ensuremath{c_{4}}}
\newcommand{\constp}[1][\lepoch]{\ensuremath{c_{5}}}

\newcommand{\constnprime}[1][\lepoch]{\ensuremath{c_{3}^\prime}}
\newcommand{\constoprime}[1][\lepoch]{\ensuremath{c_{4}^\prime}}
\newcommand{\constpprime}[1][\lepoch]{\ensuremath{c_{5}^\prime}}

\newcommand{\sumletmp}{\ensuremath{\sum_{\lepochtmp=1}^{\lepoch}}}

\newcommand\numberthis{\addtocounter{equation}{1}\tag{\theequation}}

\newcommand{\define}{\triangleq}

\newcommand{\tmpvec}[1][\client]{\mathbf{v}_{#1}}
\newcommand{\tmpvecentry}[1][j]{{v}_{\client,#1}}
\newcommand{\tmpvecentryi}[1][\client]{{v}_{#1,j}}
\newcommand{\tmpvecsorted}[1][\clienttmp]{\mathbf{v}_{\client, #1}}
\newcommand{\tmpcwtmentries}[1][j]{\ensuremath{\mathcal{X}_{#1}}}
\newcommand{\tmpnnm}[1][\client]{\bar{\mathbf{v}}_{#1}}

\newcommand{\glipub}{\ensuremath{\numconst^2\glipschitz^2\dimension}}

\newcommand{\initseed}{\ensuremath{s}}

\newcommand{\efrac}{\ensuremath{\epsilon^\prime}}

\newcommand{\probguarantee}{\ensuremath{\Delta}}

\newcommand{\ie}{\emph{i.e.,} }

\newcommand{\iid}{i.i.d.\xspace}
\newcommand{\noniid}{non-i.i.d.\xspace}

\begin{document}

\twocolumn[
\icmltitle{Byzantine-Resilient Zero-Order Optimization for \\
           Communication-Efficient Heterogeneous Federated Learning}

\icmlsetsymbol{equal}{*}

\begin{icmlauthorlist}
\icmlauthor{Maximilian Egger}{comp}
\icmlauthor{Mayank Bakshi}{sch}
\icmlauthor{Rawad Bitar}{comp}

\end{icmlauthorlist}

\icmlaffiliation{comp}{School of Computation, Information and Technology, Technical University of Munich}
\icmlaffiliation{sch}{School of Electrical, Computer, and Energy Engineering, Arizona State University}

\icmlcorrespondingauthor{Maximilian Egger}{maximilian.egger@tum.de}

\icmlkeywords{Machine Learning, ICML}

\vskip 0.3in
]

\newcommand{\algname}{\textsc{CyBeR-0}\xspace}
\printAffiliationsAndNotice{This is a follow-up work on the preliminary version \citep{neto2024communication}, where we first introduced a specific version of \algname. This project has received funding from the German Research Foundation (DFG) under Grant Agreement Nos. BI 2492/1-1 and WA 3907/7-1. The work of Mayank Bakshi is supported by the National Science Foundation under Grant No. CCF-2107526.} %

\begin{abstract}
We introduce \algname, a Byzantine-resilient federated zero-order optimization method that is robust under Byzantine attacks and provides significant savings in uplink and downlink communication costs. We introduce transformed robust aggregation to give convergence guarantees for general non-convex objectives under client data heterogeneity. Empirical evaluations for standard learning tasks and fine-tuning large language models show that \algname exhibits stable performance with only a few scalars per-round communication cost and reduced memory requirements.
\end{abstract}

\section{Introduction}

Federated Learning (FL) \citep{mcmahan2017communication} has emerged as an attractive paradigm for training a machine-learning model on the data owned by distributed clients, without exposing the clients' raw data. Despite its appeal, FL can incur prohibitive communication costs, particularly in bandwidth-limited or wireless scenarios. In a typical FL epoch, every participating client transmits its (potentially high-dimensional) local update to a central \emph{federator}, which then consolidates these updates and returns a global model to the clients. A large body of research has accordingly focused on reducing the communication overhead of transmitting and receiving model updates \citep{wen2017terngrad,karimireddy2019error,vogels2019powersgd,m2021efficient,makkuva2024laser,tang2024z,qin2023federated}. 

\paragraph{Byzantine adversaries.} The distributed nature of FL makes it vulnerable to attacks from \emph{Byzantine clients}, \emph{i.e.}, adversarial clients that craft harmful updates to derail the training process. Without suitable countermeasures, even one Byzantine client can prevent the global model from converging~\citep{blanchard2017machine}. A common countermeasure is the use of robust aggregation rules to assimilate different clients' updates by the federator~\citep{shen2016auror, blanchard2017machine,li2020learning,yin2018byzantine,rodriguez2023survey}. As such defenses typically entail slower convergence, the communication cost of FL becomes even more critical in adversarial settings.

\paragraph{Data heterogeneity.} Another key challenge that hinders FL in real-world scenarios is \emph{data heterogeneity}, \emph{i.e.}, the clients' data might stem from different underlying distributions.  Data heterogeneity can pose challenges for model convergence~\citep{zhao2018federated,zhu2021federated}, and indirectly lead to privacy loss by skewing the clients' model updates~\citep{schlegel2023codedpaddedfl,egger2023private,jahani2023swiftagg+,zhu2019applying,truex2019hybrid,bagdasaryan2019differential,wei2020federated,tang2024private}. The effect of data heterogeneity in FL is especially pronounced in the Byzantine setting. As robust aggregation rules aim to reduce the impact of outliers, if a legitimate client’s local update significantly differs due to a unique data distribution, these rules may mistakenly ignore or minimize that contribution, compromising overall performance~\citep{el2021collaborative,karimireddy2022byzantinerobust,charikar2017learning,liu2021approximate}. To counteract the effect of data heterogeneity, a pre-aggregation rule called Nearest Neighbor Mixing (NNM) is introduced in \citet{allouah2023fixing}. By averaging each client's update with a subset of its neighbors, NNM improves the performance of standard robust aggregation rules.

\paragraph{Zero-order optimization.} Zero-order (ZO) optimization methods~\citep{kiefer1952stochastic,spall1992multivariate,duchi2015optimal,ghadimi2013stochastic,liu2020primer} bypass the need for explicit gradient computations through stochastic approximations. A typical ZO optimization step involves sampling a few random \emph{perturbation vectors}, measuring the objective/loss function differences along these vectors, and then updating the model along the direction of the perturbation vector proportionately to the corresponding function differences. Beyond their utility in black-box problems where explicit gradients are unavailable or computationally prohibitive, ZO methods have also been explored for backpropagation-free neural network training in memory-constrained settings~\citep{salimans2017evolution,ilyas2018black,liu2020primer} and in federated learning~\citep{fang2022communication,qiu2023zeroth,chen2023finegrained}. In particular, ZO methods have proven attractive in fine-tuning scenarios, which exhibit low \textit{intrinsic dimensionality} \citep{salimans2017evolution,malladi2023fine}. In a preliminary version of this work~\citep{neto2024communication}, we introduced a specific version of \algname that ensures Byzantine resilience and communication efficiency in FL settings. The algorithm in~\citep{neto2024communication} is theoretically proven to be resilient to Byzantine clients for convex optimization tasks under \iid\ data. %

\paragraph{Our contribution.} We propose \algname, a framework for communication-efficient, Byzantine-resilient FL that leverages ZO optimization with Byzantine-robust aggregation. Our general algorithm is designed to work with arbitrary robust aggregation rules, such as coordinate-wise trimmed mean~\citep{yin2018byzantine} and Krum~\citep{blanchard2017machine}, that can be composed with heterogeneity-aware pre-processing steps such as NNM~\citep{allouah2023fixing}. We prove the convergence of \algname under mild assumptions about general non-convex losses and bounded data heterogeneity. We measure heterogeneity by using bounded gradient divergence among clients and applying a pseudo-Lipschitz condition to the average loss, as proposed by \citet{wang2025new}. We run comprehensive experiments for classification tasks on the MNIST dataset and fine-tuning RoBERTa-large~\citep{liu2019roberta} to three different tasks. Our experiments show that in the presence of state-of-the-art Byzantine attacks, \algname achieves accuracy comparable to gradient-based Byzantine-resilient FL methods while requiring significantly less communication, and low memory and computational cost compared to traditional gradient-based methods. %

We formally describe \algname in Section~\ref{sec:algdescription} and give here the key technical novelties that we incorporate within. We describe these in further detail in  Section~\ref{sec:innovation}. %
\begin{enumerate}[partopsep=0em, topsep=0em, itemsep=0.5em,parsep=-0.5em, label={$\bullet$}, wide, labelindent =-0pt]
    \item \textbf{Transformed robust aggregation via gradient embeddings}:~Unlike full-gradient algorithms, in \algname, the federator only has access to the ZO updates from clients, posing a challenge for the robust aggregation rule. To ensure that the global model updates lie in the space spanned by the perturbation vectors, \algname performs robust aggregation in the reduced-dimensional embedding of the ZO evaluations. Drawing on Johnson–Lindenstrauss embeddings \citep{johnson1984extension}, we show that the geometry of the original gradients is approximately preserved, so the robustness properties of existing schemes carry over. This also substantially lowers federator-side computation improving the scalability.
    \item \textbf{Low-cost uplink and downlink communication via pseudorandom perturbation directions}:~\algname leverages the structure of ZO updates enabling the clients and federator to \emph{communicate only a handful of scalars} along random perturbation directions, rather than the full model parameters. A pseudorandom generation of these directions ensures synchronization across clients and federator via a shared seed, preserving significant communication savings.
    
    \item \textbf{Multiple local ZO epochs}:~To further reduce the communication overhead, each client performs several local updates (epochs) before transmitting to the federator, thereby reducing the number of global epochs. We propose two methods: in one, the clients sample a new perturbation direction for each local epoch, and in the other, the clients use the same direction for all local epochs to further save on the communication cost. We provide theoretical and empirical convergence guarantees for the former and empirical convergence guarantees for the latter. %
\end{enumerate} 
\section{System Model and Preliminaries}
\paragraph{Notation.} The $L_2$ norm of a vector $\mathbf{x}$ and the inner product of two vectors $\mathbf{x}$ and $\mathbf{y}$ are represented by $\norm{\mathbf{x}}$ and $\innerprod{\mathbf{x}}{\mathbf{y}}$, respectively. Let $\unitball$ be the $\dimension$-dimensional unit sphere, \emph{i.e.}, $\unitball \define \{\mathbf{x} \in \mathbb{R}^\dimension: \sqnorm{\mathbf{x}} = 1\}$. $\uniform$ denotes a uniform distribution over $\unitball$. For a natural number $a$, we let $[a] \define \{1, \cdots, a\}$.  For column vectors $\mathbf{v}_i, i \in [a]$, we denote by $(\mathbf{v}_i)_{i=1}^{a}\define (\mathbf{v}_1,\dots,\mathbf{v}_a)$ the horizontal stacking operation.

\paragraph{Learning objective.} We consider an FL system with $\nclients$ clients and a federator. Each client $\client \in [\nclients]$ posses an individual dataset $\clientdata$. The global dataset is denoted by the multiset $\globaldata = \cup_{\client \in [\nclients]} \clientdata$. Let $\loss:\mathbb{R}^\dimension\times \globaldata\to\mathbb{R}^+$ be a given loss function. Let $\loss(\model,\tilde{\globaldata})\triangleq \sum_{D\in\tilde{\globaldata}}\loss(\model,D)/|\tilde{\globaldata}|$ and $\loss_\client(\model)\triangleq \loss(\model,\clientdata)$ denote the average loss of $\model$ on a subset $\tilde{\globaldata}$ of $\globaldata$ and the average loss of  $\model$ on client $\client$'s dataset, respectively. Given a set $\mathcal{A}\subseteq[n]$, we let $\loss_\mathcal{A}(\model)\define \frac{1}{|\mathcal{A}|}\sum_{\client\in \mathcal{A}}\loss_\client(\model)$ and $\loss^\star_\mathcal{A}\define \min_{\model\in\mathbb{R}^\dimension} \loss_\mathcal{A}(\model)$. The clients collectively wish to find a minimizer of $\loss_{[\nclients]}(\model)$.

All clients and the federator start with an initial model $\model^{(1)}\in \mathbb{R}^\dimension$ and construct a sequence of models $\{\model^{(\gepoch)}\}_{\gepoch\in[\ngepochs]}$, where $\ngepochs$ denotes the number of global epochs. In epoch $\gepoch$, client $\client$ observes the current global model $\model^{(\gepoch)}$, samples a \emph{mini-batch} $\tilde{\clientdata}$ of its local dataset and transmits an update to the federator. The federator aggregates all client updates, generates the new global model $\model^{(\gepoch+1)}$ and conveys it to all clients. For a mini-batch $\tilde{\clientdata}$, we let $\gradi \define \nabla  \loss(\model^{(\gepoch)},\tilde{\clientdata})$.

\paragraph{Adversarial model.} We assume $\nbyz < \nclients/2$ clients are Byzantine. We assume the strongest possible adversarial model, where all Byzantine clients can collaborate and have full knowledge of the learning algorithm, all countermeasures, and the results of honest clients. We refer to the set of honest clients as $\honest \subseteq [\nclients]$, where $\card{\honest} = \nclients-\nbyz$. The learning objective becomes finding a minimizer of $\loss_{\honest}(\model)$.%
\paragraph{Zero-order gradient estimate.} 
ZO gradient estimates are obtained by evaluating the loss function at points in the direction specified by the perturbation direction. In the following, we first define the two-point ZO estimate.\footnote{Although the estimator in Definition~\ref{def:zero_order_estimate} for $\mu=0$ is, strictly speaking, not a zero-order estimate, when $\mu$ approaches $0$, $\projtmp$ approaches $\innerprod{\nabla \loss(\model, \globaldata)}{\perturbvec}$. Despite the focus of this work being on ZO optimization, our algorithm continues to work even in the $\mu=0$ case, \emph{i.e.}, when gradients are computed by first applying backpropagation and then projected in the direction specified by $\perturbvec$.}
\begin{definition}[Two-Point Zero-Order Estimate] \label{def:zero_order_estimate} Let $\perturbvec \in\unitball$, $\tilde{\globaldata}\subseteq\globaldata\setminus\emptyset$ and $\model\in\mathbb{R}^\dimension$. The two-point ZO estimate of the gradient $\mathbf{g}\define \nabla \loss(\model, \tilde{\globaldata})$ along the \emph{perturbation direction} $\perturbvec$ is defined as $\perturbvec \projtmpargs{\model}{\perturbvec}{\tilde{\globaldata}}$, where \vspace{-.2cm}
\begin{align*}
\projtmpargs{\model}{\perturbvec}{\tilde{\globaldata}} \triangleq \begin{cases}\dimension \frac{\loss(\model + \pscale \perturbvec, \tilde{\globaldata}) - \loss(\model - \pscale \perturbvec, \tilde{\globaldata})}{2\pscale} &\mu>0\\ \dimension \innerprod{\nabla \loss(\model, \tilde{\globaldata})}{\perturbvec} & \mu =0\end{cases}
\end{align*} \vspace{-.6cm}
\end{definition}

\paragraph{Robust aggregation.} In Byzantine-resilient FL schemes, the federator aggregates the clients' updates using a robust\footnote{When all clients are honest, a common choice of the robust aggregation rule is the average.} aggregation rule $\robustaggop(\cdot)$. To measure the robustness of an aggregation rule, we adopt the following notion introduced by \citet{allouah2023fixing}.
\begin{definition}[$(\nbyz, \robustness)$-Robust Aggregation] \label{def:robustness} Let $\robustness \geq 0$ and $\nbyz < \nclients/2$. For vectors $\mathbf{v}_1, \cdots, \mathbf{v}_\nclients$ and any set $\honest \subset [\nclients]$ of size $\vert \honest \vert = \nclients-\nbyz$, letting $\bar{\mathbf{v}}_\honest = \frac{1}{\vert \honest \vert} \sum_{\client \in \honest} \mathbf{v}_\client$, an aggregation rule $\robustaggop(\{\mathbf{v}_\client\}_{\client = 1}^\nclients)$ is $(\nbyz, \robustness)$-robust if \vspace{-.1cm}
\begin{equation*}
    \norm{\robustaggop(\{\mathbf{v}_\client\}_{\client = 1}^\nclients) - \bar{\mathbf{v}}_\honest}^2 \leq \frac{\kappa}{\vert \honest \vert} \sum_{\client \in \honest} \norm{\mathbf{v}_\client - \bar{\mathbf{v}}_\honest}^2.
\end{equation*} \vspace{-.8cm}%
\end{definition}
The parameter $\robustness$ measures the robustness of the aggregation rule against at most $\nbyz$ Byzantine clients.%

\section{Overview of \algname} 
\begin{algorithm}[!t]
\caption{\algname: Robust Efficient Zero-Order FL}
\label{alg:federated_zero_order}
\begin{algorithmic}[1]
\REQUIRE Shared seed for PRNG, $\pscale \geq 0$, $\lr>0$, $\nseeds>0$, $\robustaggop$.
\STATE Initialize and broadcast global model $\model^{(1)}$.
\FOR{$\gepoch = 1$ to $\ngepochs$} 
    \FOR{each client $\client \in [\nclients]$ \textbf{in parallel}}
        \STATE Initialize local model $\model_{\gepoch, 1}^\client = \model^{(\gepoch)}$.
        \FOR{$\lepoch = 1$ to $\nlepochs$} 
        \STATE Draw $\perturbtlr[1],\! \cdots\!, \perturbtlr[\nseeds] \sim \uniform$, let $\perturbtl \define (\perturbtlr)_{\seed \in [\nseeds]}$
            \STATE Compute $\projitlr \define \projtmpargs{\modeltl}{\perturbtlr}{\clientdata}, \seed \in [\nseeds]$ (cf. \cref{def:zero_order_estimate})
            \STATE Let $\projitl \define \frac{1}{\nseeds} ((\projitlr)_{\seed=1}^\nseeds)^\top$
            \STATE Update $\modeltll[\lepoch+1] = \modeltl - \lr \perturbtl \projitl$
        \ENDFOR
        \STATE Send $\{\projitl\}_{\lepoch=1}^{\nlepochs}$ to federator.
    \ENDFOR
    \STATE Aggregate $\raggtl = \robustaggop(\{\projitl\}_{\client=1}^{\nclients}), \lepoch \in [\nlepochs]$
    \STATE Update $\model^{(\gepoch+1)} = \model^{(\gepoch)} - \lr \sum_{\lepoch=1}^{\nlepochs} \perturbtl \raggtl.$
    \STATE Broadcast $\raggtl$.
    \STATE Clients recover $\model^{(\gepoch+1)}$ using $\raggtl$ and the known $\perturbtl$.
\ENDFOR
\end{algorithmic}
\end{algorithm}

\subsection{Description of \algname} \label{sec:algdescription}\algname is stated in Algorithm~\ref{alg:federated_zero_order}. In the beginning, we assume that the federator and clients have access to a shared seed that lets all parties sample the same random vectors $\perturbvec \sim \uniform$ through a common pseudo-random number generator (PRNG). The federator and all clients set their initial model to $\model^{(1)}$. The algorithm operates over $\ngepochs$ global epochs, each consisting of $\nlepochs$ local epochs. Set a fixed $\pscale\geq 0$ to be a parameter for the ZO estimator and let $\lr$ be the learning rate. At the beginning of global epoch $\gepoch$, all clients and the federator have an identical global model $\model^{(\gepoch)}$, and each client $\client$ initializes a model $\modeltll[1] = \model^{(\gepoch)}$ for its local epochs. Global epoch $\gepoch$ proceeds as follows:
\begin{enumerate}[partopsep=0em, topsep=0em,itemsep=1em,parsep=-0.5em, label={$\bullet$}, wide, labelindent =-0pt]
    \item \textbf{Sampling perturbation directions}: For each local epoch $\lepoch \in [\nlepochs]$, all clients and the federator first generate $\nseeds\geq 1$ pseudorandom perturbations $\perturbtlr[1], \cdots, \perturbtlr[\nseeds]$. We consider two different settings:
    \begin{enumerate}[itemsep=0em,parsep=0em, label={--}, wide, labelindent =6pt]
        \item \textbf{Unbiased ZO estimator}: In this case, $\perturbvec_{\gepoch,\lepoch}^{\seed}\sim \uniform$ for each $\lepoch\in[\nlepochs]$ and $r\in[\nseeds]$, as done in~\cref{alg:federated_zero_order}.
        \item \textbf{Biased ZO estimator}: In a modified version (see \cref{alg:federated_zero_order_biased}), we choose $\perturbvec_{\gepoch,1}^{\seed}\sim \uniform$ for each $r\in[\nseeds]$ and set $\perturbvec_{\gepoch,\lepoch}^{r} = \perturbvec_{\gepoch,1}^{r}$ for $\lepoch>1$.
    \end{enumerate} 
\item \textbf{Local epochs (client-side)}: For each local epoch $\lepoch$, client $\client$ samples a mini-batch $\tilde{\clientdata}$ and computes the two-point zero-order gradient estimate $\perturbtlr[r]g(\modeltl,\perturbtlr[r],\pscale,\tilde{\clientdata})$ for each $r\in[\nseeds]$. Subsequently, it averages the $\nseeds$ different estimates thus obtained to compute a multi-point ZO estimate and updates its local model using this gradient estimate. Mathematically, let $\modeltl$ be client $\client$'s local model and let $\projitlr\triangleq \projtmpargs{\modeltl}{\perturbtlr}{\tilde{\clientdata}}$. Let $\perturbtl \in \mathbb{R}^{\dimension \times \nseeds}$ denote the matrix $\left(\perturbtlr[1], \cdots, \perturbtlr[\nseeds]\right)$ and $ \projitl\triangleq \big((\projitlr)_{\seed=1}^{\nseeds}\big)^\top$.  The clients update their model as $\modeltll[\lepoch+1] = \modeltll[\lepoch] - \lr \perturbtl \projitl$. 
\item \textbf{Client-to-federator model update}: At the end of local epoch $\nlepochs$, each client $\client$ transmits its cumulative model update to the federator by specifying the projection along the perturbation directions. In the unbiased setting, this is performed at a communication cost of $\nlepochs\nseeds$ scalars per client by transmitting  $\{\projitl:\lepoch\in[\nlepochs]\}$, while in the biased setting, this entails a communication cost of $\nseeds$ scalars per client by transmitting  $\sum_{\lepoch=1}^\nlepochs\projitl$.
\item \textbf{Transformed robust aggregation (federator side)}: Upon receiving the model updates from al clients, the federator performs robust aggregation on the model update vectors. Let $\robustaggop:\left(\mathbb{R}^{\nseeds}\right)^\nclients\to \mathbb{R}^{\nseeds}$ be a given robust aggregagation rule. In the unbiased setting, the federator applies this on each local epoch separately, \ie it computes $\robustaggop\big(\{\projitl\}_{\client \in [\nclients]}\big)$ for each $\lepoch$, and updates the global model as 
$
\model^{(\gepoch+1)} = \model^{(\gepoch)} - \lr \sum_{\lepoch=1}^\nlepochs \perturbtl \robustaggop\big(\{\projitl\}_{\client \in [\nclients]}\big).
$
In the biased setting, the robust aggregation is performed on the cumulative updates across all local epochs, \emph{i.e.}, the federator updates the global model as 
$
\model^{(\gepoch+1)} = \model^{(\gepoch)} - \lr \perturbtl[1] \robustaggop\big(\{\sum_{\lepoch=1}^\nlepochs \projitl\}_{\client \in [\nclients]}\big).
$
\item \textbf{Global model broadcast}: Lastly, the federator conveys the updated model $\model^{(\gepoch+1)}$ by sending the projections of  $\model^{(\gepoch+1)}- \model^{(\gepoch)}$ along the perturbation vectors $\{\perturbvec_{\gepoch,\lepoch}^{\seed}:\lepoch\in[\nlepochs],\seed\in[\nseeds]\}$. Note that, all clients know the perturbation vectors and $\model^{(\gepoch)}$. Hence, they can recover $\model^{(\gepoch+1)}$ as the global model updates always lie in the subspace spanned by the current epoch's perturbation vectors.
\end{enumerate}

\hide{On the clients' side, for a global epoch $\gepoch$, and local epoch $\lepoch \in [\nlepochs]$, let $\modeltl$ and $\graditl \define \nlossi(\modeltl)$ be client $\client$'s local model and true gradient, respectively.  Next, for each perturbation $\perturbtlr[r]$, $r\in[\nseeds]$, it computes the two-point zero-order gradient estimate $g(\modeltl,\perturbtlr[r],\pscale,\clientdata)$. Subsequently, it averages the $\nseeds$ different estimates thus obtained to compute the multi-point zero-order estimate. We use the following notation in our analysis. Let $\perturbtl \in \mathbb{R}^{\dimension \times \nseeds}$ denote the perturbations $\perturbtlr[1], \cdots, \perturbtlr[\nseeds]$ stacked into a matrix. Hence, we denote the clients estimate by $\perturbtl \projitl \define \perturbtl \frac{1}{\nseeds} ((\projitlr)_{\seed=1}^\nseeds)^\mathrm{T} = \frac{1}{\nseeds} \sum_{\seed=1}^\nseeds \perturbtlr \projitlr$, where (for the case $\pscale>0$, cf \cref{def:zero_order_estimate}) $\projitlr = \dimension \frac{\lossi(\modeltl + \pscale \perturbtlr) - \lossi(\modeltl - \pscale \perturbtlr)}{2\pscale}$. The update follows \cref{def:gradient_projection} for $\pscale=0$ accordingly. Starting with global model $\modeltll[0] = \modelt$, at local epoch $\lepoch>0$, the clients update their model as $\modeltll[\lepoch+1] = \modeltl - \lr \perturbtl \projitl$. At $\lepoch=\nlepochs$, the projections $\{\projitl\}_{\lepoch \in [\nlepochs]}$ describing all local model updates are sent to the federator.}

\subsection{Choice of robust aggregation rule}
\hide{As for the robust federator $\robustaggop$, we tested Krum \citet{blanchard2017machine} and coordinate-wise trimmed mean (CWTM) \citet{yin2018byzantine}, standalone and composed with NNM, which was recently shown to improve robust aggregation in heterogeneous settings \citep{allouah2023fixing}.}

\algname is compatible with arbitrary robust aggregation rules. In our theoretical analysis, the robustness guarantees of \algname depend on the parameters $\nbyz$ and $\robustness$, which characterize the rule’s robustness (see Definition~\ref{def:robustness}). Below, we outline three widely used aggregation rules from the literature that we employ in our experimental evaluation.
\begin{enumerate}[partopsep=0em, topsep=0em,itemsep=0em,parsep=-0.5em, label={$\bullet$}, wide, labelindent =-0pt]
\item\textbf{Coordinate-wise trimmed mean (CWTM):} CWTM is a simple yet effective aggregation rule that removes the $\lfloor\beta \nclients\rfloor$ smallest and largest values on a per-coordinate basis, then averages the remaining values for a design parameter $\beta$. In our experiments, we set $\beta$ to be $\nbyz/\nclients$. 
\begin{definition}[CWTM~\citep{yin2018byzantine}] \label{def:cwtm}  Let $0 \leq \beta < 1/2$ be a design parameter, let $\mathcal{X} = \{\tmpvec[1], \cdots, \tmpvec[\nclients]\}$ be a multiset of $\nclients$ vectors $\tmpvec = (\tmpvecentry[1], \cdots, \tmpvecentry[\nseeds]) \in \mathbb{R}^{\nseeds}$ and define $\tmpcwtmentries$ as the multiset obtained from $\tmpvecentryi[0], \cdots, \tmpvecentryi[\nclients]$ by removing the smallest and largest $\lfloor \beta \nclients \rfloor$ elements. CWTM outputs a vector where the $j$-th entry is computed as $\frac{1}{\nclients-2\lfloor \beta \nclients \rfloor} \sum_{x \in \tmpcwtmentries} x$. Hence, $\text{CWTM}_\beta(\{\tmpvec\}_{\client \in [\nclients]}) = \frac{1}{\nclients-2\lfloor \beta \nclients \rfloor}(\sum_{x \in \tmpcwtmentries[1]} x, \cdots, \sum_{x \in \tmpcwtmentries[\nseeds]} x)$.
\end{definition}
\item\textbf{Krum:} Krum selects a single vector from $\nclients$ candidate vectors by focusing on local geometric consistency to limit the influence of outliers. In our experiments, we set the Byzantine budget to $\nbyz$ out of $\nclients$ clients. 
\newcommand{\graddistance}{\ensuremath{d_{\client, \clienttmp}}}
\begin{definition}[Krum~\citep{blanchard2017machine}]
Let $\mathcal{X} = \{\tmpvec[1], \cdots, \tmpvec[\nclients]\}$ be a multiset of $\nclients$ vectors, and $\graddistance$ the Euclidean distance between $\tmpvec[\client]$ and $\tmpvec[\clienttmp]$. Let $\mathcal{C}_\client$ be the set of indices of the $\nclients-\nbyz-2$ vectors closest to $\tmpvec[\client]$ in Euclidean distance. Krum outputs the vector $\tmpvec[\client^\star]$ with $\client^\star = \arg\min_{\client \in [\nclients]} \sum_{\clienttmp \in \mathcal{C}_\client} \graddistance$.
\end{definition}
\item \textbf{Nearest neighbor mixing (NNM):} NNM is a pre-processing method that can be composed with standard robust aggregation rules to improve their Byzantine-resilience under heterogeneous data. NNM mixes the clients' gradients with their nearest neighbors as follows.
\begin{definition}[NNM~\citep{allouah2023fixing}] \label{def:nnm}
    Consider $\nclients$ vectors $\tmpvec[1], \cdots, \tmpvec[\nclients]$ and a parameter $\nbyz$. For each $\tmpvec[\client]$, index the vectors into $\tmpvecsorted[1], \cdots, \tmpvecsorted[\nclients]$ such that $\norm{\tmpvecsorted[1] - \tmpvec}_2 \leq \cdots \leq \norm{\tmpvecsorted[\nclients] - \tmpvec}_2$. NNM outputs the vectors $\tmpnnm[1], \cdots, \tmpnnm[\nclients]$, where
        $\tmpnnm = \frac{1}{\nclients-\nbyz} \sum_{\clienttmp = 1}^{\nclients-\nbyz} \tmpvecsorted.$
\end{definition}
\end{enumerate}
\hide{While performing robust aggregation on the projected gradient might appear natural for CWTM, it is less obvious why pre-processing techniques like NNM or other robust aggregation rules like Krum perform well under projection. The core principle is distance preservation under projection from high to low-dimensional spaces, cf. Johnson-Lindenstrauss Lemma \citep{johnson1984extension} and application to random projections from the unit sphere \citet{li2024simple}, which are crucial to the proof of convergence for arbitrary aggregation functions $\robustaggop$.}
\subsection{Innovations in \algname}\label{sec:innovation}
 In the following, we mention a few challenges addressed by \algname and highlight its key technical novelties.
\begin{enumerate}[partopsep=0em, topsep=0em,itemsep=0em,parsep=0em, label={$\bullet$}, wide, labelindent =-0pt]
\item\textbf{Transformed robust aggregation via embeddings:} Robust aggregation of the full gradients is well understood for achieving Byzantine resilience in FL. However, in \algname the federator only has access to the model updates along the random perturbations rather than full gradients. 

A naive approach would be for the federator to first reconstruct approximate gradients from the ZO updates and then apply robust aggregation. 
However, in general, robust aggregation rules are non-linear, and the aggregate vector may lie outside the subspace spanned by the input vectors. For example, consider vectors $\mathbf{v}_1=[2,2,0]^\top, \mathbf{v}_2= [0,-1,-1]^\top$, and $\mathbf{v}_3=[4,0,-4]^\top$ that lie in the vector space $\mathcal{V}$ spanned by $\{[1,1,0]^\top,[0,1,1]^\top\}$. Then, $\text{CWTM}_{1/3}(\mathbf{v}_1,\mathbf{v}_2,\mathbf{v}_3) = [2,0,-1]^\top\notin\mathcal{V}$ (see \cref{def:cwtm}). In our context, this implies that the global model update $\model^{(\gepoch+1)}-\model^{(\gepoch)}$ does not necessarily lie in the subspace spanned by the perturbation vectors $\{\perturbtlr:\seed\in[\nseeds],\lepoch\in[\nlepochs]\}$. Thus, communicating the global model update requires either additional communication cost or projecting the global model update back onto the subspace, incurring additional variance and computation. 

To address this issue, \algname introduces \emph{transformed robust aggregation}, \emph{i.e.}, robust aggregation of the clients' model updates when viewed as vectors embedded in the perturbation space $\mathbb{R}^\nseeds$. By directly aggregating in this lower dimensional perturbation space, \algname preserves communication savings (as the aggregated updates continue to belong to the perturbation space). A key challenge is to argue that performing the aggregation in the perturbation space and projecting the result back to the gradient space preserves the robustness guarantees from Definition~\ref{def:robustness}. To prove this, we rely on Johnson–Lindenstrauss–style embeddings~\citep{johnson1984extension} to maintain the necessary geometric properties of robust aggregation. Thus, the aggregation remains both efficient and Byzantine-resilient, while limiting the attackers’ ability to manipulate the global update outside the chosen subspace.
\item\textbf{Efficient downlink and uplink communication:}  We leverage the structure of ZO updates and transformed robust aggregation to significantly reduce the communication cost on the uplink and the downlink. On the uplink, this is a consequence of having the clients perform local epochs and transmit only the resulting scalar values  for each perturbation direction. On the downlink, as the transformed robust aggregation outcome is guaranteed to lie in the span of the perturbation vectors, it is sufficient to only specify the projections along the perturbation directions. This reduces the communication burden to a handful of scalars, which is a major reduction compared to typical FL settings where the gradient dimension can range from $10^6$ to $10^{12}$. 

\item\textbf{Shared seed mechanism for synchronizing perturbations:} To ensure correct global model updates, the random perturbation directions must be synchronized between the federator and the clients. A naive, yet inefficient strategy would be to transmit the newly generated perturbation vectors in every round, but this negates any communication gains since their dimension matches that of the model. In \algname, we address this challenge by adopting a lightweight shared seed protocol, inspired by \citet{salimans2017evolution}, which enables both the clients and the federator to locally generate identical pseudorandom perturbations. Note that the (one-time) cost of sharing  a common seed determined by the federator with all clients is negligible given that standard PRNGs, \emph{e.g.}, in Tensorflow, have a cycle length in the order of $2^{128}$ \citep{salmon2011parallel}.
\item\textbf{Multiple local epochs per client:} We show that \algname works well with each client performing multiple local epochs. This results in a reduction in the number of global epochs (and hence less frequent communication). Further, \algname also offers a design choice between the unbiased ZO estimator and the biased ZO estimator. The former is more amenable to theoretical analysis as $\perturbtl$'s are independent across the local epochs. However, it has a communication cost of $\nlepochs\nseeds$ per global epoch (both on each uplink as well as the downlink). The biased ZO estimator further reduces the communication cost by a factor of $\nlepochs$ by deliberately using the same perturbation vectors for each local iteration. This, however, introduces additional bias into the training process. We explore this tradeoff under varying numbers of local epochs in \cref{app:local_iterations}.

\item\textbf{Computation and memory efficiency:}
While it is well established that ZO methods reduce the computation cost, especially when $\nseeds$ is small, the reduction in the memory requirements are especially impressive even when $\nseeds$ is large. \citet{malladi2023fine} showed that ZO methods perform inference utilizing upto a factor of $12$ lower  memory compared to backpropagation. \algname inherits this property and can operate on resource constraint edge devices in fine-tuning tasks where classical approaches are intractable. %
The details can be found in \cref{app:experimental_details}.
\end{enumerate}

\begin{table*}[h!]
\centering
\vspace{-.5cm}
\caption{Mean and standard deviation of maximum accuracies across seeds. The baseline FedAvg without robust aggregation nor Byzantine attacks achieves $91.0 \pm 0.2$}
\begin{tabular}{ccccccccc}
\toprule
Algorithm & R. Agg. & NNM &ALIE & ALIE-NNM & FOE & FOE-NNM & SF & LF\\
\midrule
\textsc{CyBeR-0} & CWTM & No  & 87.4 $\pm$ 0.6 & - & \textbf{69.9 $\pm$ 4.8} & - & 71.2 $\pm$ 2.0 & 87.6 $\pm$ 0.6 \\
\textsc{CyBeR-0} & CWTM & Yes  & 90.3 $\pm$ 0.3 & 88.9 $\pm$ 1.3 & 74.0 $\pm$ 7.3 & \textbf{58.9 $\pm$ 4.9} & 59.2 $\pm$ 3.5 & 90.1 $\pm$ 0.6 \\
\textsc{CyBeR-0} & Krum & No  & 65.1 $\pm$ 8.9 & - & \textbf{34.7 $\pm$ 5.4} & - & 44.1 $\pm$ 11.9 & 66.3 $\pm$ 9.9 \\
\textsc{CyBeR-0} & Krum & Yes  & 90.6 $\pm$ 0.4 & 89.9 $\pm$ 0.4 & 70.4 $\pm$ 4.4 & \textbf{56.9 $\pm$ 6.0} & 58.2 $\pm$ 2.7 & 87.1 $\pm$ 3.5 \\
FedAvg & CWTM & No  & 87.6 $\pm$ 0.7 & - & \textbf{41.7 $\pm$ 4.8} & - & 42.5 $\pm$ 7.9 & 80.8 $\pm$ 1.4 \\
FedAvg & CWTM & Yes  & 90.6 $\pm$ 0.3 & 86.9 $\pm$ 1.2 & 76.8 $\pm$ 3.2 & \textbf{58.5 $\pm$ 2.8} & 67.3 $\pm$ 2.6 & 87.9 $\pm$ 3.5 \\
FedAvg & Krum & No  & 75.7 $\pm$ 3.5 & - & \textbf{23.9 $\pm$ 13.5} & - & 50.0 $\pm$ 8.9 & 55.9 $\pm$ 6.9 \\
FedAvg & Krum & Yes  & 88.6 $\pm$ 0.6 & 86.9 $\pm$ 1.1 & 73.0 $\pm$ 2.0 & \textbf{50.0 $\pm$ 4.1} & 57.7 $\pm$ 5.8 & 83.9 $\pm$ 4.6 \\
\bottomrule
\end{tabular}
\label{tab:results}
\vspace{-.3cm}
\end{table*}

\hide{\section{Robust Zero-Order Federated Learning}

We introduce in the following our communication-efficient and robust algorithm based on zero-order optimization, a variant of the general FL algorithm described earlier which we term \algname.} %

\newcommand{\dirichlet}{\alpha}

\section{Experimental Evaluation}

We evaluate \algname under a range of training tasks and Byzantine attacks. First, on logistic regression with MNIST, we compare \algname (with CWTM and Krum as transformed robust aggregations) against standard FL (FedAvg) with (non-transformed) CWTM and Krum as robust aggregations, optionally combined with Nearest-Neighbor Mixing (NNM). Our results show that \algname achieves better worst-case accuracies than FedAvg, while requiring two orders of magnitude less communication (\cref{fig:wca}).

To highlight its real-world applicability, we follow \citet{malladi2023fine} in fine-tuning RoBERTa-large \citep{liu2019roberta} for various natural language processing (NLP) tasks. Here, \algname maintains competitive accuracies under a variety of attacks with communication savings of up to seven orders of magnitude (\cref{fig:nlp_comparison_dirichlet1_eval_acc_trec}).
\subsection{Setup}

\textbf{Data Distribution.}
To model data heterogeneity, we base the data allocation according to the labels of the samples. First, we consider an \iid scenario, where the data $\clientdata$ of each client stems from the same underlying distribution. Therefore, each client gets the same fraction of samples from a certain label. Detecting outliers arising from Byzantine behavior is usually less challenging in \iid regimes as each honest client computes, on expectation, similar gradient.  To model heterogeneous regimes, referred to as the \noniid scenario,  for each label, we draw a distribution over the clients that determines the portion of samples each client gets from a given label. We employ a Dirichlet distribution, where the parameter $\dirichlet$ models the level of heterogeneity. For $\dirichlet \rightarrow \infty$, we recover the \iid scenario from above. The difficulty of heterogeneous regimes increases with decreasing values for $\dirichlet$. Hence, for the \noniid case, we focus on $\dirichlet\in \{0.1,1\}$ as two challenging regimes.%

\newcommand{\fedavg}{FedAvg\xspace}
\newcommand{\zotm}{\algname-CWTM\xspace}
\newcommand{\zokr}{\algname-Krum\xspace}
\newcommand{\sgdcwtm}{CWTM\xspace}
\newcommand{\sgdkr}{CWTM\xspace}
\newcommand{\foennm}{FOE-NNM\xspace}
\newcommand{\aliennm}{ALIE-NNM\xspace}

\textbf{Attack Models.} We evaluate the robustness and efficiency of our method under a variety of established attacks, i.e., \textit{A Little is Enough} (ALIE) \citep{baruch2019little}, \textit{Fall of Empires} (FOE) \citep{xie2020fall}, \textit{Sign Flipping} (SF) \citep{allen2021byzantine} and \textit{Label Flipping} (LF) \citep{allen2021byzantine}. We provide the details of the attacks in \cref{app:experimental_details}. For \algname, all attacks operate on the projected gradients $\projitl$; while for the non-projected counterparts, the attacks operate on the gradients. This represents the strongest possible adversarial behavior. ALIE and FOE involve an optimization tailored to the used robust aggregation rule. On a high level, the manipulations are chosen to maximize the distance to the desired honest outcome of the aggregated gradient. When investigating robust aggregation rules $\robustaggop$ composed with NNM, we present two variants of ALIE and FOE, optimized to $\robustaggop$, and optimized to $\text{NNM} \circ \robustaggop$. We term the latter \aliennm and \foennm, and find they can reduce the benefits of NNM in certain cases. 
For fine-tuning tasks, we settle on CWTM as transformed robust aggregation, and additionally challenge our algorithm with a specifically tailored full-knowledge attack, inspired by \citet{fang2020local} that targets maximizing the gradient deviation compared to the honest clients. Please see \cref{sec:byzantine_attacks} for details.
\subsection{Comprehensive Study for Logistic Regression} \label{sec:experiments_mnist}
We evaluate $\algname$ compared to various baselines (\emph{cf.}~\citet{allouah2023fixing}) on ALIE, FOE, SF, and LF. We use $\nclients=40$ clients, from which $\nbyz=10$ are Byzantine, and refer to \cref{app:hyperparameters} for details of the experimental setup. CWTM is parameterized with $\beta=\frac{\nbyz}{\nclients}=0.25$. The parameter $\nseeds$ is set to $\nseeds=64$. All results show averaged results over $5$ runs with different seeds, including their standard deviations (shown after $\pm$ in the tables, and with shaded areas in the plots). 
In \cref{tab:results}, we provide results for all attacks on \algname combined with transformed CWTM and Krum, and \fedavg with CWTM and Krum. All variants are tested with and without NNM composition. If composed with NNM, we provide additionally the improved attacks described above, \ie \aliennm and \foennm.
The worst-case performance of each robust aggregation across all attacks is highlighted in bold. With the exception of \aliennm and \foennm, the robustness against all attacks improves uniformly with the introduction of NNM. 

\algname exhibits better worst-case robustness for all aggregation rules compared to standard Byzantine resilient FL and \zotm outperforms the best of all state-of-the-art methods by more than $10$ percent. We note that the goal of this section is not to optimize the absolute achieved accuracies of the presented schemes, but rather to provide a comprehensive overview and comparison for many attacks and robust aggregation rules.

\begin{figure*}[!t]
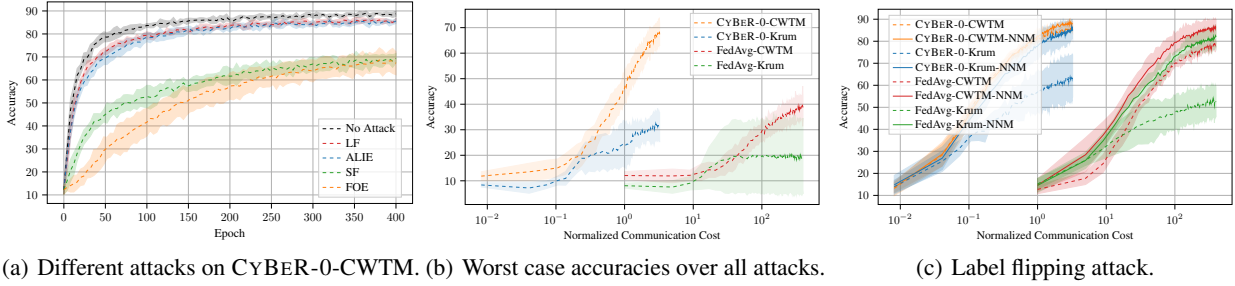

\subfigure[Different attacks on \zotm.]{\resizebox{.32\linewidth}{!}{\input{data/OURS_all_attacks}}\label{fig:ours_all_attacks}}
\subfigure[Worst case accuracies over all attacks.]{\resizebox{.31\linewidth}{!}{\input{data/wca_comm}} \label{fig:wca}}
\subfigure[Label flipping attack.]{\resizebox{.31\linewidth}{!}{\input{data/LABEL_FLIPPING_comm}}\label{fig:label_flipping_comm}}
\vspace{-.3cm}
\caption{Performance of different robust aggregation rules against different attacks for logistic regression on MNIST.} \label{fig:mnist_graphs} \vspace{-.2cm}
\end{figure*}

We provide in \cref{fig:mnist_graphs} a selection of plots showing the average accuracies as a function of global epochs and the communication cost. In \cref{fig:ours_all_attacks}, we show for \zotm the performance across all applicable attacks. The worst-case behavior is dominated by FOE. \cref{fig:wca} shows the significant performance improvements brought by \algname compared to its non-zero-order counterparts, and the reduction in communication cost by multiple orders of magnitude. While for Krum, the performance is improved by more than $10\%$, the performance of CWTM is improved by more than $25\%$ for the worst-case attack. In \cref{fig:label_flipping_comm}, we show the performance improvements brought by both \algname and NNM against LF attack. The graphs of the remaining attacks are provided in \cref{app:extended_experiments_mnist}.

\algname allows for different strategies to conduct local epochs at the clients. Instead of the method presented in \cref{alg:federated_zero_order}, we present and compare in \cref{app:local_iterations} three different methods, exposing an efficiency-bias-variance trade-off. We further study the effect of the parameter $\nseeds$ in \cref{app:primer_zero_order} in standard, non-fine-tuning learning tasks.

\subsection{Fine-Tuning Large Language Models} \label{sec:experiments_fine_tuning}

\begin{figure}[!b]
    \centering
    \resizebox{.8\linewidth}{!}{\input{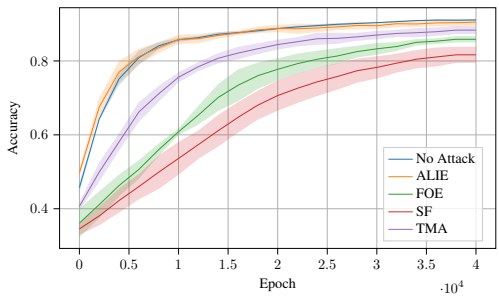}}
    \vspace{-.4cm}
    \caption{Accuracy over epochs for fine-tuning RoBERTa-large on TREC under different attack scenarios for \noniid data.}
    \label{fig:nlp_comparison_dirichlet1_eval_acc_trec}
\end{figure}

We follow the lines of \citet{malladi2023fine} and investigate different fine-tuning tasks in the realm of NLP using the well-established RoBERTa-large \citep{liu2019roberta}. We conduct sentiment analysis on SST-2 \citep{socher2013recursive}, natural language inference (NLI) on the SNLI dataset \citep{bowman2015large}, and topic classification on TREC \citep{voorhees2000building}. We fix CWTM as a robust aggregation technique, as it was earlier shown to yield the best performance in \algname. We average all results over three runs with different seeds. Due to space constraints, some of the results are shown in the Appendix.

\cref{tab:nlp_comparison_iid} shows the results of \zotm under ALIE, FOE, SF, and TMA for a \noniid data distribution with $\dirichlet=1$, and compare the results to the baseline accuracy without attack. We set $\nclients=12$, $\nbyz=3$ and $\nseeds=1$. The rightmost column shows the worst-case accuracies over all attacks. It can be found that the worst-case attacks degrade the accuracy of fine-tuning on SST-2 by only approximately one percent, and that of SNLI and TREC by $7-8\%$. We plot in \cref{fig:nlp_comparison_dirichlet1_eval_acc_trec} the evaluation accuracies throughout the fine-tuning process for all attacks, compared to the baseline. It can be clearly seen that FOE is the most challenging attack in this regime. For more plots, we refer the interested reader to \cref{app:extended_experiments_nlp}. Further, we perform the same study for an \iid data distribution ($\dirichlet=\infty$), and find that \algname exhibits extremely stable performance under both \iid and \noniid data. We refer the reader to \cref{app:extended_experiments_nlp} for the results. %

We give an extensive hyperparameter study for \algname in fine-tuning tasks. We run our experiments on SST-2 and RoBERTa-large, attacked by FOE. We use \noniid data with $\dirichlet=0.1$. Please see \cref{sec:hyperparameter_study} for detailed results. We show the robustness of \algname under varying numbers of total and Byzantine clients with $\frac{\nbyz}{\nclients} = 0.25$ and observe that the robustness increases with $\nclients$. We study the stability of \algname under varying numbers $\nseeds$ of random perturbations, $\nclients = 8$ and $\nbyz = 2$. For comparability, we fix the ratio $\nseeds \ngepochs = 20000$ and observe very similar accuracies for $\nseeds \in \{1, 2, 4, 8\}$. Hence, the number of projections trades almost inversely with the number of global epochs $\ngepochs$. A similar behavior is observed for a varying number of local epochs $\nlepochs$, fixing the ratio $\nlepochs \ngepochs$. However, for large $\nlepochs$ and small $\ngepochs$, we can see the negative impacts of local iterations. Lastly, we evaluate \algname under varying ratios of Byzantine clients, thereby fixing $\nclients=16$ for better flexibility, and using $\nseeds=1$. While the performance reduction from $\nbyz=2$ to $\nbyz=4$ is negligible, we can observe a notable difference for $\nbyz=6$, \ie when the number of Byzantine clients is close to $\nclients/2$.

\section{Convergence Analysis}
We now turn our attention to theoretical guarantees on the convergence of \algname. We provide a thorough theoretical analysis for arbitrary $(\nbyz, \robustness)$-robust aggregation rules, suitably transformed to account for ZO estimates,  under heterogeneous data distribution and general non-convex loss functions. We employ the heterogeneity model introduced by \citet{wang2025new} in our analysis. 

\begin{table*}[t]
\centering
\vspace{-.5cm}
\caption{Mean and Standard Deviation of Maximum Accuracies Across Seeds}
\begin{tabular}{ccccccc}
\toprule
Dataset & ALIE & FOE & SF & TMA & No Attack& Worst Case  \\
\midrule
SST-2 & 93.3 $\pm$ 0.4 & 91.7 $\pm$ 1.5 & 91.6 $\pm$ 0.5 & 92.1 $\pm$ 1.2 & 92.9 $\pm$ 0.1 & 91.6 \\
SNLI & 80.8 $\pm$ 1.4 & 78.4 $\pm$ 2.1 & 76.1 $\pm$ 1.0 & 80.0 $\pm$ 0.7 & 84.3 $\pm$ 0.2 & 76.1 \\
TREC & 93.3 $\pm$ 1.2 & 91.7 $\pm$ 1.0 & 87.7 $\pm$ 2.9 & 92.2 $\pm$ 1.1 & 94.7 $\pm$ 0.7 & 87.7 \\
\bottomrule
\end{tabular}
\label{tab:nlp_comparison_dirichlet1}
\vspace{-.3cm}
\end{table*}

\paragraph{Challenges and novelty.} While \citet{allouah2023fixing} provides a roadmap for analyzing robustness in non-convex regimes, the combination with ZO methods incurring bias and high variance poses particular challenges, especially given that, in expectation, the robust aggregation output does not necessarily equal the expectation of the honest clients' gradient estimate. In addition to the challenges from the gradient estimates, the robust aggregation is here conducted on a transformed space instead of on estimated gradients themselves. We show, through a novel application of projection theorems to ZO methods, that such methods can converge even under adversarial attacks in harsh heterogeneous and non-convex settings. Although \citet{fang2022communication} study the convergence of ZO FL methods in heterogeneous settings, the introduction of Byzantine resilience poses particular challenges, amplified by the new notion of heterogeneity \citep{wang2025new}. Even without robustness, the analysis of \citet{wang2025new} does not apply to ZO methods due to the bias and variance of the gradient estimates, which exhibit additional restrictions on the learning rate and incur a more complex analysis. 

Tackling the above challenges, we believe that the combination of ZO optimization, Byzantine resilience, heterogeneity models, non-convexity, and transformed robust aggregation in gradient embedding spaces exhibits unique challenges interesting from a purely theoretical perspective. 
We make the following assumptions.
\begin{assumption}[Lipschitz gradient] \label{ass:lipschitz} $\forall \model, \modelb \in \mathbb{R}^\dimension, \client \in [\nclients]$,
\begin{equation}
    \norm{\nlossi(\model) - \nlossi(\modelb)} \leq \lipschitz \norm{\model-\modelb}.
\end{equation}
\end{assumption}

\begin{assumption}[Bounded gradient variance] \label{ass:bounded_grad_var} The variance of the clients' gradient estimate is uniformly bounded as 
\begin{equation}
    \E{\sqnorm{\gradi(\model) - \nabla\lossi(\model)}} \leq \gradvar, \forall \model, \client \in [\nclients].
\end{equation}
\end{assumption}

Following \citet{wang2025new}, we assume the following two properties to model the heterogeneity of the system.
\begin{assumption}[Bounded gradient divergence] \label{ass:bounded_grad_div} It holds 
\begin{equation}
    \sqnorm{\nlossi(\model) - \nlossg(\model)} \leq \graddiv, \forall \client \in [\nclients].
\end{equation}\vspace{-.2cm}
\end{assumption}

\begin{assumption}[Pseudo-Lipschitz on averaged gradients] \label{ass:lipschitz_avg_grad} 
\begin{equation}
    \sqnormbig{\frac{1}{\nclients} \sumc \nlossi(\modeli) - \nlossg(\modelavg)}\!\!\! \leq \!\frac{\hlipschitz^2}{\nclients} \sumc\! \sqnorm{\modeli-\modelavg}\!\!\!.\!
\end{equation}\vspace{-.2cm}
\end{assumption}

For a given model $\model$, let $\nlossiu \define \mathbb{E}_{\perturbvec \sim \uniform}[\nlossi(\model + \pscale \perturbvec)]$ be a smoothened version of the gradient $\nlossi$.
Then, for the two-point zero-order estimate from \cref{def:zero_order_estimate}, we have the following well-known result \citep{flaxman2004online}.
\begin{proposition} \label{prop:estexp}
The ZO estimate satisfies on expectation
    \begin{align*}
        \e{\perturbtl \projitl} = \nlossiu(\modeltl).
    \end{align*}
\end{proposition}

\newcommand{\approxerr}{\lipschitz \mu}
\begin{proposition}[Lemma 2, \citep{tang2020distributed}] 
\label{prop:bias}
\vspace{-.1cm}
It holds
    \begin{align*}
    \norm{\nlossiu(\modeltl) - \nlossi(\modeltl)} \leq \approxerr.
\end{align*}
\vspace{-.8cm}
\end{proposition}

All proofs are deferred to \cref{app:proofs}. 
Under the assumptions above, we present the following convergence guarantee for non-convex loss landscapes. %

Let $\epsilon = \sqrt{\frac{64}{\nseeds} \log(\frac{2(\vert \honest \vert-1)}{ \delta})}$, and $\efrac \define \frac{(1+\epsilon)}{(1-\epsilon)}$. 
Let $\constl, \constm, \constn, \consto$ and $\constp$ be as defined in \eqref{eq:constl}-\eqref{eq:constp} 
in \cref{app:proofs}. These constants scale as follows: $\constl = \Theta\big(\lr^2 \nlepochs^3 \frac{\dimension}{\nseeds} + \frac{\dimension}{\nseeds} \lipschitz^2 \lr^2 \nlepochs (\nlepochs + \frac{\dimension}{\vert \honest \vert \nseeds} )\big)$, $\constm = \Theta\big(\lr^2 \nlepochs^3 (\lipschitz+(\frac{\dimension}{\nseeds} + \frac{\dimension^2}{\nseeds^2 \vert \honest \vert} \lipschitz^2 \nlepochs^2 \lr^2) \left(\graddiv +  \gradvar +  \lipschitz^2 \mu \dimension\right) \big)$, $\constn = \Theta \big(\frac{\dimension}{\nseeds} (\nlepochs^2 \efrac \robustness + \frac{\nlepochs}{\vert\honest\vert}) (1 + \lipschitz^2 \lr^2 \nlepochs \left(\nlepochs + \frac{\dimension}{\vert \honest \vert \nseeds} \right) (1+\frac{\nseeds}{\dimension}) )\big)$, $\consto = \Theta \big(\nlepochs^2 \lipschitz^2 \lr^2 (\nlepochs^2 \hlipschitz^2 + \frac{1}{\vert \honest \vert} \frac{\dimension^2}{\nseeds^2} \lipschitz^2) + \nlepochs \efrac \robustness (\hlipschitz^2 + \graddiv + \frac{\dimension}{\nseeds} \lipschitz^2)\big)$ and $\constp = \Theta\Big((\graddiv + \gradvar +  \lipschitz^2 \mu^2 \dimension) \big((\nlepochs^2 \efrac \robustness) (\frac{\dimension}{\nseeds} + \frac{\dimension^2}{\nseeds^2 \vert \honest \vert} \lipschitz^2 \nlepochs \lr^2 ) \big) + \nlepochs^2\lipschitz( \efrac \robustness  + \pscale) + (\nlepochs^2 \efrac \robustness) (\frac{\dimension}{\nseeds} \lipschitz^2 \lr^2 \nlepochs^2 \approxerr)\Big).$

\begin{theorem}[General non-convex landscapes] \label{thm:convergence_non_convex}
Let $0<\probguarantee<1$ and suppose that \cref{ass:bounded_grad_var,ass:bounded_grad_div,ass:lipschitz,ass:lipschitz_avg_grad} hold. Consider \algname with a $(\vert \honest \vert, \robustness)$-robust aggregation rule. If \emph{(a)}
$\lr^2 \leq \min\{\frac{1}{72 \nlepochs^2 \lipschitz}, \frac{\vert \honest \vert \nseeds}{96\nlepochs \dimension \lipschitz^2}, 6 \frac{\hlipschitz^2}{\nlepochs^2} + 6\frac{\graddiv}{\nlepochs^2} + \frac{32\dimension}{\nseeds \nlepochs^2} \lipschitz^2\}$, 
 \emph{(b)} $\nseeds \geq 64 \log(\frac{2(\vert \honest \vert-1) \ngepochs \nlepochs}{\probguarantee})$, and \emph{(c)} $\constn + \consto \constl \leq \frac{1}{2}$ are satisfied, then the following convergence guarantee holds with probability $1-\probguarantee$:
\begin{align*}
    \frac{1}{\ngepochs} \sum_{\gepoch=1}^{\ngepochs} \e{\sqnorm{\nlossh(\modelt)}} \leq &\frac{4(\lossh(\modelt[1]) - \optlossh)}{\lr \nlepochs \ngepochs} \\
    &+ \lr/(2\nlepochs) \left(\consto \constm + \constp \right).
\end{align*}
\end{theorem}
Note that \cref{thm:convergence_non_convex} requires $\nseeds = \Theta(\dimension)$ in order for condition \emph{(c)} to be satisfied. %
However, our empirical results demonstrate that \algname converges even when $\nseeds=1$. To bridge the gap between our empirical finding and the analysis, we make the following additional assumption.\footnote{In applications such as fine-tuning, Assumption~\ref{ass:lipschitz_objective} may be implied by Assumption~\ref{ass:lipschitz}. If it is known that the initial model $\modelt[1]$ is within a distance $\Gamma$ to a local optimal, then  Assumption~\ref{ass:lipschitz_objective} is satisfied with $G=L\Gamma^2/2$.} %
\begin{assumption}[Lipschitz Objective function]\label{ass:lipschitz_objective} $\forall \client \in [\nclients]$,
\begin{equation}
    \norm{\lossi(\model) - \lossi(\modelb)} \leq \glipschitz \norm{\model-\modelb}, \forall \model, \modelb \in \mathbb{R}^\dimension.
\end{equation}
\end{assumption}

\begin{theorem}[Lipschitz objective functions] \label{thm:convergence_non_convex_lipschitz}
Let $0<\probguarantee<1$, %
and suppose that \cref{ass:bounded_grad_var,ass:bounded_grad_div,ass:lipschitz,ass:lipschitz_avg_grad} hold. Consider \algname with $\pscale>0$ and a $(\vert \honest \vert, \robustness)$-robust aggregation rule. If \emph{(a)}
$\lr \leq \min\big\{\frac{1}{26 \nlepochs \lipschitz}, \frac{1}{4 \nlepochs \sqrt{\hlipschitz^2 + \graddiv}}\big\}$, and \emph{(b)} $\nseeds \geq 64 \log(\frac{2(\vert \honest \vert-1) \ngepochs \nlepochs}{\probguarantee})$, the following convergence guarantee holds with probability at least $1-\probguarantee$:
\begin{align*}
    &\frac{1}{\ngepochs} \sum_{\gepoch=1}^{\ngepochs} \e{\sqnorm{\nlossh(\modelt)}} \leq \frac{4(\lossh(\modelt[1]) - \optlossh)}{\lr \ngepochs \nlepochs} \\
    &+ 2 \lr \nlepochs \hlipschitz^2 \left(\frac{\glipub}{\lipschitz^2 \nseeds} + \frac{3}{\lipschitz}\right)\! \left(13 \nlepochs \! + \! 24 \efrac \robustness (1 \! + \! \frac{\graddiv}{\hlipschitz^2}) \! + \! 4\right) \\
    &+ 2 \lr \frac{\glipub}{\nseeds} \! \left(\frac{1}{5\vert \honest \vert}  \! + \! \frac{8}{\vert \honest \vert^2} \! + \! 2 \nlepochs \efrac \robustness \right) \! + \! 12\nlepochs \lipschitz \lr \left( \robustness \efrac \!+ \! \pscale\right)\!.
\end{align*}
\end{theorem}
\section{Conclusion}
In this work, we introduced \algname, a Byzantine-resilient, communication-efficient FL framework that combines zero-order (ZO) optimization with arbitrary robust aggregation rules. \algname uses a shared-seed mechanism to generate identical pseudorandom perturbations at the clients and the federator. This enables ZO updates to be communicated using only a few scalars, thus drastically reduces the communication overhead. 

A key novelty of our approach is the introduction of \emph{transformed robust aggregation}, which operates directly in the perturbation space rather than on reconstructed gradients. This  ensures  compatibility of widely-used robust aggregation rules with ZO updates while retaining Byzantine resilience. Using Johnson-Lindenstrauss embeddings, we establish provable convergence guarantees under non-convex losses, even in adversarial and heterogeneous settings.

Empirical evaluations on logistic regression (MNIST) and fine-tuning RoBERTa-large (NLP) demonstrate that the worst-case robustness of \algname is close to full-gradient Byzantine robust algorithms even though the communication costs are lower by up to seven orders of magnitude. Additionally, its low memory and computational footprint make it well-suited for resource-constrained edge devices.

\section*{Impact Statement}

This paper presents work whose goal is to advance the field of 
Machine Learning. There are many potential societal consequences 
of our work, none which we feel must be specifically highlighted here.

\bibliography{refs}

\begin{thebibliography}{56}
\providecommand{\natexlab}[1]{#1}
\providecommand{\url}[1]{\texttt{#1}}
\expandafter\ifx\csname urlstyle\endcsname\relax
  \providecommand{\doi}[1]{doi: #1}\else
  \providecommand{\doi}{doi: \begingroup \urlstyle{rm}\Url}\fi

\bibitem[Allen-Zhu et~al.(2021)Allen-Zhu, Ebrahimianghazani, Li, and Alistarh]{allen2021byzantine}
Allen-Zhu, Z., Ebrahimianghazani, F., Li, J., and Alistarh, D.
\newblock Byzantine-resilient non-convex stochastic gradient descent.
\newblock In \emph{International Conference on Learning Representations}, 2021.

\bibitem[Allouah et~al.(2023)Allouah, Farhadkhani, Guerraoui, Gupta, Pinot, and Stephan]{allouah2023fixing}
Allouah, Y., Farhadkhani, S., Guerraoui, R., Gupta, N., Pinot, R., and Stephan, J.
\newblock Fixing by mixing: A recipe for optimal byzantine ml under heterogeneity.
\newblock In \emph{International Conference on Artificial Intelligence and Statistics}, pp.\  1232--1300, 2023.

\bibitem[Bagdasaryan et~al.(2019)Bagdasaryan, Poursaeed, and Shmatikov]{bagdasaryan2019differential}
Bagdasaryan, E., Poursaeed, O., and Shmatikov, V.
\newblock Differential privacy has disparate impact on model accuracy.
\newblock \emph{Advances in neural information processing systems}, 32, 2019.

\bibitem[Baruch et~al.(2019)Baruch, Baruch, and Goldberg]{baruch2019little}
Baruch, G., Baruch, M., and Goldberg, Y.
\newblock A little is enough: Circumventing defenses for distributed learning.
\newblock \emph{Advances in Neural Information Processing Systems}, 32, 2019.

\bibitem[Blanchard et~al.(2017)Blanchard, El~Mhamdi, Guerraoui, and Stainer]{blanchard2017machine}
Blanchard, P., El~Mhamdi, E.~M., Guerraoui, R., and Stainer, J.
\newblock Machine learning with adversaries: Byzantine tolerant gradient descent.
\newblock \emph{Advances in neural information processing systems}, 30, 2017.

\bibitem[Bowman et~al.(2015)Bowman, Angeli, Potts, and Manning]{bowman2015large}
Bowman, S.~R., Angeli, G., Potts, C., and Manning, C.~D.
\newblock A large annotated corpus for learning natural language inference.
\newblock In \emph{Empirical Methods in Natural Language Processing}, pp.\  632--642, 2015.

\bibitem[Charikar et~al.(2017)Charikar, Steinhardt, and Valiant]{charikar2017learning}
Charikar, M., Steinhardt, J., and Valiant, G.
\newblock Learning from untrusted data.
\newblock In \emph{ACM SIGACT Symposium on Theory of Computing}, pp.\  47--60, 2017.

\bibitem[Chen et~al.(2023)Chen, Chen, Gu, and Deng]{chen2023finegrained}
Chen, J., Chen, H., Gu, B., and Deng, H.
\newblock Fine-grained theoretical analysis of federated zeroth-order optimization.
\newblock In \emph{Neural Information Processing Systems}, 2023.

\bibitem[Duchi et~al.(2015)Duchi, Jordan, Wainwright, and Wibisono]{duchi2015optimal}
Duchi, J.~C., Jordan, M.~I., Wainwright, M.~J., and Wibisono, A.
\newblock Optimal rates for zero-order convex optimization: The power of two function evaluations.
\newblock \emph{IEEE Transactions on Information Theory}, 61\penalty0 (5):\penalty0 2788--2806, 2015.

\bibitem[Egger et~al.(2023)Egger, Hofmeister, Wachter-Zeh, and Bitar]{egger2023private}
Egger, M., Hofmeister, C., Wachter-Zeh, A., and Bitar, R.
\newblock Private aggregation in wireless federated learning with heterogeneous clusters.
\newblock In \emph{IEEE International Symposium on Information Theory (ISIT)}, pp.\  54--59, 2023.

\bibitem[El-Mhamdi et~al.(2021)El-Mhamdi, Farhadkhani, Guerraoui, Guirguis, Hoang, and Rouault]{el2021collaborative}
El-Mhamdi, E.~M., Farhadkhani, S., Guerraoui, R., Guirguis, A., Hoang, L.-N., and Rouault, S.
\newblock Collaborative learning in the jungle (decentralized, byzantine, heterogeneous, asynchronous and nonconvex learning).
\newblock \emph{Advances in neural information processing systems}, 34:\penalty0 25044--25057, 2021.

\bibitem[Fang et~al.(2020)Fang, Cao, Jia, and Gong]{fang2020local}
Fang, M., Cao, X., Jia, J., and Gong, N.
\newblock Local model poisoning attacks to byzantine-robust federated learning.
\newblock In \emph{USENIX security symposium}, pp.\  1605--1622, 2020.

\bibitem[Fang et~al.(2022)Fang, Yu, Jiang, Shi, Jones, and Zhou]{fang2022communication}
Fang, W., Yu, Z., Jiang, Y., Shi, Y., Jones, C.~N., and Zhou, Y.
\newblock Communication-efficient stochastic zeroth-order optimization for federated learning.
\newblock \emph{IEEE Transactions on Signal Processing}, 70:\penalty0 5058--5073, 2022.

\bibitem[Flaxman et~al.(2005)Flaxman, Kalai, and McMahan]{flaxman2004online}
Flaxman, A.~D., Kalai, A.~T., and McMahan, H.~B.
\newblock Online convex optimization in the bandit setting: gradient descent without a gradient.
\newblock In \emph{ACM-SIAM Symposium on Discrete Algorithms}, pp.\  385–394, 2005.

\bibitem[Gao et~al.(2018)Gao, Jiang, and Zhang]{gao2018information}
Gao, X., Jiang, B., and Zhang, S.
\newblock On the information-adaptive variants of the admm: an iteration complexity perspective.
\newblock \emph{Journal of Scientific Computing}, 76:\penalty0 327--363, 2018.

\bibitem[Ghadimi \& Lan(2013)Ghadimi and Lan]{ghadimi2013stochastic}
Ghadimi, S. and Lan, G.
\newblock Stochastic first- and zeroth-order methods for nonconvex stochastic programming.
\newblock \emph{SIAM Journal on Optimization}, 23\penalty0 (4):\penalty0 2341--2368, 2013.

\bibitem[Ilyas et~al.(2018)Ilyas, Engstrom, Athalye, and Lin]{ilyas2018black}
Ilyas, A., Engstrom, L., Athalye, A., and Lin, J.
\newblock Black-box adversarial attacks with limited queries and information.
\newblock In \emph{International Conference on Machine Learning}, pp.\  2137--2146, 2018.

\bibitem[Jahani-Nezhad et~al.(2023)Jahani-Nezhad, Maddah-Ali, Li, and Caire]{jahani2023swiftagg+}
Jahani-Nezhad, T., Maddah-Ali, M.~A., Li, S., and Caire, G.
\newblock {SwiftAgg+}: Achieving asymptotically optimal communication loads in secure aggregation for federated learning.
\newblock \emph{IEEE Journal on Selected Areas in Communications}, 41\penalty0 (4):\penalty0 977--989, 2023.

\bibitem[Johnson \& Lindenstrauss(1984)Johnson and Lindenstrauss]{johnson1984extension}
Johnson, W. and Lindenstrauss, J.
\newblock Extensions of lipschitz maps into a hilbert space.
\newblock \emph{Contemporary Mathematics}, 26:\penalty0 189--206, 01 1984.

\bibitem[Karimireddy et~al.(2019)Karimireddy, Rebjock, Stich, and Jaggi]{karimireddy2019error}
Karimireddy, S.~P., Rebjock, Q., Stich, S., and Jaggi, M.
\newblock Error feedback fixes {SignSGD} and other gradient compression schemes.
\newblock In \emph{International Conference on Machine Learning}, pp.\  3252--3261, 2019.

\bibitem[Karimireddy et~al.(2022)Karimireddy, He, and Jaggi]{karimireddy2022byzantinerobust}
Karimireddy, S.~P., He, L., and Jaggi, M.
\newblock Byzantine-robust learning on heterogeneous datasets via bucketing.
\newblock In \emph{International Conference on Learning Representations}, 2022.

\bibitem[Kiefer \& Wolfowitz(1952)Kiefer and Wolfowitz]{kiefer1952stochastic}
Kiefer, J. and Wolfowitz, J.
\newblock {Stochastic Estimation of the Maximum of a Regression Function}.
\newblock \emph{The Annals of Mathematical Statistics}, 23\penalty0 (3):\penalty0 462 -- 466, 1952.

\bibitem[Li et~al.(2020)Li, Cheng, Wang, Liu, and Chen]{li2020learning}
Li, S., Cheng, Y., Wang, W., Liu, Y., and Chen, T.
\newblock Learning to detect malicious clients for robust federated learning.
\newblock \emph{arXiv preprint arXiv:2002.00211}, 2020.

\bibitem[Li(2024)]{li2024simple}
Li, Y.
\newblock Simple, unified analysis of johnson-lindenstrauss with applications.
\newblock \emph{arXiv preprint arXiv:2402.10232}, 2024.

\bibitem[Liu et~al.(2020)Liu, Chen, Kailkhura, Zhang, Hero~III, and Varshney]{liu2020primer}
Liu, S., Chen, P.-Y., Kailkhura, B., Zhang, G., Hero~III, A.~O., and Varshney, P.~K.
\newblock A primer on zeroth-order optimization in signal processing and machine learning: Principals, recent advances, and applications.
\newblock \emph{IEEE Signal Processing Magazine}, 37\penalty0 (5):\penalty0 43--54, 2020.

\bibitem[Liu et~al.(2021)Liu, Gupta, and Vaidya]{liu2021approximate}
Liu, S., Gupta, N., and Vaidya, N.~H.
\newblock Approximate byzantine fault-tolerance in distributed optimization.
\newblock In \emph{ACM Symposium on Principles of Distributed Computing}, pp.\  379--389, 2021.

\bibitem[Liu(2019)]{liu2019roberta}
Liu, Y.
\newblock Roberta: A robustly optimized bert pretraining approach.
\newblock \emph{arXiv preprint arXiv:1907.11692}, 364, 2019.

\bibitem[M~Abdelmoniem et~al.(2021)M~Abdelmoniem, Elzanaty, Alouini, and Canini]{m2021efficient}
M~Abdelmoniem, A., Elzanaty, A., Alouini, M.-S., and Canini, M.
\newblock An efficient statistical-based gradient compression technique for distributed training systems.
\newblock \emph{Proceedings of Machine Learning and Systems}, 3:\penalty0 297--322, 2021.

\bibitem[Makkuva et~al.(2024)Makkuva, Bondaschi, Vogels, Jaggi, Kim, and Gastpar]{makkuva2024laser}
Makkuva, A.~V., Bondaschi, M., Vogels, T., Jaggi, M., Kim, H., and Gastpar, M.
\newblock {LASER}: Linear compression in wireless distributed optimization.
\newblock In \emph{International Conference on Machine Learning}, 2024.

\bibitem[Malladi et~al.(2023)Malladi, Gao, Nichani, Damian, Lee, Chen, and Arora]{malladi2023fine}
Malladi, S., Gao, T., Nichani, E., Damian, A., Lee, J.~D., Chen, D., and Arora, S.
\newblock Fine-tuning language models with just forward passes.
\newblock \emph{Advances in Neural Information Processing Systems}, 36:\penalty0 53038--53075, 2023.

\bibitem[McMahan et~al.(2017)McMahan, Moore, Ramage, Hampson, and y~Arcas]{mcmahan2017communication}
McMahan, B., Moore, E., Ramage, D., Hampson, S., and y~Arcas, B.~A.
\newblock Communication-efficient learning of deep networks from decentralized data.
\newblock In \emph{Artificial intelligence and statistics}, pp.\  1273--1282, 2017.

\bibitem[Neto et~al.(2024)Neto, Egger, Bakshi, and Bitar]{neto2024communication}
Neto, A. d. S.~D., Egger, M., Bakshi, M., and Bitar, R.
\newblock Communication-efficient byzantine-resilient federated zero-order optimization.
\newblock \emph{submitted to ICML 2024; arXiv preprint arXiv:2406.14362}, 2024.

\bibitem[Qin et~al.(2023)Qin, Chen, Qian, Ding, Li, and Deng]{qin2023federated}
Qin, Z., Chen, D., Qian, B., Ding, B., Li, Y., and Deng, S.
\newblock Federated full-parameter tuning of billion-sized language models with communication cost under 18 kilobytes.
\newblock \emph{arXiv preprint arXiv:2312.06353}, 2023.

\bibitem[Qiu et~al.(2023)Qiu, Shanbhag, and Yousefian]{qiu2023zeroth}
Qiu, Y., Shanbhag, U., and Yousefian, F.
\newblock Zeroth-order methods for nondifferentiable, nonconvex, and hierarchical federated optimization.
\newblock \emph{Advances in Neural Information Processing Systems}, 36, 2023.

\bibitem[Rodr{\'\i}guez-Barroso et~al.(2023)Rodr{\'\i}guez-Barroso, Jim{\'e}nez-L{\'o}pez, Luz{\'o}n, Herrera, and Mart{\'\i}nez-C{\'a}mara]{rodriguez2023survey}
Rodr{\'\i}guez-Barroso, N., Jim{\'e}nez-L{\'o}pez, D., Luz{\'o}n, M.~V., Herrera, F., and Mart{\'\i}nez-C{\'a}mara, E.
\newblock Survey on federated learning threats: Concepts, taxonomy on attacks and defences, experimental study and challenges.
\newblock \emph{Information Fusion}, 90:\penalty0 148--173, 2023.

\bibitem[Salimans et~al.(2017)Salimans, Ho, Chen, Sidor, and Sutskever]{salimans2017evolution}
Salimans, T., Ho, J., Chen, X., Sidor, S., and Sutskever, I.
\newblock Evolution strategies as a scalable alternative to reinforcement learning.
\newblock \emph{arXiv preprint arXiv:1703.03864}, 2017.

\bibitem[Salmon et~al.(2011)Salmon, Moraes, Dror, and Shaw]{salmon2011parallel}
Salmon, J.~K., Moraes, M.~A., Dror, R.~O., and Shaw, D.~E.
\newblock Parallel random numbers: as easy as 1, 2, 3.
\newblock In \emph{International conference for high performance computing, networking, storage and analysis}, pp.\  1--12, 2011.

\bibitem[Schlegel et~al.(2023)Schlegel, Kumar, Rosnes, and i~Amat]{schlegel2023codedpaddedfl}
Schlegel, R., Kumar, S., Rosnes, E., and i~Amat, A.~G.
\newblock {CodedPaddedFL} and {CodedSecAgg}: Straggler mitigation and secure aggregation in federated learning.
\newblock \emph{IEEE Transactions on Communications}, 2023.

\bibitem[Shen et~al.(2016)Shen, Tople, and Saxena]{shen2016auror}
Shen, S., Tople, S., and Saxena, P.
\newblock Auror: Defending against poisoning attacks in collaborative deep learning systems.
\newblock In \emph{Annual Conference on Computer Security Applications}, pp.\  508--519, 2016.

\bibitem[Socher et~al.(2013)Socher, Perelygin, Wu, Chuang, Manning, Ng, and Potts]{socher2013recursive}
Socher, R., Perelygin, A., Wu, J., Chuang, J., Manning, C.~D., Ng, A.~Y., and Potts, C.
\newblock Recursive deep models for semantic compositionality over a sentiment treebank.
\newblock In \emph{Conference on empirical methods in natural language processing}, pp.\  1631--1642, 2013.

\bibitem[Spall(1992)]{spall1992multivariate}
Spall, J.
\newblock Multivariate stochastic approximation using a simultaneous perturbation gradient approximation.
\newblock \emph{IEEE Transactions on Automatic Control}, 37\penalty0 (3):\penalty0 332--341, 1992.

\bibitem[Tang et~al.(2024{\natexlab{a}})Tang, Panda, Nasr, Mahloujifar, and Mittal]{tang2024private}
Tang, X., Panda, A., Nasr, M., Mahloujifar, S., and Mittal, P.
\newblock Private fine-tuning of large language models with zeroth-order optimization.
\newblock \emph{CoRR}, 2024{\natexlab{a}}.

\bibitem[Tang et~al.(2020)Tang, Zhang, and Li]{tang2020distributed}
Tang, Y., Zhang, J., and Li, N.
\newblock Distributed zero-order algorithms for nonconvex multiagent optimization.
\newblock \emph{IEEE Transactions on Control of Network Systems}, 8\penalty0 (1):\penalty0 269--281, 2020.

\bibitem[Tang et~al.(2024{\natexlab{b}})Tang, Wang, and Chang]{tang2024z}
Tang, Z., Wang, Y., and Chang, T.-H.
\newblock {$z$-SignFedAvg}: A unified stochastic sign-based compression for federated learning.
\newblock In \emph{Proceedings of the AAAI Conference on Artificial Intelligence}, volume~38, pp.\  15301--15309, 2024{\natexlab{b}}.

\bibitem[Truex et~al.(2019)Truex, Baracaldo, Anwar, Steinke, Ludwig, Zhang, and Zhou]{truex2019hybrid}
Truex, S., Baracaldo, N., Anwar, A., Steinke, T., Ludwig, H., Zhang, R., and Zhou, Y.
\newblock A hybrid approach to privacy-preserving federated learning.
\newblock In \emph{ACM workshop on artificial intelligence and security}, pp.\  1--11, 2019.

\bibitem[Vogels et~al.(2019)Vogels, Karimireddy, and Jaggi]{vogels2019powersgd}
Vogels, T., Karimireddy, S.~P., and Jaggi, M.
\newblock {PowerSGD}: Practical low-rank gradient compression for distributed optimization.
\newblock \emph{Advances in Neural Information Processing Systems}, 32, 2019.

\bibitem[Voorhees \& Tice(2000)Voorhees and Tice]{voorhees2000building}
Voorhees, E.~M. and Tice, D.~M.
\newblock Building a question answering test collection.
\newblock In \emph{Annual international ACM SIGIR conference on Research and development in information retrieval}, pp.\  200--207, 2000.

\bibitem[Wang et~al.(2021)Wang, Liu, Liang, Joshi, and Poor]{wang2021novel}
Wang, J., Liu, Q., Liang, H., Joshi, G., and Poor, H.~V.
\newblock A novel framework for the analysis and design of heterogeneous federated learning.
\newblock \emph{IEEE Transactions on Signal Processing}, 69:\penalty0 5234--5249, 2021.

\bibitem[Wang et~al.(2025)Wang, Wang, Chen, and Ji]{wang2025new}
Wang, J., Wang, S., Chen, R.-R., and Ji, M.
\newblock A new theoretical perspective on data heterogeneity in federated optimization.
\newblock In \emph{International Conference on Machine Learning}, 2025.

\bibitem[Wei et~al.(2020)Wei, Li, Ding, Ma, Yang, Farokhi, Jin, Quek, and Poor]{wei2020federated}
Wei, K., Li, J., Ding, M., Ma, C., Yang, H.~H., Farokhi, F., Jin, S., Quek, T.~Q., and Poor, H.~V.
\newblock Federated learning with differential privacy: Algorithms and performance analysis.
\newblock \emph{IEEE Transactions on Information Forensics and Security}, 15:\penalty0 3454--3469, 2020.

\bibitem[Wen et~al.(2017)Wen, Xu, Yan, Wu, Wang, Chen, and Li]{wen2017terngrad}
Wen, W., Xu, C., Yan, F., Wu, C., Wang, Y., Chen, Y., and Li, H.
\newblock {TernGrad}: Ternary gradients to reduce communication in distributed deep learning.
\newblock \emph{Advances in neural information processing systems}, 30, 2017.

\bibitem[Xie et~al.(2020)Xie, Koyejo, and Gupta]{xie2020fall}
Xie, C., Koyejo, O., and Gupta, I.
\newblock Fall of empires: Breaking byzantine-tolerant sgd by inner product manipulation.
\newblock In \emph{Uncertainty in Artificial Intelligence}, pp.\  261--270, 2020.

\bibitem[Yin et~al.(2018)Yin, Chen, Kannan, and Bartlett]{yin2018byzantine}
Yin, D., Chen, Y., Kannan, R., and Bartlett, P.
\newblock Byzantine-robust distributed learning: Towards optimal statistical rates.
\newblock In \emph{International conference on machine learning}, pp.\  5650--5659, 2018.

\bibitem[Zhao et~al.(2018)Zhao, Li, Lai, Suda, Civin, and Chandra]{zhao2018federated}
Zhao, Y., Li, M., Lai, L., Suda, N., Civin, D., and Chandra, V.
\newblock Federated learning with non-iid data.
\newblock \emph{arXiv preprint arXiv:1806.00582}, 2018.

\bibitem[Zhu et~al.(2021)Zhu, Xu, Liu, and Jin]{zhu2021federated}
Zhu, H., Xu, J., Liu, S., and Jin, Y.
\newblock Federated learning on non-iid data: A survey.
\newblock \emph{Neurocomputing}, 465:\penalty0 371--390, 2021.

\bibitem[Zhu \& Philip(2019)Zhu and Philip]{zhu2019applying}
Zhu, T. and Philip, S.~Y.
\newblock Applying differential privacy mechanism in artificial intelligence.
\newblock In \emph{IEEE International Conference on Distributed Computing Systems (ICDCS)}, pp.\  1601--1609, 2019.

\end{thebibliography}
\bibliographystyle{icml2025}

\newpage
\allowdisplaybreaks
\appendix
\onecolumn

\section{Numerical Experiments}

\subsection{Experimental Details} \label{app:experimental_details}

\subsubsection{Sampling of Perturbation Vectors}
To sample the directions $\perturbvec$, for fine-tuning large language models (cf. \cref{sec:experiments_fine_tuning}) we use a practical approach similar to \citep{salimans2017evolution,malladi2023fine} that draws each coordinate independently from a standard Gaussian distribution. This minor modification has substantial practical implications by alleviating the allocation of the entire vector, and instead iteratively samples each model coordinate. Thereby, considerably reducing the memory footprint of our method.

\subsubsection{Reconstruction of the Seed}
Let $\initseed$ be a seed initially broadcast by the federator to all clients. Then, at each global and local iteration $\gepoch$ and $\lepoch$, the $\seed$-th random perturbation is sampled by setting the seed of the PRNG to $\initseed^\prime \define (\initseed, \gepoch, \lepoch, \seed)$. In this way, the perturbations $\perturbtlr$ sampled by all clients will be equivalent. The client then compute the estimate according to \cref{def:zero_order_estimate}.

\subsubsection{In-Place Model Perturbation}

Similar to \citep{salimans2017evolution,malladi2023fine}, we use in-place perturbations of the model for memory efficient zero-order optimization throughout the training phase. In particular, client $\client$ employs \cref{alg:perturb_model,alg:compute_zero_estimate} to compute the zero-order estimate as $\projitlr = \textsc{ZeroOrderEstimate}(\modeltl, \initseed^\prime, \pscale, \clientdata)$. Note that the function, instead of $\perturbtlr$, takes as input the seed $\initseed^\prime = (\initseed, \gepoch, \lepoch, \seed)$ used to reconstruct the projection $\perturbtlr$.

\begin{algorithm}[H]
   \caption{\textsc{Perturb}: Perturbing Model Parameters}
   \label{alg:perturb_model}
\begin{algorithmic}
   \STATE {\bf Input:} Model parameters $\model$, scaling factor $\pscale$, seed $\initseed^\prime$
   \STATE {\bf Output:} Perturbed model $\model$
   \STATE Initialize PRNG with seed $\initseed^\prime$
   \FOR{$p = 1$ to $\dimension$}
       \STATE Sample $z \sim \mathcal{N}(0, 1)$ 
       \STATE Perturb parameter $\model^{(p)} = \model^{(p)} + \pscale \cdot z$
   \ENDFOR
   \STATE {\bf Return:} Perturbed model $\model$
\end{algorithmic}
\end{algorithm}

\begin{algorithm}[H]
   \caption{\textsc{ZeroOrderEstimate}: Compute $\projtmpargs{\model}{\initseed^\prime}{\globaldata}$ via Model Perturbation}
   \label{alg:compute_zero_estimate}
\begin{algorithmic}
   \STATE {\bf Input:} Model $\model$, seed $\initseed^\prime$, scaling factor $\pscale$, data $\globaldata$
   \STATE {\bf Output:} Zero estimate $\projtmpargs{\model}{\initseed^\prime}{\globaldata}$
   \STATE {\bf Step 1:} $\texttt{Perturb}(\model, \pscale, \initseed^\prime)$  (cf. \cref{alg:perturb_model})
   \STATE Compute $\loss_1 = \loss(\model^{(p)}, \globaldata)$
   \STATE {\bf Step 2:} $\texttt{Perturb}(\model, -2\pscale, \initseed^\prime)$
   \STATE Compute $\loss_2 = \loss(\model^{(p)}, \globaldata)$
   \STATE {\bf Step 3:} $\texttt{Perturb}(\model, \pscale, \initseed^\prime)$ \COMMENT{Reset the model}
   \STATE {\bf Return:} $\projtmpargs{\model}{\initseed^\prime}{\globaldata} =  \frac{\loss_1 - \loss_2}{2\pscale}$
\end{algorithmic}
\end{algorithm}

\subsection{Byzantine Attacks} \label{sec:byzantine_attacks}

\newcommand{\byzscale}{\ensuremath{\omega}}
\newcommand{\attackvec}{\ensuremath{\mathbf{a}_{\gepoch, \lepoch}}}
We test and compare our algorithm using several state-of-the-art gradient attacks, i.e., \textit{A little is enough} (ALIE) \citep{baruch2019little}, \textit{Fall of Empires} (FOE) \citep{xie2020fall}, \textit{Sign Flipping} (SF) \citep{allen2021byzantine}, \textit{Label Flipping} (LF) \citep{allen2021byzantine}, and a tailored trimmed mean attack (TMA) (cf. \cref{alg:fk_attack}). For all non-zero-order experiments, we conduct the attacks on the gradients $\graditl$. For the zero-order experiments, the attacks are conducted on the projected gradients, i.e., on $\projitl \define \frac{1}{\nseeds} ((\projitlr)_{\seed=1}^\nseeds)^\mathrm{T}$, which we believe is the strongest attack scenario. Let in the following $\mathbf{g}^\client_{\gepoch, \lepoch}$ denote the contribution of client $\client$ at global epoch $\gepoch$ and local epoch $\lepoch$. The attacks are summarized as follows. Let $\bar{\mathbf{g}}_{\gepoch, \lepoch} \define \frac{1}{\vert \honest \vert} \sumhc \mathbf{g}^\client_{\gepoch, \lepoch}$ be the average of the honest clients gradients. For ALIE, FOE, and SF, the Byzantine clients $\client \in \byzantine \define [\nclients] \setminus \honest$ compute their corrupted gradient as $\mathbf{g}^\client_{\gepoch,\lepoch} = \bar{\mathbf{g}}_{\gepoch, \lepoch} + \byzscale \attackvec$ for some optimized $\byzscale$, where
\begin{itemize}
\item for ALIE, we have $\attackvec = \sigma_{\gepoch,\lepoch}$, where $\sigma_{\gepoch,\lepoch}$ is the coordinate standard deviation of $\bar{\mathbf{g}}_{\gepoch, \lepoch}$,
\item for FOE, we have $\attackvec = -\bar{\mathbf{g}}_{\gepoch, \lepoch}$, and hence $\mathbf{g}^\client_{\gepoch,\lepoch} = (1-\byzscale) \bar{\mathbf{g}}_{\gepoch, \lepoch}$,
\item for SF, we have $\attackvec = -\bar{\mathbf{g}}_{\gepoch, \lepoch}$ for fixed $\byzscale=2$, s.t. $\mathbf{g}^\client_{\gepoch,\lepoch} = -\bar{\mathbf{g}}_{\gepoch, \lepoch}$.
\end{itemize}
For ALIE and FOE, similar to \citet{allouah2023fixing}, we linearly optimize of potential choices of $\byzscale$ such that the L2 distance of the final aggregation $\raggtl$ to the honest clients' average $\bar{\mathbf{g}}^\client_{\gepoch, \lepoch}$ is maximized. For LF, each Byzantine workers manipulates the labels of its local dataset. In particular, if for a Byzantine client $\client \in \byzantine$ a sample in $\clientdata$ is labeled $\ell$, they instead train on the label $\ell^\prime = 9-\ell$ for a $10$-class classification task.

The details of the tailored trimmed mean attack can be found in \cref{alg:fk_attack}.

\begin{algorithm}[!t]
\caption{Transformed Trimmed-Mean Attack (TMA)}
\label{alg:fk_attack}
\begin{algorithmic}[1]
\REQUIRE $\beta$, $\nclients$, $\projitlr \forall \client \in [\nclients]$, honest clients $\honest$
\STATE Compute $\projitlravg = \frac{1}{\nclients} \sum_{\client=1}^\nclients \projitlr$
\FOR{Byzantine client $\client \in [\nclients]\setminus\honest$} 
    \IF{$\projitlravg > 0$}
    \STATE %
    return $\lfloor\beta \nclients\rfloor$ smallest value in $\{\projitlr\}_{\client \in \honest}$
    \ELSE
    \STATE %
    return $\lfloor\beta \nclients\rfloor$ largest value in $\{\projitlr\}_{\client \in \honest}$
    \ENDIF
\ENDFOR
\end{algorithmic}
\end{algorithm}

\subsection{Hyperparameters} \label{app:hyperparameters}

We detail in the following \cref{tab:experiments_mnist,tab:experiments_fine_tuning} the hyperparameters used through the experiments in \cref{sec:experiments_mnist,sec:experiments_fine_tuning}.
\begin{table}[H]
\caption{Simulation Parameters and Hyperparameters for \cref{sec:experiments_mnist}}
\label{tab:experiments_mnist}
\vskip 0.15in
\begin{center}
\begin{tabular}{lc}
\toprule
\multicolumn{2}{c}{MNIST}  \\
\midrule
Global Train Samples   & $60000$ \\
Number of Clients      & $40$ \\
Number of Byzantine Clients & $10$ \\
Scaling Factor $\pscale$ & $0.001$ \\
Learning Rate $\lr$         & $0.01$ \\
Batch Size      & $64$ \\
Global Epochs $\ngepochs$  & 400 \\
\bottomrule
\end{tabular}
\end{center}
\vskip -0.1in
\end{table}

\begin{table}[H]
\caption{Simulation Parameters and Hyperparameters for \cref{sec:experiments_fine_tuning}}
\label{tab:experiments_fine_tuning}
\vskip 0.15in
\begin{center}
\begin{tabular}{lccc}
\toprule
& SST-2 & SNLI & TREC \\
\midrule
Global Train Samples   & \multicolumn{3}{c}{\xrfill[3pt]{0.5pt} $\;$$512$ \xrfill[3pt]{0.5pt}} \\
Scaling Factor $\pscale$ & \multicolumn{3}{c}{\xrfill[3pt]{0.5pt} $\;$$0.001$ \xrfill[3pt]{0.5pt}} \\
Learning Rate $\lr$         & \multicolumn{3}{c}{\xrfill[3pt]{0.5pt} $\;$$10^{-6}$ \xrfill[3pt]{0.5pt}} \\
Batch Size      & \multicolumn{3}{c}{\xrfill[3pt]{0.5pt} $\;$64 \xrfill[3pt]{0.5pt}} \\
Global Epochs $\ngepochs$         & 20,000      & 20,000       & 40,000       \\
\bottomrule
\end{tabular}
\end{center}
\vskip -0.1in
\end{table}

The numerical experiments were conducted on the following cluster of simulation servers.
\begin{table}[H]
\centering
\begin{tabular}{|l|l|l|l|}
\hline
\textbf{CPU(s)}                                  & \textbf{RAM}   & \textbf{GPU(s)}                         & \textbf{VRAM} \\ \hline
2x Intel Xeon Platinum 8176 (56 cores)         & 256 GB         & 2x NVIDIA GeForce GTX 1080 Ti        & 11 GB                 \\ 
2x AMD EPYC 7282 (32 cores)                    & 512 GB         & NVIDIA GeForce RTX 4090              & 24 GB                 \\ 
2x AMD EPYC 7282 (32 cores)                    & 640 GB         & NVIDIA GeForce RTX 4090              & 24 GB                 \\ 
2x AMD EPYC 7282 (32 cores)                    & 448 GB         & NVIDIA GeForce RTX 4080              & 16 GB                 \\ 
2x AMD EPYC 7282 (32 cores)                    & 256 GB         & NVIDIA GeForce RTX 4080              & 16 GB                 \\ 
HGX-A100 (96 cores)                            & 1 TB           & 4x NVIDIA A100                      & 80 GB                 \\ 
DGX-A100 (252 cores)                           & 2 TB           & 8x NVIDIA Tesla A100                & 80 GB                 \\ 
DGX-1-V100 (76 cores)                          & 512 GB         & 8x NVIDIA Tesla V100                & 16 GB                 \\ 
DGX-1-P100 (76 cores)                          & 512 GB         & 8x NVIDIA Tesla P100                & 16 GB                 \\ 
HPE-P100 (28 cores)                            & 256 GB         & 4x NVIDIA Tesla P100                & 16 GB                 \\ \hline
\end{tabular}
\caption{System specifications of our simulation cluster.}
\end{table}

\subsection{Local Iterations} \label{app:local_iterations}

\algname offers different option to conduct local epochs at the clients. We term the approach for local epochs introduced in \cref{alg:federated_zero_order} ``Unbiased''.

A second approach is to follow \cref{alg:federated_zero_order}, but to replace the sending of $\{\projitl\}_{\lepoch=1}^{\nlepochs}$ from clients to the federator, followed by $\raggtl = \robustaggop(\{\projitl\}_{\client=1}^{\nclients}), \lepoch \in [\nlepochs]$ and $\model^{(\gepoch+1)} = \model^{(\gepoch)} - \lr \sum_{\lepoch=1}^{\mathcal{L}} \perturbtl \raggtl$. Instead of transmitting the results $\{\projitl\}_{\lepoch=1}^{\nlepochs}$ for all local epochs, the clients can instead reconstruct the aggregated local gradient updates as $\gradi = \sum_{\lepoch=1}^\nlepochs \perturbtl \projitl$ and project this gradient approximation onto random directions $\perturbttx$ (known to the federator and all clients) according to \cref{def:zero_order_estimate}, and only transmit the result of $\perturbttx^\mathrm{T} \gradi$ to the federator. The federator conducts the transformed aggregation on $\raggtl = \robustaggop(\{\perturbttx^\mathrm{T} \gradi\}_{\client=1}^{\nclients})$ and updates the global model as $\model^{(\gepoch+1)} = \model^{(\gepoch)} - \lr \perturbttx \raggtl$. This approach is by a factor of $\nlepochs$ more communication efficient. However, it can introduce significant additional variance, especially for small values of $\nseeds$. This is because two randomly drawn vectors in high dimensions are likely almost orthogonal, and hence the subspaces resulting from $\perturbttx$ and $\{\perturbtl\}_{\lepoch=1}^\nlepochs$ might only be weakly dependent. However, for large values of $\nseeds$, this approach might be beneficial due to the drastic savings in the cost of communication. We term this approach ``Unbiased Compressed''.

\begin{algorithm}[!t]
\caption{\algname: Robust Efficient Zero-Order FL with Biased ZO Estimator}
\label{alg:federated_zero_order_biased}
\begin{algorithmic}[1]
\REQUIRE Shared seed for PRNG, $\pscale \geq 0$, $\lr>0$, $\nseeds>0$, $\robustaggop$.
\STATE Initialize and broadcast global model $\model^{(1)}$.
\FOR{$\gepoch = 1$ to $\ngepochs$} 
    \FOR{each client $\client \in [\nclients]$ \textbf{in parallel}}
        \STATE Initialize local model $\model_{\gepoch, 1}^\client = \model^{(\gepoch)}$.
        \STATE Draw $\perturbvec_{\gepoch,1}^{1},\! \cdots\!, \perturbvec_{\gepoch,1}^{\nseeds} \sim \uniform$, let $\perturbtl[1] \define (\perturbtoner)_{\seed \in [\nseeds]}$
        \FOR{$\lepoch = 1$ to $\nlepochs$} 
            \STATE Compute $\projitoner \define \projtmpargs{\modeltl}{\perturbtoner}{\clientdata}, \seed \in [\nseeds]$ (cf. \cref{def:zero_order_estimate})
            \STATE Let $\projitone \define \frac{1}{\nseeds} ((\projitoner)_{\seed=1}^\nseeds)^\top$
            \STATE Update $\modeltll[\lepoch+1] = \modeltl - \lr \perturbtl[1] \projitone$
        \ENDFOR
        \STATE Send $\{\sum_{\lepoch=1}^\nlepochs \projitone\}$ to federator.
    \ENDFOR
    \STATE Aggregate $\raggt = \robustaggop(\{\sum_{\lepoch=1}^\nlepochs \projitl\}_{\client=1}^{\nclients})$
    \STATE Update $\model^{(\gepoch+1)} = \model^{(\gepoch)} - \lr \sum_{\lepoch=1}^{\nlepochs} \perturbtl[1] \raggt.$
    \STATE Broadcast $\raggt$; clients accordingly update using $\perturbtl[1]$
\ENDFOR
\STATE Clients recover the updated global model $\model^{(\gepoch+1)}$ using known perturbations $\perturbtl[1]$.
\end{algorithmic}
\end{algorithm}

A third approach, termed ``Biased'', is to make the clients reuse the directions $\perturbtl$ at each iteration, i.e., $\perturbtm[\lepoch] = \perturbtm[\lepochtmp]$ for $\lepoch \neq \lepochtmp \in [\nlepochs]$. It suffices for the clients to communicate to the federator $\sum_{\lepoch=1}^{\nlepochs} \projitl$, thus reducing the communication cost by the same factor of $\nlepochs$ as for the second approach above. However, this strategy incurs bias in the local training process, since the gradient updates are not uniformly and independently chosen at each local iteration $\lepoch \in [\nlepochs]$. We summarize this approach in \cref{alg:federated_zero_order_biased}.

\begin{figure*}[!t]
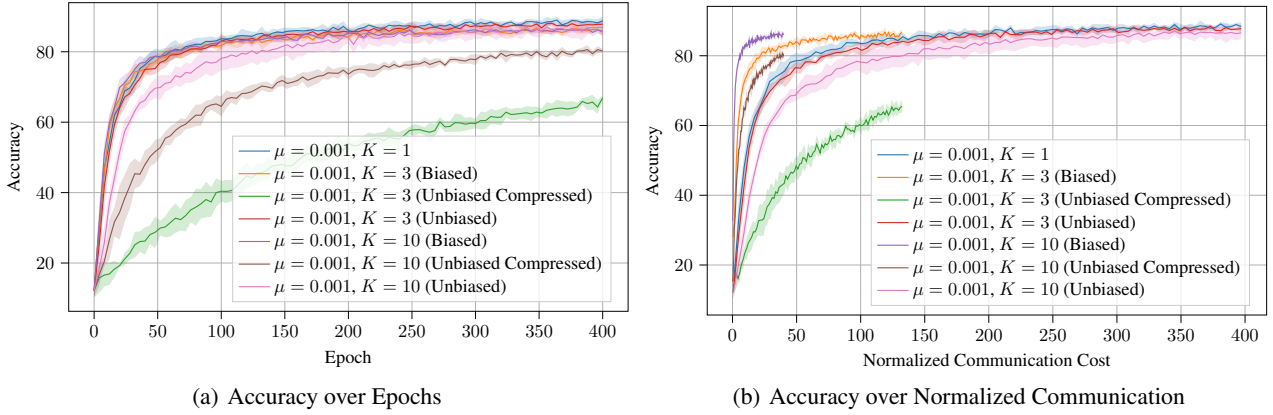

\subfigure[Accuracy over Epochs]{\resizebox{.49\linewidth}{!}{\input{data/mu_0.001_le_sampling_comparison}}\label{fig:mu0001_le_epochs}}
\subfigure[Accuracy over Normalized Communication]{\resizebox{.49\linewidth}{!}{\input{data/mu_0.001_le_sampling_comparison_comm}} \label{fig:mu0_le_epochs}}
\vspace{-.1cm}
\caption{Comparisons of Local Epoch Strategies for $\pscale=0.001$} \label{fig:lr_comparison_mu0001} \vspace{-.2cm}
\end{figure*}

\begin{figure*}[!t]
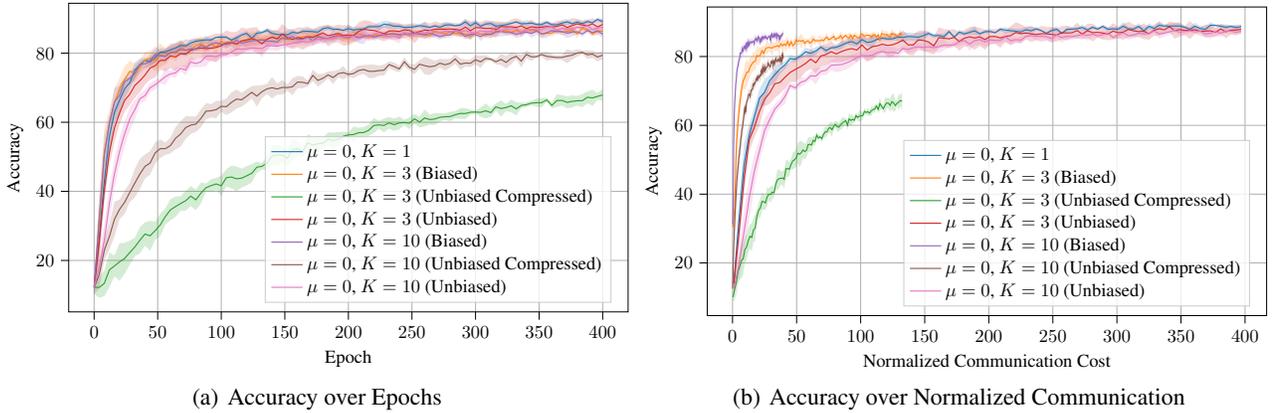

\subfigure[Accuracy over Epochs]{\resizebox{.49\linewidth}{!}{\input{data/mu_0_le_sampling_comparison}}\label{fig:mu0001_le_epochsa}}
\subfigure[Accuracy over Normalized Communication]{\resizebox{.49\linewidth}{!}{\input{data/mu_0_le_sampling_comparison_comm}} \label{fig:mu0_le_epochs_comm}}
\vspace{-.1cm}
\caption{Comparisons of Local Epoch Strategies for $\pscale=0$} \label{fig:lr_comparison_mu0} \vspace{-.2cm}
\end{figure*}

The above mentioned approaches expose an efficiency-bias-variance trade-off that we will examine in the following. In \cref{fig:mu0001_le_epochs}, we provide a study for $\pscale=0.001$ in terms of accuracies over epochs. It can be observed that Unbiased Compressed local epochs are harmful, expecially for smalle $\nlepochs$. The larger $\nlepochs$, the larger the space covered during the local training process, and the smaller the loss incurred by projection the approximated overall local gradient onto an independent subspace. Looking at the accuracies over the normlized communication cost in \cref{fig:lr_comparison_mu0001}, we can observe that Biased local epochs with reasonably large values of \nlepochs can indeed significantly improve the performance when normalized by communication cost. The Unbiased approach, although reducing the number of packets to be transmitted, does not significantly improve the factual communication cost. Results for $\pscale=0$ in \cref{fig:mu0_le_epochs} and \cref{fig:lr_comparison_mu0} exhibit the same trade offs.

\begin{figure}[H]
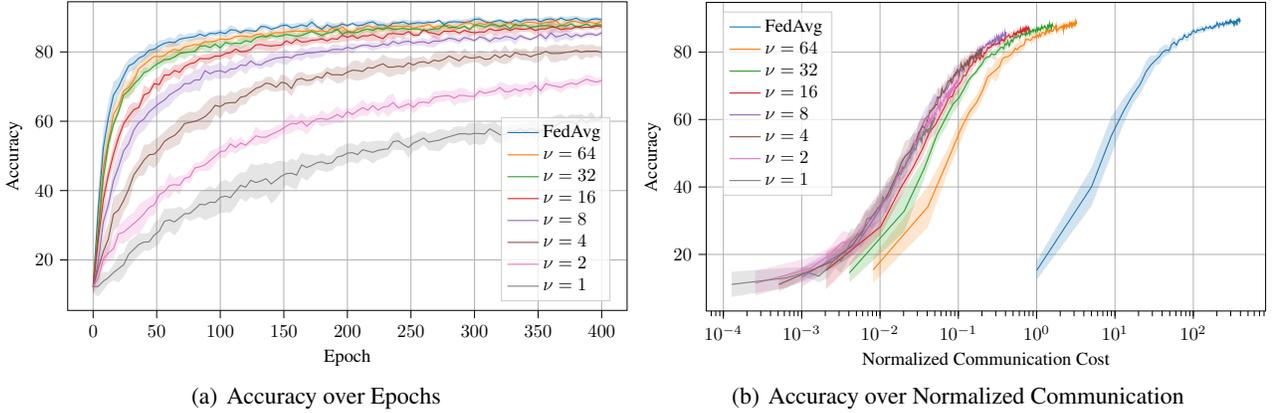

\subfigure[Accuracy over Epochs]{\resizebox{.49\linewidth}{!}{\input{data/k_comparison}}\label{fig:}}
\subfigure[Accuracy over Normalized Communication]{\resizebox{.49\linewidth}{!}{\input{data/k_comparison_comm}} \label{fig:}}
\vspace{-.1cm}
\caption{Comparison of Zero-Order Optimization for Different Values of $\nseeds$ Compared to the Baseline FedAvg.} \label{fig:k_comparison} \vspace{-.2cm}
\end{figure}

{\subsection{Effect of Number of Perturbations} \label{app:primer_zero_order}
To highlight the effect of the number of perturbations $\nseeds$ on the convergence of zero-order optimization in standard learning tasks, we show in \cref{fig:k_comparison} the performance of \algname compared to \fedavg \citep{mcmahan2017communication} for different values of perturbations $\nseeds$. While $\nseeds=1$ exhibits a substantial performance gap to \fedavg, this gap decreases with increasing $\nseeds$, until nearly vanishing with $\nseeds=64$.}

\subsection{Accuracies over Epochs for all Attacks on MNIST} \label{app:extended_experiments_mnist}

We present in the following, extending the case of LF (cf. \cref{fig:label_flipping_comm}), the comparison of all countermeasures for ALIE, \aliennm, FOE, \foennm, and SF. We present the results for accuracies over epochs, and accuracies over normalized communication cost.

\begin{figure}[H]
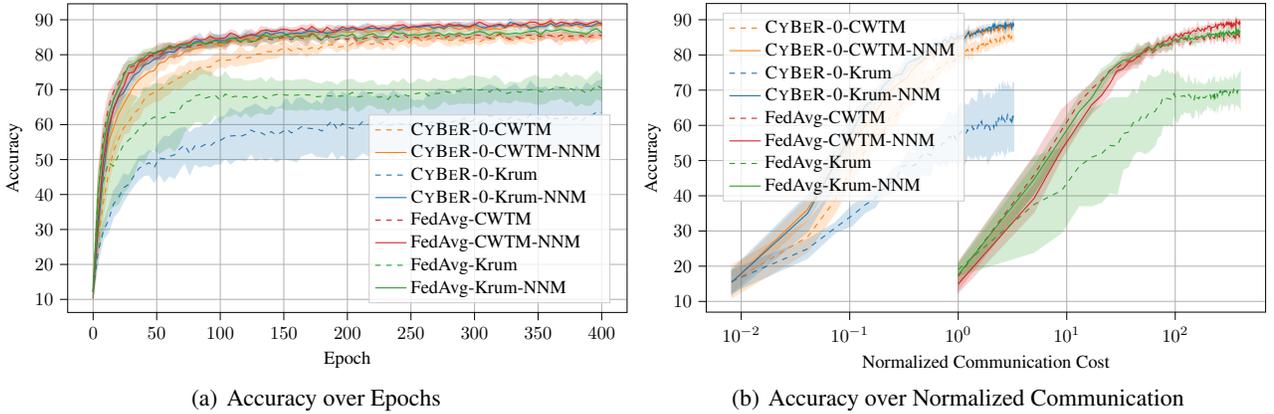

\subfigure[Accuracy over Epochs]{\resizebox{.49\linewidth}{!}{\input{data/ALIE}}\label{fig:}}
\subfigure[Accuracy over Normalized Communication]{\resizebox{.49\linewidth}{!}{\input{data/ALIE_comm}} \label{fig:}}
\vspace{-.1cm}
\caption{ALIE attack on logistic regression on MNIST.} \label{fig:} \vspace{-.2cm}
\end{figure}

\begin{figure}[H]
\subfigure[Accuracy over Epochs]{\resizebox{.49\linewidth}{!}{\input{data/ALIE-NNM}}\label{fig:}}
\subfigure[Accuracy over Normalized Communication]{\resizebox{.49\linewidth}{!}{\input{data/ALIE-NNM_comm}} \label{fig:}}
\vspace{-.1cm}
\caption{\aliennm attack on logistic regression on MNIST.} \label{fig:} \vspace{-.2cm}
\end{figure}

\begin{figure}[H]
\subfigure[Accuracy over Epochs]{\resizebox{.49\linewidth}{!}{\input{data/FOE}}\label{fig:}}
\subfigure[Accuracy over Normalized Communication]{\resizebox{.49\linewidth}{!}{\input{data/FOE_comm}} \label{fig:}}
\vspace{-.1cm}
\caption{FOE attack on logistic regression on MNIST.} \label{fig:} \vspace{-.2cm}
\end{figure}

\begin{figure}[H]
\subfigure[Accuracy over Epochs]{\resizebox{.49\linewidth}{!}{\input{data/FOE-NNM}}\label{fig:}}
\subfigure[Accuracy over Normalized Communication]{\resizebox{.49\linewidth}{!}{\input{data/FOE-NNM_comm}} \label{fig:}}
\vspace{-.1cm}
\caption{\foennm attack on logistic regression on MNIST.} \label{fig:} \vspace{-.2cm}
\end{figure}

\begin{figure}[H]
\subfigure[Accuracy over Epochs]{\resizebox{.49\linewidth}{!}{\input{data/SF}}\label{fig:}}
\subfigure[Accuracy over Normalized Communication]{\resizebox{.49\linewidth}{!}{\input{data/SF_comm}} \label{fig:}}
\vspace{-.1cm}
\caption{SF attack on logistic regression on MNIST.} \label{fig:} \vspace{-.2cm}
\end{figure}

\begin{figure}[H]
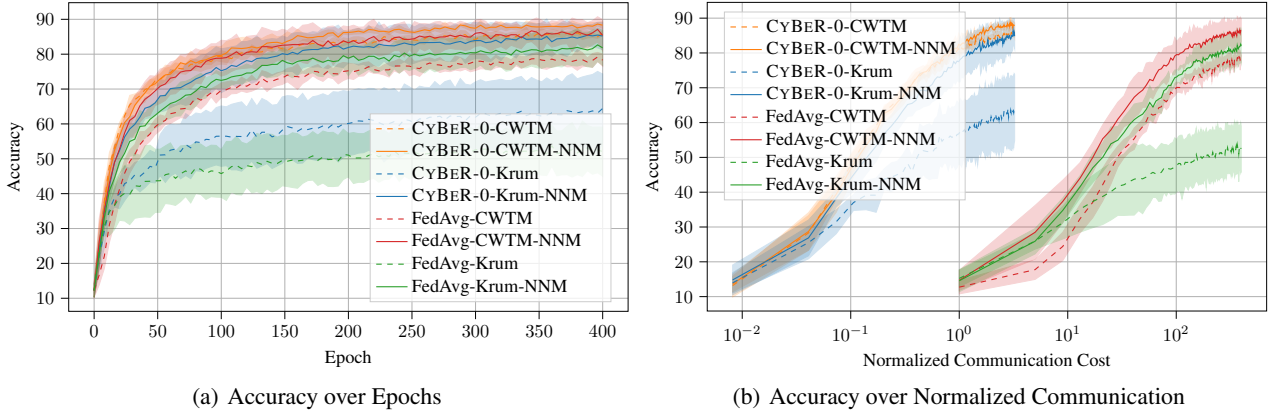

\subfigure[Accuracy over Epochs]{\resizebox{.49\linewidth}{!}{\input{data/LABEL_FLIPPING}}\label{fig:}}
\subfigure[Accuracy over Normalized Communication]{\resizebox{.49\linewidth}{!}{\input{data/LABEL_FLIPPING_comm}} \label{fig:}}
\vspace{-.1cm}
\caption{LF attack on logistic regression on MNIST.} \label{fig:} \vspace{-.2cm}
\end{figure}

\subsection{Accuracies over Epochs for all Attacks on Fine-Tuning Tasks} \label{app:extended_experiments_nlp}

We provide in \cref{tab:nlp_comparison_iid} extensive results on an \iid data distribution, analog to the \noniid results in \cref{tab:nlp_comparison_dirichlet1}. It can be found that our algorithm exhibits stable performance for both \iid and \noniid distributions, and is not significantly affected by heterogeneity. Further, we provide in the following plots for the accuracies over the epochs for all attacks, datasets, and both \iid and \noniid data distributions. \algname exhibits stable performance in all settings.

\begin{table}[H]
\centering
\begin{tabular}{ccccccc}
\toprule
Dataset & ALIE & FOE & SF & TMA & No Attack& Worst Case  \\
\midrule
SST-2 & 93.0 $\pm$ 0.4 & 91.6 $\pm$ 0.2 & 91.9 $\pm$ 0.1 & 92.1 $\pm$ 0.6 & 92.9 $\pm$ 0.6 & 91.6 \\
SNLI & 83.5 $\pm$ 0.5 & 77.0 $\pm$ 0.8 & 78.7 $\pm$ 1.0 & 79.6 $\pm$ 0.9 & 84.9 $\pm$ 0.2 & 77.0 \\
TREC & 95.5 $\pm$ 0.3 & 88.5 $\pm$ 0.9 & 90.5 $\pm$ 0.8 & 91.4 $\pm$ 1.9 & 95.6 $\pm$ 0.4 & 88.5 \\
\bottomrule
\end{tabular}
\caption{Mean and Standard Deviation of Maximum Accuracies Across Seeds}
\label{tab:nlp_comparison_iid}
\end{table}

\begin{figure}[H]
    \centering
    \resizebox{.6\linewidth}{!}{\begin{tikzpicture}

\definecolor{crimson2143940}{RGB}{214,39,40}
\definecolor{darkgray176}{RGB}{176,176,176}
\definecolor{darkorange25512714}{RGB}{255,127,14}
\definecolor{forestgreen4416044}{RGB}{44,160,44}
\definecolor{lightgray204}{RGB}{204,204,204}
\definecolor{mediumpurple148103189}{RGB}{148,103,189}
\definecolor{steelblue31119180}{RGB}{31,119,180}

\begin{axis}[
legend cell align={left},
legend style={
  fill opacity=0.8,
  draw opacity=1,
  text opacity=1,
  at={(0.03,0.97)},
  anchor=north west,
  draw=lightgray204
},
tick align=outside,
tick pos=left,
x grid style={darkgray176},
xlabel={Epoch},
xmajorgrids,
xmin=-1000, xmax=21000,
xtick style={color=black},
y grid style={darkgray176},
ylabel={Accuracy},
ymajorgrids,
ymin=0.819815132794967, ymax=0.932506093339437,
ytick style={color=black}
]
\path [fill=steelblue31119180, fill opacity=0.2]
(axis cs:0,0.878264778338004)
--(axis cs:0,0.856110221661996)
--(axis cs:2000,0.879112420933888)
--(axis cs:4000,0.887175807961502)
--(axis cs:6000,0.894850263018752)
--(axis cs:8000,0.894301627542704)
--(axis cs:10000,0.895884320749941)
--(axis cs:12000,0.899729415769501)
--(axis cs:14000,0.901624683245881)
--(axis cs:16000,0.903887832585732)
--(axis cs:18000,0.90279329008444)
--(axis cs:20000,0.903887832585732)
--(axis cs:20000,0.925539250747601)
--(axis cs:20000,0.925539250747601)
--(axis cs:18000,0.924029626582227)
--(axis cs:16000,0.925539250747601)
--(axis cs:14000,0.921943025087453)
--(axis cs:12000,0.920583084230499)
--(axis cs:10000,0.916615679250059)
--(axis cs:8000,0.912990039123963)
--(axis cs:6000,0.911139320314581)
--(axis cs:4000,0.899933567038498)
--(axis cs:2000,0.894325079066112)
--(axis cs:0,0.878264778338004)
--cycle;

\path [fill=darkorange25512714, fill opacity=0.2]
(axis cs:0,0.875657549481248)
--(axis cs:0,0.859368492185419)
--(axis cs:2000,0.882224291731404)
--(axis cs:4000,0.893241705274474)
--(axis cs:6000,0.899364689516413)
--(axis cs:8000,0.902030758472821)
--(axis cs:10000,0.902216751584259)
--(axis cs:12000,0.907185977811318)
--(axis cs:14000,0.909847025604408)
--(axis cs:16000,0.907021137766908)
--(axis cs:18000,0.907997700266908)
--(axis cs:20000,0.910506848048948)
--(axis cs:20000,0.927383776951052)
--(axis cs:20000,0.927383776951052)
--(axis cs:18000,0.927288758066425)
--(axis cs:16000,0.926312195566425)
--(axis cs:14000,0.922835266062259)
--(axis cs:12000,0.922892147188682)
--(axis cs:10000,0.920699915082408)
--(axis cs:8000,0.918932783193846)
--(axis cs:6000,0.914437393816921)
--(axis cs:4000,0.901680169725526)
--(axis cs:2000,0.899025708268596)
--(axis cs:0,0.875657549481248)
--cycle;

\path [fill=forestgreen4416044, fill opacity=0.2]
(axis cs:0,0.851494842483315)
--(axis cs:0,0.824937449183352)
--(axis cs:2000,0.846787151156667)
--(axis cs:4000,0.861379393626182)
--(axis cs:6000,0.868091380731639)
--(axis cs:8000,0.88140337630015)
--(axis cs:10000,0.884505872730751)
--(axis cs:12000,0.88254144965625)
--(axis cs:14000,0.884780436718717)
--(axis cs:16000,0.887191654625328)
--(axis cs:18000,0.888202401661711)
--(axis cs:20000,0.888202401661711)
--(axis cs:20000,0.900860098338289)
--(axis cs:20000,0.900860098338289)
--(axis cs:18000,0.900860098338289)
--(axis cs:16000,0.899266678708005)
--(axis cs:14000,0.899073729947949)
--(axis cs:12000,0.896104383677083)
--(axis cs:10000,0.890884752269249)
--(axis cs:8000,0.889429957033183)
--(axis cs:6000,0.887116952601694)
--(axis cs:4000,0.884063314707151)
--(axis cs:2000,0.869358682176666)
--(axis cs:0,0.851494842483315)
--cycle;

\path [fill=crimson2143940, fill opacity=0.2]
(axis cs:0,0.853467139174927)
--(axis cs:0,0.829475569158406)
--(axis cs:2000,0.853580947639485)
--(axis cs:4000,0.867944878906077)
--(axis cs:6000,0.873655232268975)
--(axis cs:8000,0.878526662777608)
--(axis cs:10000,0.882444294768975)
--(axis cs:12000,0.885017542066767)
--(axis cs:14000,0.886092785548948)
--(axis cs:16000,0.887315725277608)
--(axis cs:18000,0.888083666731404)
--(axis cs:20000,0.889473566213581)
--(axis cs:20000,0.904797267119753)
--(axis cs:20000,0.904797267119753)
--(axis cs:18000,0.904885083268596)
--(axis cs:16000,0.903048858055726)
--(axis cs:14000,0.902969714451052)
--(axis cs:12000,0.899487666266567)
--(axis cs:10000,0.894899455231025)
--(axis cs:8000,0.894259795555726)
--(axis cs:6000,0.886110392731025)
--(axis cs:4000,0.87879991276059)
--(axis cs:2000,0.872981552360515)
--(axis cs:0,0.853467139174927)
--cycle;

\path [fill=mediumpurple148103189, fill opacity=0.2]
(axis cs:0,0.860356811470879)
--(axis cs:0,0.840164021862454)
--(axis cs:2000,0.868040967709315)
--(axis cs:4000,0.878929541166518)
--(axis cs:6000,0.885325229995558)
--(axis cs:8000,0.886883106260428)
--(axis cs:10000,0.888499464153494)
--(axis cs:12000,0.890095447375577)
--(axis cs:14000,0.891187177232349)
--(axis cs:16000,0.889428738124686)
--(axis cs:18000,0.890761427815978)
--(axis cs:20000,0.8910504112646)
--(axis cs:20000,0.914288130402067)
--(axis cs:20000,0.914288130402067)
--(axis cs:18000,0.913926072184022)
--(axis cs:16000,0.911352511875314)
--(axis cs:14000,0.911547197767651)
--(axis cs:12000,0.910034760957756)
--(axis cs:10000,0.907724494179839)
--(axis cs:8000,0.903481477072905)
--(axis cs:6000,0.900482061671109)
--(axis cs:4000,0.892554833833482)
--(axis cs:2000,0.879354865624018)
--(axis cs:0,0.860356811470879)
--cycle;

\addplot [semithick, steelblue31119180]
table {%
0 0.8671875
2000 0.88671875
4000 0.8935546875
6000 0.902994791666667
8000 0.903645833333333
10000 0.90625
12000 0.91015625
14000 0.911783854166667
16000 0.914713541666667
18000 0.913411458333333
20000 0.914713541666667
};
\addlegendentry{No Attack}
\addplot [semithick, darkorange25512714]
table {%
0 0.867513020833333
2000 0.890625
4000 0.8974609375
6000 0.906901041666667
8000 0.910481770833333
10000 0.911458333333333
12000 0.9150390625
14000 0.916341145833333
16000 0.916666666666667
18000 0.917643229166667
20000 0.9189453125
};
\addlegendentry{ALIE}
\addplot [semithick, forestgreen4416044]
table {%
0 0.838216145833333
2000 0.858072916666667
4000 0.872721354166667
6000 0.877604166666667
8000 0.885416666666667
10000 0.8876953125
12000 0.889322916666667
14000 0.891927083333333
16000 0.893229166666667
18000 0.89453125
20000 0.89453125
};
\addlegendentry{FOE}
\addplot [semithick, crimson2143940]
table {%
0 0.841471354166667
2000 0.86328125
4000 0.873372395833333
6000 0.8798828125
8000 0.886393229166667
10000 0.888671875
12000 0.892252604166667
14000 0.89453125
16000 0.895182291666667
18000 0.896484375
20000 0.897135416666667
};
\addlegendentry{SF}
\addplot [semithick, mediumpurple148103189]
table {%
0 0.850260416666667
2000 0.873697916666667
4000 0.8857421875
6000 0.892903645833333
8000 0.895182291666667
10000 0.898111979166667
12000 0.900065104166667
14000 0.9013671875
16000 0.900390625
18000 0.90234375
20000 0.902669270833333
};
\addlegendentry{TMA}
\end{axis}

\end{tikzpicture}}
    \caption{Accuracy comparison of different attacks on fine-tuning RoBERTa-large on SST-2 with \iid data, compared to the baseline.}
    \label{fig:nlp_comparison_iid_eval_acc_SST}
\end{figure}

\begin{figure}[H]
    \centering
    \resizebox{.6\linewidth}{!}{\begin{tikzpicture}

\definecolor{crimson2143940}{RGB}{214,39,40}
\definecolor{darkgray176}{RGB}{176,176,176}
\definecolor{darkorange25512714}{RGB}{255,127,14}
\definecolor{forestgreen4416044}{RGB}{44,160,44}
\definecolor{lightgray204}{RGB}{204,204,204}
\definecolor{mediumpurple148103189}{RGB}{148,103,189}
\definecolor{steelblue31119180}{RGB}{31,119,180}

\begin{axis}[
legend cell align={left},
legend style={
  fill opacity=0.8,
  draw opacity=1,
  text opacity=1,
  at={(0.03,0.97)},
  anchor=north west,
  draw=lightgray204
},
tick align=outside,
tick pos=left,
x grid style={darkgray176},
xlabel={Epoch},
xmajorgrids,
xmin=-1000, xmax=21000,
xtick style={color=black},
y grid style={darkgray176},
ylabel={Accuracy},
ymajorgrids,
ymin=0.596365110573217, ymax=0.84102043747765,
ytick style={color=black}
]
\path [fill=steelblue31119180, fill opacity=0.2]
(axis cs:0,0.694039948578898)
--(axis cs:0,0.679657968087769)
--(axis cs:2000,0.730214157245535)
--(axis cs:4000,0.754996252852736)
--(axis cs:6000,0.778552977877794)
--(axis cs:8000,0.791388694997039)
--(axis cs:10000,0.798215696611362)
--(axis cs:12000,0.804436147731225)
--(axis cs:14000,0.810074217533157)
--(axis cs:16000,0.815670542722839)
--(axis cs:18000,0.817669703644269)
--(axis cs:20000,0.817669703644269)
--(axis cs:20000,0.829899740800176)
--(axis cs:20000,0.829899740800176)
--(axis cs:18000,0.829899740800176)
--(axis cs:16000,0.827992651721605)
--(axis cs:14000,0.822304254689065)
--(axis cs:12000,0.822299963379886)
--(axis cs:10000,0.811593331166416)
--(axis cs:8000,0.801493249447406)
--(axis cs:6000,0.791325494344428)
--(axis cs:4000,0.774083608258375)
--(axis cs:2000,0.751556676087799)
--(axis cs:0,0.694039948578898)
--cycle;

\path [fill=darkorange25512714, fill opacity=0.2]
(axis cs:0,0.687853035082949)
--(axis cs:0,0.674560159361496)
--(axis cs:2000,0.728791772849746)
--(axis cs:4000,0.755442065365965)
--(axis cs:6000,0.772830006759657)
--(axis cs:8000,0.790951159170838)
--(axis cs:10000,0.802644484903999)
--(axis cs:12000,0.807214809786445)
--(axis cs:14000,0.810139093282034)
--(axis cs:16000,0.816487105686204)
--(axis cs:18000,0.81240449429398)
--(axis cs:20000,0.816587219629516)
--(axis cs:20000,0.827075974814928)
--(axis cs:20000,0.827075974814928)
--(axis cs:18000,0.827786477928242)
--(axis cs:16000,0.826308033202684)
--(axis cs:14000,0.822673406717966)
--(axis cs:12000,0.816483106880222)
--(axis cs:10000,0.803258292873779)
--(axis cs:8000,0.797590507495829)
--(axis cs:6000,0.785763743240343)
--(axis cs:4000,0.766693351300701)
--(axis cs:2000,0.743864477150254)
--(axis cs:0,0.687853035082949)
--cycle;

\path [fill=forestgreen4416044, fill opacity=0.2]
(axis cs:0,0.619944748304864)
--(axis cs:0,0.607485807250691)
--(axis cs:2000,0.656590657962368)
--(axis cs:4000,0.674744135549755)
--(axis cs:6000,0.690968722908303)
--(axis cs:8000,0.706462234457023)
--(axis cs:10000,0.717885964599292)
--(axis cs:12000,0.730673726109968)
--(axis cs:14000,0.739155693847987)
--(axis cs:16000,0.747845401849569)
--(axis cs:18000,0.751569107503971)
--(axis cs:20000,0.751569107503971)
--(axis cs:20000,0.767528114718251)
--(axis cs:20000,0.767528114718251)
--(axis cs:18000,0.767528114718251)
--(axis cs:16000,0.753890709261542)
--(axis cs:14000,0.746521389485347)
--(axis cs:12000,0.73851030166781)
--(axis cs:10000,0.729596674289597)
--(axis cs:8000,0.71671484887631)
--(axis cs:6000,0.704430582647253)
--(axis cs:4000,0.6854989200058)
--(axis cs:2000,0.66979823092652)
--(axis cs:0,0.619944748304864)
--cycle;

\path [fill=crimson2143940, fill opacity=0.2]
(axis cs:0,0.624726922175949)
--(axis cs:0,0.623102938935162)
--(axis cs:2000,0.670920990005562)
--(axis cs:4000,0.687942068087472)
--(axis cs:6000,0.706864087749569)
--(axis cs:8000,0.721519656214434)
--(axis cs:10000,0.728562247215375)
--(axis cs:12000,0.742617787270596)
--(axis cs:14000,0.747185604443079)
--(axis cs:16000,0.757781654595545)
--(axis cs:18000,0.766674389322272)
--(axis cs:20000,0.766674389322272)
--(axis cs:20000,0.769349916233283)
--(axis cs:20000,0.769349916233283)
--(axis cs:18000,0.769349916233283)
--(axis cs:16000,0.763919734293344)
--(axis cs:14000,0.761060923334699)
--(axis cs:12000,0.755212073840515)
--(axis cs:10000,0.746698169451292)
--(axis cs:8000,0.739851871563343)
--(axis cs:6000,0.727597717805986)
--(axis cs:4000,0.711363487468083)
--(axis cs:2000,0.679773454438883)
--(axis cs:0,0.624726922175949)
--cycle;

\path [fill=mediumpurple148103189, fill opacity=0.2]
(axis cs:0,0.646600955857613)
--(axis cs:0,0.636385155253498)
--(axis cs:2000,0.684986785210652)
--(axis cs:4000,0.708398655972818)
--(axis cs:6000,0.722410060418631)
--(axis cs:8000,0.739369252600301)
--(axis cs:10000,0.751137236685442)
--(axis cs:12000,0.761830877870324)
--(axis cs:14000,0.767447412051423)
--(axis cs:16000,0.771245872501391)
--(axis cs:18000,0.774384075152564)
--(axis cs:20000,0.774384075152564)
--(axis cs:20000,0.780303424847436)
--(axis cs:20000,0.780303424847436)
--(axis cs:18000,0.780303424847436)
--(axis cs:16000,0.773458988609721)
--(axis cs:14000,0.771181060170799)
--(axis cs:12000,0.765078844351898)
--(axis cs:10000,0.761015541092336)
--(axis cs:8000,0.755422414066366)
--(axis cs:6000,0.745037856248036)
--(axis cs:4000,0.727799260693849)
--(axis cs:2000,0.700429881456014)
--(axis cs:0,0.646600955857613)
--cycle;

\addplot [semithick, steelblue31119180]
table {%
0 0.686848958333333
2000 0.740885416666667
4000 0.764539930555556
6000 0.784939236111111
8000 0.796440972222222
10000 0.804904513888889
12000 0.813368055555555
14000 0.816189236111111
16000 0.821831597222222
18000 0.823784722222222
20000 0.823784722222222
};
\addlegendentry{No Attack}
\addplot [semithick, darkorange25512714]
table {%
0 0.681206597222222
2000 0.736328125
4000 0.761067708333333
6000 0.779296875
8000 0.794270833333333
10000 0.802951388888889
12000 0.811848958333333
14000 0.81640625
16000 0.821397569444444
18000 0.820095486111111
20000 0.821831597222222
};
\addlegendentry{ALIE}
\addplot [semithick, forestgreen4416044]
table {%
0 0.613715277777778
2000 0.663194444444444
4000 0.680121527777778
6000 0.697699652777778
8000 0.711588541666667
10000 0.723741319444444
12000 0.734592013888889
14000 0.742838541666667
16000 0.750868055555555
18000 0.759548611111111
20000 0.759548611111111
};
\addlegendentry{FOE}
\addplot [semithick, crimson2143940]
table {%
0 0.623914930555555
2000 0.675347222222222
4000 0.699652777777778
6000 0.717230902777778
8000 0.730685763888889
10000 0.737630208333333
12000 0.748914930555555
14000 0.754123263888889
16000 0.760850694444445
18000 0.768012152777778
20000 0.768012152777778
};
\addlegendentry{SF}
\addplot [semithick, mediumpurple148103189]
table {%
0 0.641493055555555
2000 0.692708333333333
4000 0.718098958333333
6000 0.733723958333333
8000 0.747395833333333
10000 0.756076388888889
12000 0.763454861111111
14000 0.769314236111111
16000 0.772352430555556
18000 0.77734375
20000 0.77734375
};
\addlegendentry{TMA}
\end{axis}

\end{tikzpicture}}
    \caption{Accuracy comparison of different attacks on fine-tuning RoBERTa-large on SNLI with \iid data, compared to the baseline.}
    \label{fig:nlp_comparison_iid_eval_acc_SNLI}
\end{figure}

\begin{figure}[H]
    \centering
    \resizebox{.6\linewidth}{!}{\input{data/nlp_comparison_iid_eval_acc_trec}}
    \caption{Accuracy comparison of different attacks on fine-tuning RoBERTa-large on SNLI with \iid data, compared to the baseline.}
    \label{fig:nlp_comparison_iid_eval_acc_trec}
\end{figure}

\begin{figure}[H]
    \centering
    \resizebox{.6\linewidth}{!}{\begin{tikzpicture}

\definecolor{crimson2143940}{RGB}{214,39,40}
\definecolor{darkgray176}{RGB}{176,176,176}
\definecolor{darkorange25512714}{RGB}{255,127,14}
\definecolor{forestgreen4416044}{RGB}{44,160,44}
\definecolor{lightgray204}{RGB}{204,204,204}
\definecolor{mediumpurple148103189}{RGB}{148,103,189}
\definecolor{steelblue31119180}{RGB}{31,119,180}

\begin{axis}[
legend cell align={left},
legend style={
  fill opacity=0.8,
  draw opacity=1,
  text opacity=1,
  at={(0.03,0.97)},
  anchor=north west,
  draw=lightgray204
},
tick align=outside,
tick pos=left,
x grid style={darkgray176},
xlabel={Epoch},
xmajorgrids,
xmin=-1000, xmax=21000,
xtick style={color=black},
y grid style={darkgray176},
ylabel={Accuracy},
ymajorgrids,
ymin=0.782519840948601, ymax=0.925422798337665,
ytick style={color=black}
]
\path [fill=steelblue31119180, fill opacity=0.2]
(axis cs:0,0.869591398538688)
--(axis cs:0,0.857622143127978)
--(axis cs:2000,0.87423611197882)
--(axis cs:4000,0.88667330397788)
--(axis cs:6000,0.888581912957692)
--(axis cs:8000,0.891072009875577)
--(axis cs:10000,0.893025134875577)
--(axis cs:12000,0.897625233528977)
--(axis cs:14000,0.896411693231639)
--(axis cs:16000,0.8980496088192)
--(axis cs:18000,0.9000027338192)
--(axis cs:20000,0.9000027338192)
--(axis cs:20000,0.917054557847467)
--(axis cs:20000,0.917054557847467)
--(axis cs:18000,0.917054557847467)
--(axis cs:16000,0.915101432847467)
--(axis cs:14000,0.915437265101694)
--(axis cs:12000,0.914874766471023)
--(axis cs:10000,0.912964448457756)
--(axis cs:8000,0.911011323457756)
--(axis cs:6000,0.906339962042308)
--(axis cs:4000,0.903040237688787)
--(axis cs:2000,0.895946179687847)
--(axis cs:0,0.869591398538688)
--cycle;

\path [fill=darkorange25512714, fill opacity=0.2]
(axis cs:0,0.87678907275262)
--(axis cs:0,0.856283843914047)
--(axis cs:2000,0.880513548696547)
--(axis cs:4000,0.888497003713581)
--(axis cs:6000,0.893669634536332)
--(axis cs:8000,0.89287355652074)
--(axis cs:10000,0.894329748417773)
--(axis cs:12000,0.898379555466403)
--(axis cs:14000,0.896834973048948)
--(axis cs:16000,0.899218522365837)
--(axis cs:18000,0.901518330633864)
--(axis cs:20000,0.901385290634565)
--(axis cs:20000,0.918927209365435)
--(axis cs:20000,0.918927209365435)
--(axis cs:18000,0.917492086032803)
--(axis cs:16000,0.91783876930083)
--(axis cs:14000,0.913711901951052)
--(axis cs:12000,0.916724611200264)
--(axis cs:10000,0.91556608491556)
--(axis cs:8000,0.910511860145927)
--(axis cs:6000,0.906460573797001)
--(axis cs:4000,0.903820704619753)
--(axis cs:2000,0.890970826303453)
--(axis cs:0,0.87678907275262)
--cycle;

\path [fill=forestgreen4416044, fill opacity=0.2]
(axis cs:0,0.836329037655305)
--(axis cs:0,0.794530337344695)
--(axis cs:2000,0.835575035807218)
--(axis cs:4000,0.857792694611301)
--(axis cs:6000,0.862860102405997)
--(axis cs:8000,0.867767791958901)
--(axis cs:10000,0.872821244597933)
--(axis cs:12000,0.871635450078112)
--(axis cs:14000,0.874578852405997)
--(axis cs:16000,0.878150389126477)
--(axis cs:18000,0.87904592267246)
--(axis cs:20000,0.87904592267246)
--(axis cs:20000,0.905459285660873)
--(axis cs:20000,0.905459285660873)
--(axis cs:18000,0.905459285660873)
--(axis cs:16000,0.900495444206856)
--(axis cs:14000,0.900160730927336)
--(axis cs:12000,0.899848924921888)
--(axis cs:10000,0.8960589637354)
--(axis cs:8000,0.893299916374432)
--(axis cs:6000,0.888441980927336)
--(axis cs:4000,0.881139597055365)
--(axis cs:2000,0.864294755859448)
--(axis cs:0,0.836329037655305)
--cycle;

\path [fill=crimson2143940, fill opacity=0.2]
(axis cs:0,0.838588736745835)
--(axis cs:0,0.789015429920831)
--(axis cs:2000,0.831530643129556)
--(axis cs:4000,0.848089502747945)
--(axis cs:6000,0.857554987344371)
--(axis cs:8000,0.862808521633533)
--(axis cs:10000,0.863306308612308)
--(axis cs:12000,0.867625577108225)
--(axis cs:14000,0.872116894833036)
--(axis cs:16000,0.875316957459745)
--(axis cs:18000,0.879241349120141)
--(axis cs:20000,0.879241349120141)
--(axis cs:20000,0.901357609213192)
--(axis cs:20000,0.901357609213192)
--(axis cs:18000,0.901357609213192)
--(axis cs:16000,0.900073667540255)
--(axis cs:14000,0.899367480166964)
--(axis cs:12000,0.896697339558442)
--(axis cs:10000,0.897110358054358)
--(axis cs:8000,0.891748770033133)
--(axis cs:6000,0.885283554322296)
--(axis cs:4000,0.878472997252055)
--(axis cs:2000,0.863781856870444)
--(axis cs:0,0.838588736745835)
--cycle;

\path [fill=mediumpurple148103189, fill opacity=0.2]
(axis cs:0,0.87217367419281)
--(axis cs:0,0.821836742473857)
--(axis cs:2000,0.860934669606976)
--(axis cs:4000,0.873203376722993)
--(axis cs:6000,0.871688423368302)
--(axis cs:8000,0.874248859150371)
--(axis cs:10000,0.872715327036368)
--(axis cs:12000,0.874234793956118)
--(axis cs:14000,0.874981924802422)
--(axis cs:16000,0.877374109436062)
--(axis cs:18000,0.878239860403809)
--(axis cs:20000,0.878614542766357)
--(axis cs:20000,0.91305212390031)
--(axis cs:20000,0.91305212390031)
--(axis cs:18000,0.912775764596191)
--(axis cs:16000,0.911688390563938)
--(axis cs:14000,0.910174325197578)
--(axis cs:12000,0.909619372710548)
--(axis cs:10000,0.907883631296965)
--(axis cs:8000,0.905048015849629)
--(axis cs:6000,0.903702201631698)
--(axis cs:4000,0.902838289943673)
--(axis cs:2000,0.896877830393024)
--(axis cs:0,0.87217367419281)
--cycle;

\addplot [semithick, steelblue31119180]
table {%
0 0.863606770833333
2000 0.885091145833333
4000 0.894856770833333
6000 0.8974609375
8000 0.901041666666667
10000 0.902994791666667
12000 0.90625
14000 0.905924479166667
16000 0.906575520833333
18000 0.908528645833333
20000 0.908528645833333
};
\addlegendentry{No Attack}
\addplot [semithick, darkorange25512714]
table {%
0 0.866536458333333
2000 0.8857421875
4000 0.896158854166667
6000 0.900065104166667
8000 0.901692708333333
10000 0.904947916666667
12000 0.907552083333333
14000 0.9052734375
16000 0.908528645833333
18000 0.909505208333333
20000 0.91015625
};
\addlegendentry{ALIE}
\addplot [semithick, forestgreen4416044]
table {%
0 0.8154296875
2000 0.849934895833333
4000 0.869466145833333
6000 0.875651041666667
8000 0.880533854166667
10000 0.884440104166667
12000 0.8857421875
14000 0.887369791666667
16000 0.889322916666667
18000 0.892252604166667
20000 0.892252604166667
};
\addlegendentry{FOE}
\addplot [semithick, crimson2143940]
table {%
0 0.813802083333333
2000 0.84765625
4000 0.86328125
6000 0.871419270833333
8000 0.877278645833333
10000 0.880208333333333
12000 0.882161458333333
14000 0.8857421875
16000 0.8876953125
18000 0.890299479166667
20000 0.890299479166667
};
\addlegendentry{SF}
\addplot [semithick, mediumpurple148103189]
table {%
0 0.847005208333333
2000 0.87890625
4000 0.888020833333333
6000 0.8876953125
8000 0.8896484375
10000 0.890299479166667
12000 0.891927083333333
14000 0.892578125
16000 0.89453125
18000 0.8955078125
20000 0.895833333333333
};
\addlegendentry{TMA}
\end{axis}

\end{tikzpicture}}
    \caption{Accuracy comparison of different attacks on fine-tuning RoBERTa-large on SST-2 with \noniid data, compared to the baseline.}
    \label{fig:nlp_comparison_dirichlet1_eval_acc_SST}
\end{figure}

\begin{figure}[H]
    \centering
    \resizebox{.6\linewidth}{!}{\begin{tikzpicture}

\definecolor{crimson2143940}{RGB}{214,39,40}
\definecolor{darkgray176}{RGB}{176,176,176}
\definecolor{darkorange25512714}{RGB}{255,127,14}
\definecolor{forestgreen4416044}{RGB}{44,160,44}
\definecolor{lightgray204}{RGB}{204,204,204}
\definecolor{mediumpurple148103189}{RGB}{148,103,189}
\definecolor{steelblue31119180}{RGB}{31,119,180}

\begin{axis}[
legend cell align={left},
legend style={
  fill opacity=0.8,
  draw opacity=1,
  text opacity=1,
  at={(0.03,0.97)},
  anchor=north west,
  draw=lightgray204
},
tick align=outside,
tick pos=left,
x grid style={darkgray176},
xlabel={Epoch},
xmajorgrids,
xmin=-1000, xmax=21000,
xtick style={color=black},
y grid style={darkgray176},
ylabel={Accuracy},
ymajorgrids,
ymin=0.523012346718598, ymax=0.839210092595889,
ytick style={color=black}
]
\path [fill=steelblue31119180, fill opacity=0.2]
(axis cs:0,0.686365078917097)
--(axis cs:0,0.66563144886068)
--(axis cs:2000,0.728409128132979)
--(axis cs:4000,0.755695389594151)
--(axis cs:6000,0.771617753632779)
--(axis cs:8000,0.78410425921324)
--(axis cs:10000,0.791873448602539)
--(axis cs:12000,0.80015193388861)
--(axis cs:14000,0.803702190730712)
--(axis cs:16000,0.807880125989984)
--(axis cs:18000,0.808843087772271)
--(axis cs:20000,0.808843087772271)
--(axis cs:20000,0.824837467783285)
--(axis cs:20000,0.824837467783285)
--(axis cs:18000,0.824837467783285)
--(axis cs:16000,0.824498346232238)
--(axis cs:14000,0.823901975935955)
--(axis cs:12000,0.819639732778057)
--(axis cs:10000,0.814897384730794)
--(axis cs:8000,0.804003379675649)
--(axis cs:6000,0.794788496367221)
--(axis cs:4000,0.772516415961405)
--(axis cs:2000,0.734264482978132)
--(axis cs:0,0.686365078917097)
--cycle;

\path [fill=darkorange25512714, fill opacity=0.2]
(axis cs:0,0.681239512544614)
--(axis cs:0,0.671191043010942)
--(axis cs:2000,0.720133076028162)
--(axis cs:4000,0.742557973871752)
--(axis cs:6000,0.759299997294088)
--(axis cs:8000,0.775863529625328)
--(axis cs:10000,0.775275684435663)
--(axis cs:12000,0.782755197489312)
--(axis cs:14000,0.780477854432844)
--(axis cs:16000,0.782269150855875)
--(axis cs:18000,0.793777426541776)
--(axis cs:20000,0.794062178516316)
--(axis cs:20000,0.799687821483684)
--(axis cs:20000,0.799687821483684)
--(axis cs:18000,0.797802434569335)
--(axis cs:16000,0.800196126921903)
--(axis cs:14000,0.801987423344934)
--(axis cs:12000,0.799710080288466)
--(axis cs:10000,0.788092371119893)
--(axis cs:8000,0.787938553708005)
--(axis cs:6000,0.78019653048369)
--(axis cs:4000,0.757442026128248)
--(axis cs:2000,0.733425951749615)
--(axis cs:0,0.681239512544614)
--cycle;

\path [fill=forestgreen4416044, fill opacity=0.2]
(axis cs:0,0.652285167357687)
--(axis cs:0,0.537384971531202)
--(axis cs:2000,0.594400587664702)
--(axis cs:4000,0.640480963107002)
--(axis cs:6000,0.670702212200194)
--(axis cs:8000,0.700567667094666)
--(axis cs:10000,0.720375434151771)
--(axis cs:12000,0.733344273555203)
--(axis cs:14000,0.739096525484359)
--(axis cs:16000,0.750503834133533)
--(axis cs:18000,0.758242911381837)
--(axis cs:20000,0.758242911381837)
--(axis cs:20000,0.784291810840385)
--(axis cs:20000,0.784291810840385)
--(axis cs:18000,0.784291810840385)
--(axis cs:16000,0.779444082533133)
--(axis cs:14000,0.773056252293419)
--(axis cs:12000,0.766221698667019)
--(axis cs:10000,0.758357204737117)
--(axis cs:8000,0.747783027349779)
--(axis cs:6000,0.734679732244251)
--(axis cs:4000,0.72236625911522)
--(axis cs:2000,0.69813413455752)
--(axis cs:0,0.652285167357687)
--cycle;

\path [fill=crimson2143940, fill opacity=0.2]
(axis cs:0,0.630154889081469)
--(axis cs:0,0.546494416474087)
--(axis cs:2000,0.582245091306832)
--(axis cs:4000,0.62846619463995)
--(axis cs:6000,0.656225899736563)
--(axis cs:8000,0.682295685901885)
--(axis cs:10000,0.701689680641872)
--(axis cs:12000,0.717687560751017)
--(axis cs:14000,0.724834429198583)
--(axis cs:16000,0.738580943389627)
--(axis cs:18000,0.749835153352109)
--(axis cs:20000,0.749835153352109)
--(axis cs:20000,0.759713457759002)
--(axis cs:20000,0.759713457759002)
--(axis cs:18000,0.759713457759002)
--(axis cs:16000,0.752738501054818)
--(axis cs:14000,0.748255848579195)
--(axis cs:12000,0.738475633693427)
--(axis cs:10000,0.724091569358128)
--(axis cs:8000,0.717443897431448)
--(axis cs:6000,0.703149100263437)
--(axis cs:4000,0.690978249804495)
--(axis cs:2000,0.656470186470946)
--(axis cs:0,0.630154889081469)
--cycle;

\path [fill=mediumpurple148103189, fill opacity=0.2]
(axis cs:0,0.644830445357597)
--(axis cs:0,0.594318860197958)
--(axis cs:2000,0.647774918305508)
--(axis cs:4000,0.695347690322548)
--(axis cs:6000,0.721105552849644)
--(axis cs:8000,0.744333181776865)
--(axis cs:10000,0.760074688172138)
--(axis cs:12000,0.771540438935162)
--(axis cs:14000,0.774761357546495)
--(axis cs:16000,0.780711086203657)
--(axis cs:18000,0.787507722655606)
--(axis cs:20000,0.787507722655606)
--(axis cs:20000,0.790183249566617)
--(axis cs:20000,0.790183249566617)
--(axis cs:18000,0.790183249566617)
--(axis cs:16000,0.783959052685232)
--(axis cs:14000,0.781228225786838)
--(axis cs:12000,0.773164422175949)
--(axis cs:10000,0.765098922938974)
--(axis cs:8000,0.757836957112024)
--(axis cs:6000,0.742002086039246)
--(axis cs:4000,0.726093281899674)
--(axis cs:2000,0.68859660947227)
--(axis cs:0,0.644830445357597)
--cycle;

\addplot [semithick, steelblue31119180]
table {%
0 0.675998263888889
2000 0.731336805555555
4000 0.764105902777778
6000 0.783203125
8000 0.794053819444445
10000 0.803385416666667
12000 0.809895833333333
14000 0.813802083333333
16000 0.816189236111111
18000 0.816840277777778
20000 0.816840277777778
};
\addlegendentry{No Attack}
\addplot [semithick, darkorange25512714]
table {%
0 0.676215277777778
2000 0.726779513888889
4000 0.75
6000 0.769748263888889
8000 0.781901041666667
10000 0.781684027777778
12000 0.791232638888889
14000 0.791232638888889
16000 0.791232638888889
18000 0.795789930555555
20000 0.796875
};
\addlegendentry{ALIE}
\addplot [semithick, forestgreen4416044]
table {%
0 0.594835069444444
2000 0.646267361111111
4000 0.681423611111111
6000 0.702690972222222
8000 0.724175347222222
10000 0.739366319444444
12000 0.749782986111111
14000 0.756076388888889
16000 0.764973958333333
18000 0.771267361111111
20000 0.771267361111111
};
\addlegendentry{FOE}
\addplot [semithick, crimson2143940]
table {%
0 0.588324652777778
2000 0.619357638888889
4000 0.659722222222222
6000 0.6796875
8000 0.699869791666667
10000 0.712890625
12000 0.728081597222222
14000 0.736545138888889
16000 0.745659722222222
18000 0.754774305555555
20000 0.754774305555555
};
\addlegendentry{SF}
\addplot [semithick, mediumpurple148103189]
table {%
0 0.619574652777778
2000 0.668185763888889
4000 0.710720486111111
6000 0.731553819444445
8000 0.751085069444445
10000 0.762586805555556
12000 0.772352430555555
14000 0.777994791666667
16000 0.782335069444445
18000 0.788845486111111
20000 0.788845486111111
};
\addlegendentry{TMA}
\end{axis}

\end{tikzpicture}}
    \caption{Accuracy comparison of different attacks on fine-tuning RoBERTa-large on SNLI with \noniid data, compared to the baseline.}
    \label{fig:nlp_comparison_dirichlet1_eval_acc_SNLI}
\end{figure}

\begin{figure}[H]
    \centering
    \resizebox{.6\linewidth}{!}{\input{data/nlp_comparison_dirichlet1_eval_acc_trec}}
    \caption{Accuracy comparison of different attacks on fine-tuning RoBERTa-large on SNLI with \noniid data, compared to the baseline.}
\end{figure}

\subsection{Hyperparameter Study} \label{sec:hyperparameter_study}
We provide in the following tables \cref{tab:nlp_filtered_results_nworkers_16_stprnd_1,tab:nlp_filtered_results_nworkers_8_nbyzworkers_2_nsmpl_var,tab:nlp_filtered_results_steps_20000_stprnd_1,tab:nlp_filtered_results_nworkers_8_nbyzworkers_2_stprnd_var} a sensitivity analysis of \algname with respect to the number of global epochs $\ngepochs$, the number of local epochs $\nlepochs$, the number of perturbations $\nseeds$, the number of clients $\nclients$, and the number of Byzantine clients $\byzantine$. 

We run our experiments on SST-2 and RoBERTa-large, attacked by FOE. We use \noniid data with $\dirichlet=0.1$. We first show in \cref{tab:nlp_filtered_results_steps_20000_stprnd_1} the robustness of \algname under varying numbers of total and Byzantine clients with $\frac{\nbyz}{\nclients} = 0.25$ and observe that the robustness increases with $\nclients$. We fix $\nclients=8$, and $\nbyz = 2$ as the most challenging setting for the following experiments. %
We show in \cref{tab:nlp_filtered_results_nworkers_8_nbyzworkers_2_nsmpl_var} the stability of \algname under varying number $\nseeds$ of random perturbations. For comparability, we fix the ratio $\nseeds \ngepochs = 20000$ and observe very similar accuracies for $\nseeds \in \{1, 2, 4, 8\}$. Hence, the number of projections trades almost inversely with the number of global epochs $\ngepochs$. A similar behavior can be observed in \cref{tab:nlp_filtered_results_nworkers_8_nbyzworkers_2_stprnd_var} for a varying number of local epochs $\nlepochs$, fixing the ratio $\nlepochs \ngepochs$. However, for large $\nlepochs$ and small $\ngepochs$, we can see the negative impacts of local iterations. Lastly, we evaluate \algname under varying ratios of Byzantine clients, thereby fixing $\nclients=16$ for better flexibility, and using $\nseeds=1$. The results are depicted in \cref{tab:nlp_filtered_results_nworkers_16_stprnd_1}, While the performance reduction from $\nbyz=2$ to $\nbyz=4$ is negligible, we can observe a notable difference for $\nbyz=6$, i.e., when the number of Byzantine clients is close to $\nclients/2$.
\begin{multicols}{2}  %
\begin{table}[H]
\centering
\caption{Accuracy over $\nclients, \nbyz$ for $\nbyz/\nclients = 0.25$}
\begin{tabular}{ccc}
\toprule
Clients \nclients & Byzantine \nbyz & Acc $\pm$ Std \\
\midrule
8 & 2 & 0.86 $\pm$ 0.02 \\
12 & 3 & 0.87 $\pm$ 0.06 \\
16 & 4 & 0.89 $\pm$ 0.02 \\
32 & 8 & 0.87 $\pm$ 0.03 \\
\bottomrule
\end{tabular}
\label{tab:nlp_filtered_results_steps_20000_stprnd_1}
\end{table}

\begin{table}[H]
\centering
\caption{Accuracy over $\ngepochs$ and $\nseeds$.}
\begin{tabular}{ccc}
\toprule
Global epochs $\ngepochs$ & $\nseeds$ & Acc $\pm$ Std \\
\midrule
2500 & 8 & 0.87 $\pm$ 0.02 \\
5000 & 4 & 0.88 $\pm$ 0.02 \\
10000 & 2 & 0.86 $\pm$ 0.02 \\
20000 & 1 & 0.86 $\pm$ 0.02 \\
\bottomrule
\end{tabular}
\label{tab:nlp_filtered_results_nworkers_8_nbyzworkers_2_nsmpl_var}
\end{table}

\begin{table}[H]
\centering
\caption{Accuracy over $\ngepochs$ and $\nlepochs$.}
\begin{tabular}{ccc}
\toprule
Global epochs $\ngepochs$ & Local epochs $\nlepochs$ & Acc $\pm$ Std \\
\midrule
2000 & 10 & 0.81 $\pm$ 0.03 \\
4000 & 5 & 0.86 $\pm$ 0.03 \\
20000 & 1 & 0.86 $\pm$ 0.02 \\
\bottomrule
\end{tabular}
\label{tab:nlp_filtered_results_nworkers_8_nbyzworkers_2_stprnd_var}
\end{table}

\begin{table}[H]
\centering
\caption{Accuracy over $\nbyz/\nclients$ for $\nclients=16$.}
\begin{tabular}{ccc}
\toprule
Clients \nclients & Byzantine \nbyz & Acc $\pm$ Std \\
\midrule
16 & 2 & 0.90 $\pm$ 0.01 \\
16 & 4 & 0.89 $\pm$ 0.02 \\
16 & 6 & 0.78 $\pm$ 0.05 \\
\bottomrule
\end{tabular}
\label{tab:nlp_filtered_results_nworkers_16_stprnd_1}
\end{table}

\end{multicols}  %

\section{Proofs} \label{app:proofs}

\subsection{Proof of \cref{thm:convergence_non_convex}.}

\begin{proof}[Sketch of Proof]
    We provide a brief proof outline in the following for the case when $\pscale>0$. The proof for $\pscale=0$ follows similar steps with some modifications, since $\sqnorm{\left(\nlossi(\modeltl) - \nlossiu(\modeltl)\right)}=0$ by definition and the gradient estimate is bounded differently. We define the following quantities $\modeltlavg \define \frac{1}{\vert \honest \vert} \sumhc \modeltl$ and $\projitlavgh \define \frac{1}{\vert \honest \vert} \sumhc \projitl$, and focus on the conceptual strategy and omit all factors. We first decompose the difference of two consecutive models into $\sqnorm{\nlossh(\modelt)}$, $\sqnorm{\sum_{\lepoch=1}   ^{\nlepochs} \raggtl \perturbtl}$ and $\sqnorm{\nlepochs \nlossh(\modelt)-\sumle \perturbtl \raggtl}$. The former term is the quantity of interest. The second term can be made negative by an appropriate choice of the learning rate. On expectation and using \cref{ass:lipschitz,ass:lipschitz_avg_grad}, the latter can be bounded by  i) $\ec{\hist}{\sqnorm{\modelt-\modeltlavg}}$, ii) $\ec{\histl}{\sumle \sqnorm{\perturbtl\projitlavgh - \perturbtl \raggtl}}$, iii) $\ec{\hist}{\sqnorm{\modeltl-\modeltlavg}}$ and iv) $\ec{\hist}{\sqnorm{\sumle \left(\sumhc \frac{\nlossi(\modeltl)}{\vert\honest\vert} - \perturbtl \projitlavgh\right)}}$. \cref{lemma:model_progress} (in turn requiring similar derivations as for the proof of \cref{lemma:zovar}) relates the term ii) to $\ec{\histl[\lepoch-1]}{\sqnorm{\nlossg(\modelt)}}$ and a term like iii). iv) can be bounded from above by $\sqnorm{\left(\nlossi(\modeltl) - \nlossiu(\modeltl)\right)}$ and $\ec{\hist}{\sqnorm{\sumle \left( \nlossiu(\modeltl) - \perturbtl \projitl \right)}}$. While the first is bounded by \cref{prop:bias}, the latter is bounded by \cref{lemma:zovar} (in turn requiring \cref{ass:bounded_grad_var} and \cref{lem:est_var,lemma:model_progress} in terms of $\ec{\histl[\lepoch-1]}{\sqnorm{\nlossg(\modelt)}}$ and terms like iii). ii) is bounded by a twice application of a particular Johnson-Lindenstrauss-type Lemma (\cref{lemma:modified_jl}) and \cref{lemma:graddiv} (which relies on \citep[Lemma B.1]{wang2025new} and \cref{lemma:zovar}. All terms of the kind iii) are bounded using \cref{lemma:model_divergence} (that relies on \cref{lemma:graddiv}) in terms of $\ec{\histl[\lepoch-1]}{\sqnorm{\nlossg(\modelt)}}$. By appropriate choices of the learning rates so that \cref{lemma:model_progress} and \cref{lemma:model_divergence} hold and the term that multiplies the quantity $\ec{\histl[\lepoch-1]}{\sqnorm{\nlossg(\modelt)}}$ of interest is negative and can hence be rearranged and bounded, the proof is completed by telescoping over all global iterations.    
\end{proof}

\subsubsection{Proof of \cref{thm:convergence_non_convex} for $\pscale>0$}

We start with proving convergence for $\pscale>0$, and apply similar steps to prove the result for $\pscale=0$ in \cref{subsubsec:proof_mu_0}.

\begin{proof}
To prove the convergence of our algorithm for general non-convex functions with local iterations, byzantine resilience and heterogeneity, we rely on the following intermediate lemmas that we state in the following. We assume throughout that \cref{ass:bounded_grad_div,ass:bounded_grad_var,ass:lipschitz_avg_grad,ass:lipschitz} hold, and the robust aggregator satisfies \cref{def:robustness}.

\begin{lemma} \label{lemma:model_progress}
Let
\begin{align*}
    \conste &\define 5\cdot 32 \lr^2 \nlepochs \left(\nlepochs + \frac{\dimension}{\vert \honest \vert \nseeds} \right), \\
    \constf &\define 5\nlepochs \lr^2 \frac{\dimension}{\vert \honest \vert\nseeds} \left(32 \graddiv + 8 \gradvar +  \lipschitz^2 \mu^2 \dimension\right) + 5 \lr^2 24\nlepochs^2 \approxerr,\ \mathrm{and} \\
    \constg &\define 5 \cdot 32\lr^2 \left(\nlepochs \hlipschitz^2 + \frac{1}{\vert \honest \vert} \frac{\dimension}{\nseeds} \lipschitz^2\right).
\end{align*}

Let $\modeltlavg \define \frac{1}{\vert \honest \vert} \sumhc \modeltl$. For a learning rate satisfying $24 \nlepochs \eta^2 \lipschitz^2 \leq \frac{1}{3\nlepochs}$ and $2\eta^2 \frac{1}{\vert \honest \vert} \frac{4\dimension}{\nseeds} 4 \lipschitz^2 \leq \frac{1}{3\nlepochs}$, we have the following upper bound on the averaged local model divergence:
\begin{align*}
    &\ec{\histl[\lepoch-1]}{\sqnorm{\modeltlavg-\modelt}} \leq \conste \ec{\histl[\lepoch-1]}{\sqnorm{\nlossg(\modelt)}} + \constf + \constg \sum_{\lepoch^\prime=1}^{\lepoch-1} \frac{1}{\vert \honest \vert} \sumhc \ec{\histl[\lepoch^\prime-1]}{\sqnorm{\modeltll[\lepoch^\prime] - \modeltmavg[\lepoch^\prime]}}.
\end{align*}
\end{lemma}

\begin{lemma} \label{lemma:zovar}
Let
\begin{align*}
    \consth &\define \frac{16\dimension}{\nseeds} \lipschitz^2, \\
    \consti &\define \frac{80 \cdot 32\dimension}{\nseeds} \lipschitz^2 \lr^2 \left(\nlepochs \hlipschitz^2 + \frac{1}{\vert \honest \vert} \frac{\dimension}{\nseeds} \lipschitz^2\right), \\
    \constj &\define \frac{16\dimension}{\nseeds} + \frac{80\cdot 32\dimension}{\nseeds} \lipschitz^2 \lr^2 \nlepochs \left(\nlepochs + \frac{\dimension}{\vert \honest \vert \nseeds} \right),\ \mathrm{and} \\
    \constk &\define \left(\frac{\dimension}{2\nseeds} + \frac{80\dimension^2}{\nseeds^2 \vert \honest \vert} \lipschitz^2 \nlepochs \lr^2 \right) \left(32 \graddiv + 8 \gradvar +  \lipschitz^2 \mu^2 \dimension\right) + \frac{80 \cdot 24\dimension}{\nseeds} \lipschitz^2 \lr^2 \nlepochs^2 \approxerr.
\end{align*}
We have the following bound on the gradient estimate variance based on multiple independent perturbations
    \begin{align*}
    &\e{\sqnorm{\perturbtl \projitl - \nlossiu(\modeltl)}} \\
    &\leq \consth  \ec{\histl[\lepoch-1]}{\sqnorm{\modeltl - \modeltlavg}} + \constk + \constj \ec{\histl[\lepoch-1]}{\sqnorm{\nlossg(\modelt)}} + \consti \sum_{\lepoch^\prime=1}^{\lepoch-1} \frac{1}{\vert \honest \vert} \sumhc \ec{\histl[\lepoch^\prime-1]}{\sqnorm{\modeltll[\lepoch^\prime] - \modeltmavg[\lepoch^\prime]}}.
\end{align*}
\end{lemma}

\begin{lemma} \label{lemma:graddiv}
Let
\begin{align*}
    \consta[\lepoch] &\define 6 \hlipschitz^2 + 6\graddiv + 2\consth[\lepochtmp] + 2\consti[\lepochtmp] \nlepochs^2 = 6 \hlipschitz^2 + 6\graddiv + 2 \frac{4\dimension}{\nseeds} 4 \lipschitz^2 + \frac{8\dimension}{\nseeds} 4 \lipschitz^2 5 \cdot 32\lr^2 \left(\nlepochs \hlipschitz^2 + \frac{1}{\vert \honest \vert} \frac{\dimension}{\nseeds} \lipschitz^2\right) \nlepochs^2, \\
    \constb &\define 2 \constk[\lepochtmp] = 2 \left(\frac{\dimension}{2\nseeds} + \frac{80\dimension^2}{\nseeds^2 \vert \honest \vert} \lipschitz^2 \nlepochs \lr^2 \right) \left(32 \graddiv + 8 \gradvar +  \lipschitz^2 \mu^2 \dimension\right) + \frac{8\dimension}{\nseeds} 4 \lipschitz^2 5 \lr^2 24\nlepochs^2 \approxerr, \\
    \constc &\define 6 \lipschitz,\ \mathrm{and} \\
    \constd &\define 2 \constj[\lepochtmp] = \frac{8\dimension}{\nseeds} 4 + \frac{8\dimension}{\nseeds} 4 \lipschitz^2 5\cdot 32 \lr^2 \nlepochs \left(\nlepochs + \frac{\dimension}{\vert \honest \vert \nseeds} \right).
\end{align*}
Then the local gradient divergence is bounded from above by
\begin{align*}
    &\sumletmp \frac{1}{\vert \honest \vert} \sumhc \e{\sqnorm{\perturbtm \projitl - \frac{1}{\vert \honest \vert} \sumhcj \perturbtm \projjtm}} \\
    &\leq \sum_{\lepochtmp=1}^{\lepoch} \frac{1}{\vert \honest \vert} \sumhc \consta \e{ \sqnorm{\modeltmavg[\lepochtmp] - \modeltm}} + \sum_{\lepochtmp=1}^\lepoch (\constb + \constc) + \sum_{\lepochtmp=1}^\lepoch \constd \ec{\histl[\lepochtmp-1]}{\sqnorm{\nlossg(\modelt)}}.
\end{align*}
\end{lemma}

\begin{lemma} \label{lemma:model_divergence}
Let
\begin{align*}
    \constl &\define 4 \lr^2 \nlepochs^3 \frac{16\dimension}{\nseeds} + \frac{4\dimension}{\nseeds} 4 \lipschitz^2 5\cdot 32 \lr^2 \nlepochs \left(\nlepochs + \frac{\dimension}{\vert \honest \vert \nseeds} \right)\ \mathrm{and} \\
    \constm &\define 4\lr^2 \nlepochs^3 \left(\left(\frac{\dimension}{2\nseeds} + \frac{80\dimension^2}{\nseeds^2 \vert \honest \vert} \lipschitz^2 \nlepochs \lr^2 \right) \left(32 \graddiv + 8 \gradvar +  \lipschitz^2 \mu^2 \dimension\right) + \frac{80 \cdot 24\dimension}{\nseeds} \lipschitz^2 \lr^2 \nlepochs^2 \approxerr\right) + 12\lr^2 \nlepochs^3 \lipschitz.
\end{align*}
For a learning rate that satisfies $\lr \leq \sqrt{6 \frac{\hlipschitz^2}{\nlepochs^2} + 6\frac{\graddiv}{\nlepochs^2} + \frac{24\dimension}{\nseeds \nlepochs^2} \lipschitz^2}$, %
we have the following upper bound on the local model divergence
\begin{align*}
    \sumle \frac{1}{\vert \honest \vert} \sumhc \e{\sqnorm{\modeltl-\modeltlavg}} &\leq \constl \ec{\histl[\lepochtmp-1]}{\sqnorm{\nlossg(\modelt)}} + \constm.
\end{align*}
\end{lemma}

Now we are ready to prove \cref{thm:convergence_non_convex}. 
With $\raggtl \define \Raggtl$, we have by \cref{ass:lipschitz} that
\begin{align*}
    \lossh(\modelt[\gepoch+1]) - \lossh(\modelt) &\leq \innerprod{\nlossh(\modelt)}{\modelt[\gepoch+1] - \modelt} + \frac{\lipschitz}{2} \sqnorm{\modelt[\gepoch+1] - \modelt} \\
    &= -\lr \innerprod{\nlossh(\modelt)}{\sumle \raggtl \perturbtlr} + \frac{\lr^2 \lipschitz}{2} \sqnorm{\sumle \raggtl \perturbtlr} \numberthis \label{eq:base_equation}
\end{align*}

We first seek an upper bound to the first term:
\begin{align*}
    &-\lr/\nlepochs \innerprod{\nlepochs \nlossh(\modelt)}{\sumle \raggtl \perturbtl} \\
    &=-\lr/(2\nlepochs) \left( \nlepochs^2 \sqnorm{\nlossh(\modelt)} + \sqnorm{\sumle \raggtl \perturbtl} - \sqnorm{\nlepochs \nlossh(\modelt)-\sumle \raggtl \perturbtl} \right).
\end{align*}
The leftmost term is the quantity of interest and will later be brought to the LHS of the equation. The prefactor of the middle term will, by an appropriate choice of the learning rate, be made small enough such that $c \define \frac{\lr^2 \lipschitz}{2}-\frac{\lr}{2\nlepochs} \leq 0$, and hence we can bound the term $c \sqnorm{\sumle \raggtl \perturbtl}$ by $0$. It remains to find a bound for the rightmost term $\sqnorm{\nlepochs \nlossh(\modelt)-\sumle \raggtl \perturbtl}$. Let $\projitlavgh \define \frac{1}{\vert \honest \vert} \sumhc \projitl$. 
By expansion, we can obtain
\begin{align*}
    &\sqnorm{\nlepochs \nlossh(\modelt)-\sumle \perturbtl \raggtl} \\
    &=\sqnormbig{\nlepochs \nlossh(\modelt)+&\sumle \bigg(-\nlossg(\modeltlavg) + \nlossg(\modeltlavg) - \sumhc \frac{\nlossi(\modeltl)}{\vert\honest\vert} \\
    &+ \sumhc \frac{\nlossi(\modeltl)}{\vert\honest\vert} - \perturbtl \projitlavgh + \perturbtl\projitlavgh - \perturbtl \raggtl \bigg)} \\
    &\leq 4 \nlepochs \sumle \sqnorm{\nlossh(\modelt)-\nlossg(\modeltlavg)} + 4 \sqnorm{\sumle \left(\perturbtl\projitlavgh - \perturbtl \raggtl\right)} \\
    &+ 4\nlepochs \sumle \sqnorm{\nlossg(\modeltlavg) - \sumhc \frac{\nlossi(\modeltl)}{\vert\honest\vert}} + 4 \sqnorm{\sumle \left(\sumhc \frac{\nlossi(\modeltl)}{\vert\honest\vert} - \perturbtl \projitlavgh\right)} \\
    &\overset{(a)}{\leq} 4 \nlepochs \sumle \lipschitz^2 \sqnorm{\modelt-\modeltlavg} + 4 \sqnorm{\sumle \left(\perturbtl\projitlavgh - \perturbtl \raggtl\right)} + 4\nlepochs \sumle  \frac{\hlipschitz^2}{\vert \honest \vert} \sumhc \sqnorm{\modeltl-\modeltlavg} \\
    &+ 4 \sqnorm{\sumle \left(\sumhc \frac{\nlossi(\modeltl)}{\vert\honest\vert} - \perturbtl \projitlavgh\right)},
\end{align*}
where $(a)$ holds by \cref{ass:lipschitz} and \cref{ass:lipschitz_avg_grad}.
We take the expectation on both sides and obtain
\begin{align}
    &\ec{\hist}{\sqnorm{\nlepochs \nlossh(\modelt)-\sumle \perturbtl \raggtl}} \nonumber \\
    &\leq 4 \nlepochs \sumle \lipschitz^2 \ec{\hist}{\sqnorm{\modelt-\modeltlavg}} + 4 \ec{\histl}{\sumle \sqnorm{\perturbtl\projitlavgh - \perturbtl \raggtl}} \nonumber \\
    &+ 4\nlepochs \sumle  \frac{\hlipschitz^2}{\vert \honest \vert} \sumhc \ec{\hist}{\sqnorm{\modeltl-\modeltlavg}} + 4 \ec{\hist}{\sqnorm{\sumle \left(\sumhc \frac{\nlossi(\modeltl)}{\vert\honest\vert} - \perturbtl \projitlavgh\right)}}. \label{eq:main_equation}
\end{align}
We continue with bounding the individual terms, and start with the latter term.

\begin{align*}
    &\ec{\hist}{\sqnorm{\sumle \left( \sumhc \frac{1}{\vert\honest\vert} \nlossi(\modeltl) - \perturbtl \projitlavgh \right)}} \\
    &\leq 2\ec{\hist}{\sqnorm{\sumle \sumhc \frac{1}{\vert\honest\vert}\left(\nlossi(\modeltl) - \nlossiu(\modeltl)\right)}} + 2 \ec{\hist}{\sqnorm{\sumle \sumhc \frac{1}{\vert\honest\vert}\left( \nlossiu(\modeltl) - \perturbtl \projitl \right)}} \\
    &\overset{(a)}{\leq} 2 \frac{\nlepochs}{\vert\honest\vert} \sumhc \sumle \sqnorm{\left(\nlossi(\modeltl) - \nlossiu(\modeltl)\right)} + 2\frac{1}{\vert\honest\vert^2} \sumhc \ec{\hist}{\sqnorm{\sumle \left( \nlossiu(\modeltl) - \perturbtl \projitl \right)}} \\
    &\overset{(b)}{\leq} 2 \nlepochs^2 \approxerr + 2\frac{1}{\vert\honest\vert^2} \sumhc \sumle \ec{\randl}{\sqnorm{\nlossiu(\modeltl) - \perturbtl \projitl}} \\
    &\overset{(c)}{\leq} \al{2 \nlepochs^2 \approxerr + 2\frac{1}{\vert\honest\vert^2} \sumhc \sumle &\bigg(\consth  \ec{\histl[\lepoch-1]}{\sqnorm{\modeltl - \modeltlavg}} + \constk \\
    &+ \constj \ec{\histl[\lepoch-1]}{\sqnorm{\nlossg(\modelt)}} + \consti \sum_{\lepoch^\prime=1}^{\lepoch-1} \frac{1}{\vert \honest \vert} \sumhc \ec{\histl[\lepoch^\prime-1]}{\sqnorm{\modeltll[\lepoch^\prime] - \modeltmavg[\lepoch^\prime]}} \bigg)}\\
    &\overset{(d)}{\leq} \al{2 \nlepochs^2 \approxerr + 2\frac{1}{\vert\honest\vert^2} \sumhc \sumle &\bigg(\consth  \ec{\histl[\lepoch-1]}{\sqnorm{\modeltl - \modeltlavg}} + \constk \\
    &+ \constj \ec{\histl[\lepoch-1]}{\sqnorm{\nlossg(\modelt)}} + \frac{\consti \nlepochs (\nlepochs-1)}{\vert \honest \vert} \sumhc \ec{\histl[\lepoch-1]}{\sqnorm{\modeltll[\lepoch] - \modeltmavg[\lepoch]}} \bigg)}\\
    &= \al{2 \nlepochs^2 \approxerr + 2\frac{1}{\vert\honest\vert^2} \sumhc \sumle \bigg((\consth + \consti \nlepochs (\nlepochs-1)) \ec{\histl[\lepoch-1]}{\sqnorm{\modeltl - \modeltlavg}} + \constk + \constj \ec{\histl[\lepoch-1]}{\sqnorm{\nlossg(\modelt)}} \bigg)} \numberthis \label{eq:gradaggvar},
\end{align*}
where $(a)$ is due to the independence of $\sumle \left( \nlossiu(\modeltl) - \perturbtl \projitl \right)$ and $\sumle \left( \nlossju(\modeltlj) - \perturbtl \projjtl \right)$ for $\client \neq \clienttmp$. $(b)$ follows from \cref{prop:estexp} and \citep[Lemma 2]{wang2021novel}. $(c)$ is by the application of \cref{lemma:zovar}. $(d)$ holds since $\sum_{\lepoch=1}^\nlepochs \sum_{\lepoch^\prime=1}^{\lepoch-1} x_{\lepoch^\prime} \leq \sum_{\lepoch=1}^\nlepochs \sum_{\lepoch^\prime=1}^{\lepoch} x_{\lepoch^\prime} \leq \frac{\nlepochs (\nlepochs-1)}{2} \sum_{\lepoch} x_\lepoch$.

We continue with bounding the robustness term using  a double-sided application of an extension of the Johnson-Lindenstrauss Lemma as stated in the following \cref{lemma:modified_jl} and the application of \cref{lemma:graddiv}.
\begin{lemma}[Proposition 8, \citep{li2024simple}] \label{lemma:modified_jl}
    Let $\perturb = (\mathbf{z}_1, \cdots, \mathbf{z}_\nseeds)^T \in \mathbb{R}^{\nseeds \times \dimension}$ with $\mathbf{z}_\seed \sim \uniform[\unitball], \forall \seed \in [\nseeds]$. For a given vector $\mathbf{x} \in \mathbb{R}^{\dimension}$, we have for $\epsilon>0$, $\delta < 1/2$ with probability at least $1-\delta$ that
    \begin{align*}
        (1-\epsilon) \sqnorm{\mathbf{x}} \leq \sqnorm{\perturb \mathbf{x}} \leq (1-\epsilon) \sqnorm{\mathbf{x}}
    \end{align*}
    for $\nseeds \geq 64 \epsilon^{-2} \log(2/\delta)$.
\end{lemma}

For the robustness term, we have
\begin{align*}
    &\e{\sqnorm{\sumle \perturbtl \left(\raggtl - \projitlavgh\right)}} \\
    &\leq \nlepochs \sumle \sqnorm{\perturbtl \left(\raggtl - \projitlavgh\right)} \\
    &\overset{(a)}{\leq} \nlepochs \sumle (1+\epsilon) \frac{\robustness}{\card{\honest}} \sumhc \e{\sqnorm{\projitl -\projitlavgh}} \\
    &\overset{(b)}{\leq} \nlepochs \sumle \frac{(1+\epsilon)}{(1-\epsilon)} \frac{\robustness}{\card{\honest}} \sumhc \e{\sqnorm{\perturbtl (\projitl -\projitlavgh)}} \\
    &\overset{(c)}{\leq} \nlepochs \frac{(1+\epsilon)}{(1-\epsilon)} \robustness \left(\sum_{\lepoch=1}^{\nlepochs} \frac{1}{\vert \honest \vert} \sumhc \consta[\lepoch] \e{ \sqnorm{\modeltmavg[\lepoch] - \modeltl}} + \sum_{\lepoch=1}^\nlepochs (\constb[\lepoch] + \constc[\lepoch]) + \sum_{\lepoch=1}^\nlepochs \constd[\lepoch] \ec{\histl[\lepoch-1]}{\sqnorm{\nlossg(\modelt)}}\right) \numberthis \label{eq:robustnessa} \\
\end{align*}
where $(a)$ and $(b)$ are by the application of \cref{lemma:modified_jl} for in total $\vert \honest \vert+1$ projections. By a union bound over \cref{lemma:modified_jl}, the distance preservation holds with probability $1-\delta$ for $\epsilon \geq \sqrt{\frac{64}{\nseeds} \log(\frac{2(\vert \honest \vert-1)}{ \delta})}$. We choose the smallest possible $\epsilon$. This must hold for each iteration, so the distance preservation holds w.p. at least $1-\nlepochs \delta$ for all local epochs. $(c)$ is by \cref{lemma:graddiv}.

To bound the local model divergence from the global model, by \cref{lemma:model_progress}, we have
\begin{align*}
    \sumle \ec{\histl[\lepoch-1]}{\sqnorm{\modeltlavg-\modelt}} &\leq \sumle \conste \ec{\histl[\lepoch-1]}{\sqnorm{\nlossg(\modelt)}} + \sumle \constf + \sumle \constg \sum_{\lepoch^\prime=1}^{\lepoch-1} \frac{1}{\vert \honest \vert} \sumhc \ec{\histl[\lepoch^\prime-1]}{\sqnorm{\modeltll[\lepoch^\prime] - \modeltmavg[\lepoch^\prime]}} \\
    &\leq \sumle \conste \ec{\histl[\lepoch-1]}{\sqnorm{\nlossg(\modelt)}} + \sumle \constf + \sumle \frac{\constg \nlepochs (\nlepochs-1)}{\vert \honest \vert} \sumhc \ec{\histl[\lepoch-1]}{\sqnorm{\modeltll[\lepoch] - \modeltmavg[\lepoch]}} \numberthis \label{eq:model_progressa},
\end{align*}
where the latter follows since $\sum_{\lepoch=1}^\nlepochs \sum_{\lepoch^\prime=1}^{\lepoch-1} x_{\lepoch^\prime} \leq \sum_{\lepoch=1}^\nlepochs \sum_{\lepoch^\prime=1}^{\lepoch} x_{\lepoch^\prime} \leq \frac{\nlepochs (\nlepochs-1)}{2} \sum_{\lepoch} x_\lepoch$.

Plugging \eqref{eq:gradaggvar}, \eqref{eq:robustnessa}, and \eqref{eq:model_progressa} into \eqref{eq:main_equation}, we obtain
\begin{align*}
    &\ec{\hist}{\sqnorm{\nlepochs \nlossh(\modelt)-\sumle \perturbtl \raggtl}} \nonumber \\
    &\leq 4 \nlepochs \lipschitz^2 \left(\sumle \conste \ec{\histl[\lepoch-1]}{\sqnorm{\nlossg(\modelt)}} + \sumle \constf + \sumle \frac{\constg \nlepochs (\nlepochs-1)}{\vert \honest \vert} \sumhc \ec{\histl[\lepoch-1]}{\sqnorm{\modeltll[\lepoch] - \modeltmavg[\lepoch]}}\right) \\
    &+ 4 \left(\nlepochs \frac{(1+\epsilon)}{(1-\epsilon)} \robustness \left(\sum_{\lepoch=1}^{\nlepochs} \frac{1}{\vert \honest \vert} \sumhc \consta[\lepoch] \e{ \sqnorm{\modeltmavg[\lepoch] - \modeltl}} + \sum_{\lepoch=1}^\nlepochs (\constb[\lepoch] + \constc[\lepoch]) + \sum_{\lepoch=1}^\nlepochs \constd[\lepoch] \ec{\histl[\lepoch-1]}{\sqnorm{\nlossg(\modelt)}}\right)\right) \nonumber \\
    &+ 4\nlepochs \sumle  \frac{\hlipschitz^2}{\vert \honest \vert} \sumhc \ec{\hist}{\sqnorm{\modeltl-\modeltlavg}} \\
    &+ 4 \left(2 \nlepochs^2 \approxerr + 2\frac{1}{\vert\honest\vert^2} \sumhc \sumle \bigg((\consth + \consti \nlepochs (\nlepochs-1)) \ec{\histl[\lepoch-1]}{\sqnorm{\modeltl - \modeltlavg}} + \constk + \constj \ec{\histl[\lepoch-1]}{\sqnorm{\nlossg(\modelt)}} \bigg)\right) \\
    &\leq \constnprime \ec{\histl[\lepoch-1]}{\sqnorm{\nlossg(\modelt)}} + \constoprime \sum_{\lepoch=1}^{\nlepochs} \frac{1}{\vert \honest \vert} \sumhc \e{ \sqnorm{\modeltmavg[\lepoch] - \modeltl}} + \constpprime \\
    &\leq \constnprime \ec{\histl[\lepoch-1]}{\sqnorm{\nlossg(\modelt)}} + \constoprime \constl \ec{\histl[\lepochtmp-1]}{\sqnorm{\nlossg(\modelt)}} + \constoprime \constm + \constpprime, \\
    &\leq (\constnprime + \constoprime \constl) \ec{\histl[\lepoch-1]}{\sqnorm{\nlossg(\modelt)}} + \constoprime \constm + \constpprime, \numberthis \label{eq:intermediate_bound}
\end{align*}
where the penultimate step is by the application of \cref{lemma:model_divergence}, and the constants read and can be bounded as
\begin{align*}
    \constnprime &\define 4 \nlepochs \lipschitz^2 \sumle \conste + 4 \nlepochs \frac{(1+\epsilon)}{(1-\epsilon)} \robustness \sum_{\lepoch=1}^\nlepochs \constd[\lepoch] + 4 \cdot 2\frac{1}{\vert\honest\vert^2} \sumhc \sumle \constj \\
    &\leq 4 \nlepochs^2 \lipschitz^2 \conste + \left(8 \nlepochs^2 \frac{(1+\epsilon)}{(1-\epsilon)} \robustness + 8\frac{1}{\vert\honest\vert} \nlepochs\right) \constj \\
    &\leq \constn \define 4 \nlepochs^2 \lipschitz^2 5\cdot 32 \lr^2 \nlepochs \left(\nlepochs + \frac{\dimension}{\vert \honest \vert \nseeds} \right) + \left(8 \nlepochs^2 \frac{(1+\epsilon)}{(1-\epsilon)} \robustness + 8\frac{1}{\vert\honest\vert} \nlepochs\right) \left(\frac{16\dimension}{\nseeds} + \frac{80\cdot 32\dimension}{\nseeds} \lipschitz^2 \lr^2 \nlepochs \left(\nlepochs + \frac{\dimension}{\vert \honest \vert \nseeds} \right)\right) \\
    \constoprime &\define 4 \nlepochs \lipschitz^2 \constg \nlepochs (\nlepochs-1) + 4 \nlepochs \frac{(1+\epsilon)}{(1-\epsilon)} \robustness \consta[\nlepochs] + 4\nlepochs \hlipschitz^2 + 4 \cdot 2\frac{1}{\vert\honest\vert} (\consth + \consti \nlepochs (\nlepochs-1)) \\
    &\leq 4 \nlepochs^3 \lipschitz^2 \constg + 4 \nlepochs \frac{(1+\epsilon)}{(1-\epsilon)} \robustness \consta[\nlepochs] + 4\nlepochs \hlipschitz^2 + 8\frac{1}{\vert\honest\vert} (\consth + \consti \nlepochs^2) \\
    &\leq 4 \nlepochs^3 \lipschitz^2 \constg + 8 \nlepochs \frac{(1+\epsilon)}{(1-\epsilon)} \robustness \left(3 \hlipschitz^2 + 3\graddiv + \consth[\lepochtmp] + \consti[\lepochtmp] \nlepochs^2\right) + 4\nlepochs \hlipschitz^2 + 8\frac{1}{\vert\honest\vert} (\consth + \consti \nlepochs^2) \\
    &\leq 4 \nlepochs^3 \lipschitz^2 \constg + 8 \nlepochs \frac{(1+\epsilon)}{(1-\epsilon)} \robustness \left(3 \hlipschitz^2 + 3\graddiv\right) + 4\nlepochs \hlipschitz^2 + 8\left(\frac{1}{\vert\honest\vert}+\nlepochs \frac{(1+\epsilon)}{(1-\epsilon)} \robustness\right) (\consth + \consti \nlepochs^2) \\
    &\leq 20 \cdot 32 \nlepochs^3 \lipschitz^2 \lr^2 \left(\nlepochs \hlipschitz^2 + \frac{1}{\vert \honest \vert} \frac{\dimension}{\nseeds} \lipschitz^2\right) + 8 \nlepochs \frac{(1+\epsilon)}{(1-\epsilon)} \robustness \left(3 \hlipschitz^2 + 3\graddiv\right) + 4\nlepochs \hlipschitz^2 \\
    &+ 8\left(\frac{1}{\vert\honest\vert}+\nlepochs \frac{(1+\epsilon)}{(1-\epsilon)} \robustness\right) \left(\frac{16\dimension}{\nseeds} \lipschitz^2 + \frac{80 \cdot 32\dimension}{\nseeds} \lipschitz^2 \lr^2 \left(\nlepochs \hlipschitz^2 + \frac{1}{\vert \honest \vert} \frac{\dimension}{\nseeds} \lipschitz^2\right) \nlepochs^2\right) \\
    &\leq \consto \define \al{&20 \cdot 32 \nlepochs^2 \lipschitz^2 \lr^2 \left(\nlepochs \hlipschitz^2 + \frac{1}{\vert \honest \vert} \frac{\dimension}{\nseeds} \lipschitz^2\right) \left(\nlepochs + \frac{4\dimension}{\nseeds}\right) \\
    &+ 8 \nlepochs \frac{(1+\epsilon)}{(1-\epsilon)} \robustness \left(3 \hlipschitz^2 + 3\graddiv + \frac{16\dimension}{\nseeds} \lipschitz^2\right) + 4\nlepochs \hlipschitz^2 + 8 \frac{1}{\vert\honest\vert} \frac{16\dimension}{\nseeds} \lipschitz^2 } \\
    \constpprime &\define 4 \nlepochs \lipschitz^2 \sumle \constf + 4 \nlepochs \frac{(1+\epsilon)}{(1-\epsilon)} \robustness \sum_{\lepoch=1}^\nlepochs (\constb[\lepoch] + \constc[\lepoch]) + 4 \cdot 2 \nlepochs^2 \approxerr + 4 \cdot 2\frac{1}{\vert\honest\vert^2} \sumhc \sumle \constk \\
    &\leq 4 \nlepochs^2 \lipschitz^2 \constf + 4 \nlepochs^2 \frac{(1+\epsilon)}{(1-\epsilon)} \robustness (\constb[\lepoch] + \constc[\lepoch]) + 8 \nlepochs^2 \approxerr + 8\frac{1}{\vert\honest\vert} \nlepochs \constk \\
    &\leq 4 \nlepochs^2 \lipschitz^2 \constf + 4 \nlepochs^2 \frac{(1+\epsilon)}{(1-\epsilon)} \robustness (2 \constk[\lepochtmp] + 6\lipschitz) + 8 \nlepochs^2 \approxerr + 8\frac{1}{\vert\honest\vert} \nlepochs \constk \\
    &\leq 4 \nlepochs^2 \lipschitz^2 \constf + 4 \nlepochs^2 \frac{(1+\epsilon)}{(1-\epsilon)} \robustness 6\lipschitz + 8 \nlepochs^2 \approxerr + \left(8\frac{1}{\vert\honest\vert} \nlepochs + 8 \nlepochs^2 \frac{(1+\epsilon)}{(1-\epsilon)} \robustness \right) \constk \\
    &\leq 4 \nlepochs^2 \lipschitz^2 \left(5\nlepochs \lr^2 \frac{\dimension}{\vert \honest \vert\nseeds} \left(32 \graddiv + 8 \gradvar +  \lipschitz^2 \mu^2 \dimension\right) + 5 \lr^2 24\nlepochs^2 \approxerr\right) + 4 \nlepochs^2 \frac{(1+\epsilon)}{(1-\epsilon)} \robustness 6\lipschitz + 8 \nlepochs^2 \approxerr \\
    &+ \left(8\frac{1}{\vert\honest\vert} \nlepochs + 8 \nlepochs^2 \frac{(1+\epsilon)}{(1-\epsilon)} \robustness \right) \left(\left(\frac{\dimension}{2\nseeds} + \frac{80\dimension^2}{\nseeds^2 \vert \honest \vert} \lipschitz^2 \nlepochs \lr^2 \right) \left(32 \graddiv + 8 \gradvar +  \lipschitz^2 \mu^2 \dimension\right) + \frac{80 \cdot 24\dimension}{\nseeds} \lipschitz^2 \lr^2 \nlepochs^2 \approxerr\right) \\
    &\leq \constp \define \al{&\left(32 \graddiv + 8 \gradvar +  \lipschitz^2 \mu^2 \dimension\right) \left(20 \nlepochs^3 \lipschitz^2 \lr^2 \frac{\dimension}{\vert \honest \vert\nseeds} + \left(8\frac{1}{\vert\honest\vert} \nlepochs + 8 \nlepochs^2 \frac{(1+\epsilon)}{(1-\epsilon)} \robustness \right) \left(\frac{\dimension}{2\nseeds} + \frac{80\dimension^2}{\nseeds^2 \vert \honest \vert} \lipschitz^2 \nlepochs \lr^2 \right) \right) \\
    &+ 4 \nlepochs^2 \lipschitz^2 5 \lr^2 24\nlepochs^2 \approxerr + 4 \nlepochs^2 \frac{(1+\epsilon)}{(1-\epsilon)} \robustness 6\lipschitz + 8 \nlepochs^2 \approxerr + \left(8\frac{1}{\vert\honest\vert} \nlepochs + 8 \nlepochs^2 \frac{(1+\epsilon)}{(1-\epsilon)} \robustness \right) \left(\frac{80 \cdot 24\dimension}{\nseeds} \lipschitz^2 \lr^2 \nlepochs^2 \approxerr\right)}.
\end{align*}

By taking the expectation over \eqref{eq:base_equation} and replacing $\e{\sqnorm{\nlepochs \nlossh(\modelt)-\sumle \perturbtl \raggtl}}$ by \eqref{eq:intermediate_bound}, we can write
\begin{align*}
    &\e{\lossh(\modelt[\gepoch+1])} - \e{\lossh(\modelt)} \\
    &\leq -\lr/(2) \nlepochs \e{\sqnorm{\nlossh(\modelt)}} + \left(\frac{\lr^2 \lipschitz}{2}-\frac{\lr}{2\nlepochs}\right) \e{\sqnorm{\sumle \raggtl \perturbtl}} + \lr/(2\nlepochs) \e{\sqnorm{\nlepochs \nlossh(\modelt)-\sumle \raggtl \perturbtl}} \\
    &\overset{(a)}{\leq} -\lr/(2) \nlepochs \e{\sqnorm{\nlossh(\modelt)}} +  \lr/(2\nlepochs) \left((\constn + \consto \constl) \ec{\histl[\lepoch-1]}{\sqnorm{\nlossg(\modelt)}} + \consto \constm + \constp \right) \\
    &\overset{(b)}{\leq} -\frac{\lr \nlepochs}{4} \e{\sqnorm{\nlossh(\modelt)}} +  \lr/(2\nlepochs) \left(\consto \constm + \constp \right),
\end{align*}
where $(a)$ holds when $\lr \leq \frac{1}{\nlepochs \lipschitz}$ and $(b)$ assumes that $\frac{\lr}{2\nlepochs} (\constn + \consto \constl) \leq \frac{\lr \nlepochs}{4}$.

Reordering and telescoping over $\gepoch$, we obtain
\begin{align*}
    \frac{1}{\ngepochs} \sum_{\gepoch=1}^{\ngepochs} \e{\sqnorm{\nlossh(\modelt)}} \leq \frac{4(\e{\lossh(\modelt[1])} - \e{\lossh(\modelt[\ngepochs+1])})}{\ngepochs \lr \nlepochs} + \lr/(2\nlepochs) \left(\consto \constm + \constp \right) \\
\end{align*}
with probability $1-\delta \nlepochs \ngepochs$ by a union bound argument over all global iterations $\ngepochs$. We let $\probguarantee \define \delta \nlepochs \ngepochs$, and obtain $\epsilon \geq \sqrt{\frac{64}{\nseeds} \log(\frac{2(\vert \honest \vert-1)}{ \delta})} = \sqrt{\frac{64}{\nseeds} \log(\frac{2(\vert \honest \vert-1) \ngepochs \nlepochs}{\probguarantee})}$. Since it is required to satisfy $\epsilon < 1$, the proof holds for $\nseeds \geq 64 \log(\frac{2(\vert \honest \vert-1) \ngepochs \nlepochs}{\probguarantee})$. 
Noting that $\lossh(\modelt[\ngepochs+1]) \geq \optlossh$ by definition concludes the proof. The requirements on the learning rate are summarized as follows:
\begin{itemize}
    \item $24 \nlepochs \eta^2 \lipschitz^2 \leq \frac{1}{3\nlepochs} \rightarrow \lr \leq \frac{1}{\sqrt{72 \nlepochs^2 \lipschitz}}$
    \item $2\eta^2 \frac{1}{\vert \honest \vert} \frac{4\dimension}{\nseeds} 4 \lipschitz^2 \leq \frac{1}{3\nlepochs} \rightarrow \lr \leq \sqrt{\frac{\vert \honest \vert \nseeds}{96\nlepochs \dimension \lipschitz^2}}$
    \item $\lr \leq \sqrt{6 \frac{\hlipschitz^2}{\nlepochs^2} + 6\frac{\graddiv}{\nlepochs^2} + \frac{32\dimension}{\nseeds \nlepochs^2} \lipschitz^2}$
\end{itemize}

The constants are summarized as
\begin{align*}
    \constl &\define 4 \lr^2 \nlepochs^3 \frac{16}{\nseeds} + \frac{4\dimension}{\nseeds} 4 \lipschitz^2 5\cdot 32 \lr^2 \nlepochs \left(\nlepochs + \frac{\dimension}{\vert \honest \vert \nseeds} \right) \\
    \constm &\define 4\lr^2 \nlepochs^3 \left(\left(\frac{\dimension}{2\nseeds} + \frac{80\dimension^2}{\nseeds^2 \vert \honest \vert} \lipschitz^2 \nlepochs \lr^2 \right) \left(32 \graddiv + 8 \gradvar +  \lipschitz^2 \mu^2 \dimension\right) + \frac{80 \cdot 24\dimension}{\nseeds} \lipschitz^2 \lr^2 \nlepochs^2 \approxerr\right) + 12\lr^2 \nlepochs^3 \lipschitz \\
    \constn &\define 4 \nlepochs^2 \lipschitz^2 5\cdot 32 \lr^2 \nlepochs \left(\nlepochs + \frac{\dimension}{\vert \honest \vert \nseeds} \right) + \left(8 \nlepochs^2 \frac{(1+\epsilon)}{(1-\epsilon)} \robustness + 8\frac{1}{\vert\honest\vert} \nlepochs\right) \left(\frac{16\dimension}{\nseeds} + \frac{80\cdot 32\dimension}{\nseeds} \lipschitz^2 \lr^2 \nlepochs \left(\nlepochs + \frac{\dimension}{\vert \honest \vert \nseeds} \right)\right) \\
    \consto &\define 20 \cdot 32 \nlepochs^2 \lipschitz^2 \lr^2 \left(\nlepochs \hlipschitz^2 + \frac{1}{\vert \honest \vert} \frac{\dimension}{\nseeds} \lipschitz^2\right) \left(\nlepochs + \frac{4\dimension}{\nseeds}\right) + 8 \nlepochs \frac{(1+\epsilon)}{(1-\epsilon)} \robustness \left(3 \hlipschitz^2 + 3\graddiv + \frac{16\dimension}{\nseeds} \lipschitz^2\right) + 4\nlepochs \hlipschitz^2 + 8 \frac{1}{\vert\honest\vert} \frac{16\dimension}{\nseeds} \lipschitz^2  \\
    \constp &\define \left(32 \graddiv + 8 \gradvar +  \lipschitz^2 \mu^2 \dimension\right) \left(20 \nlepochs^3 \lipschitz^2 \lr^2 \frac{\dimension}{\vert \honest \vert\nseeds} + \left(8\frac{1}{\vert\honest\vert} \nlepochs + 8 \nlepochs^2 \frac{(1+\epsilon)}{(1-\epsilon)} \robustness \right) \left(\frac{\dimension}{2\nseeds} + \frac{80\dimension^2}{\nseeds^2 \vert \honest \vert} \lipschitz^2 \nlepochs \lr^2 \right) \right) + 4 \nlepochs^2 \lipschitz^2 5 \lr^2 24\nlepochs^2 \approxerr \\
    &+ 4 \nlepochs^2 \frac{(1+\epsilon)}{(1-\epsilon)} \robustness 6\lipschitz + 8 \nlepochs^2 \approxerr + \left(8\frac{1}{\vert\honest\vert} \nlepochs + 8 \nlepochs^2 \frac{(1+\epsilon)}{(1-\epsilon)} \robustness \right) \left(\frac{80 \cdot 24\dimension}{\nseeds} \lipschitz^2 \lr^2 \nlepochs^2 \approxerr\right),
\end{align*}
and, with $\efrac \define \frac{(1+\epsilon)}{(1-\epsilon)}$, can be approximated by
\begin{align}
    \constl &= \Theta\left(\lr^2 \nlepochs^3 \frac{\dimension}{\nseeds} + \frac{\dimension}{\nseeds} \lipschitz^2 \lr^2 \nlepochs \left(\nlepochs + \frac{\dimension}{\vert \honest \vert \nseeds} \right)\right) \label{eq:constl} \\
    \constm &= \Theta\left(\lr^2 \nlepochs^3 \left(\lipschitz+\left(\frac{\dimension}{\nseeds} + \frac{\dimension^2}{\nseeds^2 \vert \honest \vert} \lipschitz^2 \nlepochs^2 \lr^2 \right) \left(\graddiv +  \gradvar +  \lipschitz^2 \mu \dimension\right)\right) \right) \label{eq:constm} \\
    \constn &= \Theta \left(\frac{\dimension}{\nseeds} \left(\nlepochs^2 \efrac \robustness + \frac{\nlepochs}{\vert\honest\vert}\right) \left(1 + \lipschitz^2 \lr^2 \nlepochs \left(\nlepochs + \frac{\dimension}{\vert \honest \vert \nseeds} \right) (1+\frac{\nseeds}{\dimension}) \right)\right) \label{eq:constn} \\
    \consto &= \Theta \left(\nlepochs^2 \lipschitz^2 \lr^2 \left(\nlepochs^2 \hlipschitz^2 + \frac{1}{\vert \honest \vert} \frac{\dimension^2}{\nseeds^2} \lipschitz^2\right) + \nlepochs \efrac \robustness \left(\hlipschitz^2 + \graddiv + \frac{\dimension}{\nseeds} \lipschitz^2\right)\right) \label{eq:consto} \\
    \constp &= \Theta\bigg(\left(\graddiv + \gradvar +  \lipschitz^2 \mu^2 \dimension\right) \left(\left(\nlepochs^2 \efrac \robustness \right) \left(\frac{\dimension}{\nseeds} + \frac{\dimension^2}{\nseeds^2 \vert \honest \vert} \lipschitz^2 \nlepochs \lr^2 \right) \right) \\
    &+ \nlepochs^2\lipschitz( \efrac \robustness  + \pscale) + \left(\nlepochs^2 \efrac \robustness \right) \left(\frac{\dimension}{\nseeds} \lipschitz^2 \lr^2 \nlepochs^2 \approxerr\right)\bigg). \label{eq:constp}
\end{align}

\end{proof}

We now continue to prove all intermediate \cref{lemma:graddiv,lemma:model_progress,lemma:model_divergence,lemma:zovar}.

\begin{proof}[Proof of \cref{lemma:model_progress}]
By definition, $\ec{\histl[\lepoch-1]}{\sqnorm{\modeltmavg[1]-\modelt}}=0$. For $\lepoch \in \{2, \cdots, \nlepochs\}$, we have
\begin{align*}
    &\ec{\histl[\lepoch-1]}{\sqnorm{\modeltlavg-\modelt}} \\
    &= \ec{\histl[\lepoch-1]}{\sqnorm{\modeltmavg[\lepoch-1]- \frac{\lr}{\vert \honest \vert} \sumhc \perturbtl[\lepoch-1] \projitll[\lepoch-1] - \modelt}} \\
    &= \ecbig{\histl[\lepoch-1]}{\sqnormbignal{&\modeltmavg[\lepoch-1] - \modelt -\lr \bigg(\frac{1}{\vert \honest \vert} \sumhc \perturbtl[\lepoch-1] \projitll[\lepoch-1] - \frac{1}{\vert \honest \vert} \sumhc \nlossiu(\modeltll[\lepoch-1]) + \frac{1}{\vert \honest \vert} \sumhc \nlossiu(\modeltll[\lepoch-1]) \\
    &- \frac{1}{\vert \honest \vert} \sumhc \nlossi(\modeltll[\lepoch-1]) + \frac{1}{\vert \honest \vert} \sumhc \nlossi(\modeltll[\lepoch-1]) - \nlossg(\modeltmavg[\lepoch-1]) + \nlossg(\modeltmavg[\lepoch-1]) - \nlossg(\modelt) + \nlossg(\modelt) \bigg)}} \\
    &\leq (1+\frac{1}{\tradetmp}) \ec{\histl[\lepoch-1]}{\sqnorm{\modeltmavg[\lepoch-1] - \modelt}} + \\
    &(1+\tradetmp) \ecbig{\histl[\lepoch-1]}{\sqnormbignal{&-\lr \bigg(\frac{1}{\vert \honest \vert} \sumhc \perturbtl[\lepoch-1] \projitll[\lepoch-1] - \frac{1}{\vert \honest \vert} \sumhc \nlossiu(\modeltll[\lepoch-1]) + \frac{1}{\vert \honest \vert} \sumhc \nlossiu(\modeltll[\lepoch-1]) \\
    &- \frac{1}{\vert \honest \vert} \sumhc \nlossi(\modeltll[\lepoch-1]) + \frac{1}{\vert \honest \vert} \sumhc \nlossi(\modeltll[\lepoch-1]) - \nlossg(\modeltmavg[\lepoch-1]) + \nlossg(\modeltmavg[\lepoch-1]) - \nlossg(\modelt) + \nlossg(\modelt) \bigg)}} \\
    &\overset{(b)}{\leq} \al{&(1+\frac{1}{\tradetmp}) \ec{\histl[\lepoch-1]}{\sqnorm{\modeltmavg[\lepoch-1] - \modelt}} \\
    &+ 2(1+\tradetmp) \eta^2 \ecbig{\histl[\lepoch-1]}{\sqnormbignal{&\frac{1}{\vert \honest \vert} \sumhc \nlossiu(\modeltll[\lepoch-1]) - \frac{1}{\vert \honest \vert} \sumhc \nlossi(\modeltll[\lepoch-1]) + \frac{1}{\vert \honest \vert} \sumhc \nlossi(\modeltll[\lepoch-1]) - \nlossg(\modeltmavg[\lepoch-1]) \\
    &+ \nlossg(\modeltmavg[\lepoch-1]) - \nlossg(\modelt) + \nlossg(\modelt)}}
    \\
    &+ 2\eta^2 \frac{1}{\vert \honest \vert^2} \sumhc \ec{\histl[\lepoch-1]}{\sqnorm{\perturbtl[\lepoch-1] \projitll[\lepoch-1] - \nlossiu(\modeltll[\lepoch-1])}}} \\
    &\leq \al{&(1+\frac{1}{\tradetmp}) \ec{\histl[\lepoch-1]}{\sqnorm{\modeltmavg[\lepoch-1] - \modelt}} + 8(1+\tradetmp) \eta^2 \ec{\histl[\lepoch-1]}{\sqnorm{\frac{1}{\vert \honest \vert} \sumhc \nlossiu(\modeltll[\lepoch-1]) - \frac{1}{\vert \honest \vert} \sumhc \nlossi(\modeltll[\lepoch-1])}}
    \\
    &+ 8(1+\tradetmp) \eta^2 \ec{\histl[\lepoch-1]}{\sqnorm{\nlossg(\modelt)}} \\
    &+ 8(1+\tradetmp) \eta^2 \ec{\histl[\lepoch-1]}{\sqnorm{\frac{1}{\vert \honest \vert} \sumhc \nlossi(\modeltll[\lepoch-1]) - \nlossg(\modeltmavg[\lepoch-1])}}
    + 8(1+\tradetmp) \eta^2 \ec{\histl[\lepoch-1]}{\sqnorm{\nlossg(\modeltmavg[\lepoch-1]) - \nlossg(\modelt)}}
    \\
    &+ 2\eta^2 \frac{1}{\vert \honest \vert^2} \sumhc \ec{\histl[\lepoch-1]}{\sqnorm{\perturbtl[\lepoch-1] \projitll[\lepoch-1] - \nlossiu(\modeltll[\lepoch-1])}}} \numberthis \label{eq:model_progress_lipschitz} \\
    &\overset{(c)}{\leq} \al{&(1+\frac{1}{\tradetmp}) \ec{\histl[\lepoch-1]}{\sqnorm{\modeltmavg[\lepoch-1] - \modelt}} + 8(1+\tradetmp) \eta^2 \approxerr 
    + 8(1+\tradetmp) \eta^2 \ec{\histl[\lepoch-1]}{\sqnorm{\nlossg(\modelt)}} \\
    &+ 8(1+\tradetmp) \eta^2 \frac{\hlipschitz^2}{\vert \honest \vert} \sumhc \ec{\histl[\lepoch-1]}{\sqnorm{\modeltll[\lepoch-1] - \modeltmavg[\lepoch-1]}}
    + 8(1+\tradetmp) \eta^2 \lipschitz^2 \ec{\histl[\lepoch-1]}{\sqnorm{\modeltmavg[\lepoch-1] - \modelt}}
    \\
    &+ 2\eta^2 \frac{1}{\vert \honest \vert^2} \sumhc \frac{4\dimension}{\nseeds} \left(4 \lipschitz^2 \ec{\histl[\lepoch-1]}{\sqnorm{\modeltll[\lepoch-1] - \modeltmavg[\lepoch-1]}} + 4 \graddiv + 4 \lipschitz^2 \ec{\histl[\lepoch-1]}{\sqnorm{\modeltmavg[\lepoch-1]-\modelt}} + 4\ec{\histl[\lepoch-1]}{\sqnorm{\nlossg(\modelt)}}\right) \\
    &+ 2\eta^2 \frac{1}{\vert \honest \vert} \frac{4\dimension}{\nseeds} \gradvar +  2\eta^2 \frac{1}{\vert \honest \vert} \frac{\lipschitz^2 \mu^2 \dimension^2}{2\nseeds}} \\
    &\overset{(d)}{\leq} \al{&\left((1+\frac{1}{\tradetmp}) + 8(1+\tradetmp) \eta^2 \lipschitz^2 + 2\eta^2 \frac{1}{\vert \honest \vert} \frac{4\dimension}{\nseeds} 4 \lipschitz^2 \right) \ec{\histl[\lepoch-1]}{\sqnorm{\modeltmavg[\lepoch-1] - \modelt}} + 8(1+\tradetmp) \eta^2 \approxerr 
    \\
    &+ \left(8(1+\tradetmp) \eta^2 + 2\eta^2 \frac{1}{\vert \honest \vert} \frac{4\dimension}{\nseeds} 4\right) \ec{\histl[\lepoch-1]}{\sqnorm{\nlossg(\modelt)}} \\
    &+ \left(8(1+\tradetmp) \eta^2 \frac{\hlipschitz^2}{\vert \honest \vert} + 2\eta^2 \frac{1}{\vert \honest \vert^2} \frac{4\dimension}{\nseeds} 4 \lipschitz^2\right)  \sumhc \ec{\histl[\lepoch-1]}{\sqnorm{\modeltll[\lepoch-1] - \modeltmavg[\lepoch-1]}} + 2\eta^2 \frac{1}{\vert \honest \vert} \frac{4\dimension}{\nseeds} 4 \graddiv \\
    &+ 2\eta^2 \frac{1}{\vert \honest \vert} \frac{4\dimension}{\nseeds} \gradvar +  2\eta^2 \frac{1}{\vert \honest \vert} \frac{\lipschitz^2 \mu^2 \dimension^2}{2\nseeds}}
\end{align*}
where $(a)$ is by independence for $i \neq j$, and $(b)$ is because $\sqnorm{\mathbf{x}+\mathbf{y}} = (1+1/\tradetmp) \sqnorm{\mathbf{x}} + (1+\tradetmp) \sqnorm{\mathbf{y}}$, $\tradetmp>0$. $(c)$ follows from \cref{ass:lipschitz}, \cref{ass:lipschitz_avg_grad} and an intermediate step in the proof of \cref{lemma:zovar}.

To ensure the bound holds uniformly for all $\lepoch \in [\nlepochs]$, we now choose $\tradetmp=3\nlepochs-1$ and the learning rate small enough so that $\left((1+\frac{1}{\tradetmp}) + 8(1+\tradetmp) \eta^2 \lipschitz^2 + 2\eta^2 \frac{1}{\vert \honest \vert} \frac{4\dimension}{\nseeds} 4 \lipschitz^2 \right) \leq 1 + \frac{1}{\nlepochs-1}$, i.e., that $8(1+\tradetmp) \eta^2 \lipschitz^2 \leq \frac{1}{3\nlepochs}$ and $2\eta^2 \frac{1}{\vert \honest \vert} \frac{4\dimension}{\nseeds} 4 \lipschitz^2 \leq \frac{1}{3\nlepochs}$. 
With this choice of the learning rate, we have
\begin{align*}
    &\ec{\histl[\lepoch-1]}{\sqnorm{\modeltlavg-\modelt}} \\
    &\leq (1+\frac{1}{\nlepochs-1}) \ec{\histl[\lepoch-1]}{\sqnorm{\modeltmavg[\lepoch-1] - \modelt}} + 8(1+\tradetmp) \eta^2 \approxerr 
    + \left(8(1+\tradetmp) \eta^2 + 2\eta^2 \frac{1}{\vert \honest \vert} \frac{4\dimension}{\nseeds} 4\right) \ec{\histl[\lepoch-1]}{\sqnorm{\nlossg(\modelt)}} \\
    &+ \left(8(1+\tradetmp) \eta^2 \frac{\hlipschitz^2}{\vert \honest \vert} + 2\eta^2 \frac{1}{\vert \honest \vert^2} \frac{4\dimension}{\nseeds} 4 \lipschitz^2\right)  \sumhc \ec{\histl[\lepoch-1]}{\sqnorm{\modeltll[\lepoch-1] - \modeltmavg[\lepoch-1]}} + 2\eta^2 \frac{1}{\vert \honest \vert} \frac{4\dimension}{\nseeds} 4 \graddiv + 2\eta^2 \frac{1}{\vert \honest \vert} \frac{4\dimension}{\nseeds} \gradvar +  2\eta^2 \frac{1}{\vert \honest \vert} \frac{\lipschitz^2 \mu^2 \dimension^2}{2\nseeds} \\
    &\overset{(d)}{\leq} \consteprime \ec{\histl[\lepoch-1]}{\sqnorm{\nlossg(\modelt)}} + \constfprime + \constgprime \sum_{\lepoch^\prime=1}^{\lepoch-1} \sumhc \ec{\histl[\lepoch^\prime-1]}{\sqnorm{\modeltll[\lepoch^\prime] - \modeltmavg[\lepoch^\prime]}},
\end{align*}
where $(e)$ is by the recursive application of $(d)$ and the fact that $(1+\frac{1}{\nlepochs})^\lepoch \leq (1+\frac{1}{\lepoch})^\lepoch \leq e$ for all $\lepoch \in [\nlepochs]$. 
This concludes the proof. The constants are given as
\begin{align*}
    \consteprime &\define 5(\lepoch-1)  \left(8(1+\tradetmp) \eta^2 + 2\eta^2 \frac{1}{\vert \honest \vert} \frac{4\dimension}{\nseeds} 4\right) \leq \conste \define 5\cdot 32 \lr^2 \nlepochs \left(\nlepochs + \frac{\dimension}{\vert \honest \vert \nseeds} \right) \\
    \constfprime &\define 5(\lepoch-1) \left(2\eta^2 \frac{1}{\vert \honest \vert} \frac{4\dimension}{\nseeds} 4 \graddiv + 2\eta^2 \frac{1}{\vert \honest \vert} \frac{4\dimension}{\nseeds} \gradvar +  2\eta^2 \frac{1}{\vert \honest \vert} \frac{\lipschitz^2 \mu^2 \dimension^2}{2\nseeds} + 8(1+\tradetmp) \eta^2 \approxerr\right) \\
    &\leq \constf \define 5\nlepochs \lr^2 \frac{\dimension}{\vert \honest \vert\nseeds} \left(32 \graddiv + 8 \gradvar +  \lipschitz^2 \mu^2 \dimension\right) + 5 \lr^2 24\nlepochs^2 \approxerr \\
    \constgprime &\define 5\left(8(1+\tradetmp) \eta^2 \hlipschitz^2 + 2\eta^2 \frac{1}{\vert \honest \vert} \frac{4\dimension}{\nseeds} 4 \lipschitz^2\right) \leq \constg \define 5 \cdot 32\lr^2 \left(\nlepochs \hlipschitz^2 + \frac{1}{\vert \honest \vert} \frac{\dimension}{\nseeds} \lipschitz^2\right)
\end{align*}
\end{proof}

\begin{proof}[Proof of \cref{lemma:zovar}]
For the proof of the zero-order approximated gradient variance, we rely on the following intermediate lemma.
\begin{lemma} \label{lem:est_var}
The second moment of the gradient estimate can be bounded from above as
\begin{align*}
    \ec{\mathbf{z}}{\sqnorm{\perturbvec \projtmp}} \leq 2\dimension \sqnorm{\nlossg(\model, \globaldata)} + \frac{\lipschitz^2 \mu^2 \dimension^2}{2}.
\end{align*}
\end{lemma}
\begin{proof}
\begin{align*}
    &\ec{\perturbvec \sim \uniform}{\sqnorm{\perturbvec \projtmp}} = \ec{\perturbvec \sim \uniform}{\sqnorm{\dimension \frac{\loss(\model + \pscale \perturbvec) - \loss(\model - \pscale \perturbvec)}{2\pscale}}} \\
    &= \ec{\perturbvec \sim \uniform}{\sqnorm{\dimension \frac{\loss(\model + \pscale \perturbvec) - \loss(\model) + \loss(\model) - \loss(\model - \pscale \perturbvec)}{2\pscale}}} \\
    &\leq \frac{1}{2} \ec{\perturbvec \sim \uniform}{\sqnorm{\dimension \frac{\loss(\model + \pscale \perturbvec) - \loss(\model)}{\pscale}}} + \frac{1}{2} \ec{\perturbvec \sim \uniform}{\sqnorm{\dimension \frac{\loss(\model - \pscale \perturbvec)-\loss(\model)}{\pscale}}} \\
    &= \ec{\perturbvec \sim \uniform}{\sqnorm{\dimension \frac{\loss(\model + \pscale \perturbvec) - \loss(\model)}{\pscale}}} \\
    &\leq 2\dimension \sqnorm{\nlossg(\model)} + \frac{\lipschitz^2 \mu^2 \dimension^2}{2},
\end{align*}
where the penultimate step holds by symmetry and the last step is from \citep[Lemma 4.1]{gao2018information}.
\end{proof}
We bound the zero-order approximated gradient variance as follows:
Since $\e{\sqnorm{Z-\e{Z}}} \leq \e{\sqnorm{Z}}$, and $\e{\perturbtlr \projitlr} = \nlossiu(\modeltl)$, we have
\begin{align*}
    &\e{\sqnorm{\perturbtl \projitl - \nlossiu(\modeltl)}} = \e{\sqnorm{\frac{1}{\nseeds} \sum_{\seed = 1}^\nseeds \perturbtlr \projitlr - \nlossiu(\modeltl)}} \\
    &\overset{(a)}{=} \frac{1}{\nseeds^2} \sum_{\seed = 1}^\nseeds \e{\sqnorm{\perturbtlr \projitlr - \nlossiu(\modeltl)}} \\
    &\leq \frac{1}{\nseeds^2} \sum_{\seed = 1}^\nseeds \e{\sqnorm{\perturbtlr \projitlr}} \\
    &\overset{(b)}{\leq} \frac{2\dimension}{\nseeds^2} \sum_{\seed = 1}^\nseeds \ec{\histl[\lepoch-1]}{\ec{\randl[\lepoch]}{\sqnorm{\graditl}}} + \frac{\lipschitz^2 \mu^2 \dimension^2}{2\nseeds} \\
    &\leq \frac{2\dimension}{\nseeds^2} \sum_{\seed = 1}^\nseeds \ec{\histl[\lepoch-1]}{\ec{\randl[\lepoch]}{\sqnorm{\graditl - \nlossi(\modeltl) + \nlossi(\modeltl)}}} + \frac{\lipschitz^2 \mu^2 \dimension^2}{2\nseeds} \\
    &\leq \frac{2\dimension}{\nseeds^2} \sum_{\seed = 1}^\nseeds \ec{\histl[\lepoch-1]}{2\ec{\randl[\lepoch]}{\sqnorm{\graditl - \nlossi(\modeltl)}} + 2\sqnorm{\nlossi(\modeltl)}} + \frac{\lipschitz^2 \mu^2 \dimension^2}{2\nseeds} \\
    &\overset{(c)}{\leq} \frac{4\dimension}{\nseeds^2} \sum_{\seed = 1}^\nseeds \ec{\histl[\lepoch-1]}{\sqnorm{\nlossi(\modeltl)}} + \frac{4\dimension}{\nseeds} \gradvar +  \frac{\lipschitz^2 \mu^2 \dimension^2}{2\nseeds} \numberthis \label{eq:grad_var_intermediate}
\end{align*}
where $(a)$ is due to the independence of $\perturbtlr \projitlr$ and $\perturbtlr[\seed^\prime] \projitlr[\seed^\prime]$ for $\seed \neq \seed^\prime$. $(b)$ is by \cref{lem:est_var}. $(c)$ follows from \cref{ass:bounded_grad_var}.
\begin{align*}
    \ec{\histl[\lepoch-1]}{\sqnorm{\nlossi(\modeltl)}} &= \ec{\histl[\lepoch-1]}{\sqnorm{\nlossi(\modeltl) - \nlossi(\modeltlavg) + \nlossi(\modeltlavg) - \nlossg(\modeltlavg) + \nlossg(\modeltlavg) - \nlossg(\modelt) + \nlossg(\modelt)}} \\
    &\leq 4\ec{\histl[\lepoch-1]}{\sqnorm{\nlossi(\modeltl) - \nlossi(\modeltlavg)}} + 4\ec{\histl[\lepoch-1]}{\sqnorm{\nlossi(\modeltlavg) - \nlossg(\modeltlavg)}} \\
    &+ 4\ec{\histl[\lepoch-1]}{\sqnorm{\nlossg(\modeltlavg) - \nlossg(\modelt)}} + 4\ec{\histl[\lepoch-1]}{\sqnorm{\nlossg(\modelt)}} \\
    &\leq 4 \lipschitz^2 \ec{\histl[\lepoch-1]}{\sqnorm{\modeltl - \modeltlavg}} + 4 \graddiv + 4 \lipschitz^2 \ec{\histl[\lepoch-1]}{\sqnorm{\modeltlavg-\modelt}} + 4\ec{\histl[\lepoch-1]}{\sqnorm{\nlossg(\modelt)}}.
\end{align*}

Substituting the result in \eqref{eq:grad_var_intermediate}, we obtain
\begin{align*}
    &\e{\sqnorm{\perturbtl \projitl - \nlossiu(\modeltl)}} = \e{\sqnorm{\frac{1}{\nseeds} \sum_{\seed = 1}^\nseeds \perturbtlr \projitlr - \nlossiu(\modeltl)}} \\
    &\leq \frac{4\dimension}{\nseeds^2} \sum_{\seed = 1}^\nseeds \left(4 \lipschitz^2 \ec{\histl[\lepoch-1]}{\sqnorm{\modeltl - \modeltlavg}} + 4 \graddiv + 4 \lipschitz^2 \ec{\histl[\lepoch-1]}{\sqnorm{\modeltlavg-\modelt}} + 4\ec{\histl[\lepoch-1]}{\sqnorm{\nlossg(\modelt)}}\right) + \frac{4\dimension}{\nseeds} \gradvar +  \frac{\lipschitz^2 \mu^2 \dimension^2}{2\nseeds} \\
    &= \frac{4\dimension}{\nseeds} \left(4 \lipschitz^2 \ec{\histl[\lepoch-1]}{\sqnorm{\modeltl - \modeltlavg}} + 4 \graddiv + 4 \lipschitz^2 \ec{\histl[\lepoch-1]}{\sqnorm{\modeltlavg-\modelt}} + 4\ec{\histl[\lepoch-1]}{\sqnorm{\nlossg(\modelt)}}\right) + \frac{4\dimension}{\nseeds} \gradvar +  \frac{\lipschitz^2 \mu^2 \dimension^2}{2\nseeds} \\
    &\leq \frac{4\dimension}{\nseeds} \left(4 \lipschitz^2 \ec{\histl[\lepoch-1]}{\sqnorm{\modeltl - \modeltlavg}} + 4 \graddiv + 4\ec{\histl[\lepoch-1]}{\sqnorm{\nlossg(\modelt)}}\right) + \frac{4\dimension}{\nseeds} \gradvar +  \frac{\lipschitz^2 \mu^2 \dimension^2}{2\nseeds} \\
    & +\frac{4\dimension}{\nseeds} 4 \lipschitz^2 \bigg(\conste \ec{\histl[\lepoch-1]}{\sqnorm{\nlossg(\modelt)}} + \constf + \constg \sum_{\lepoch^\prime=1}^{\lepoch-1} \frac{1}{\vert \honest \vert} \sumhc \ec{\histl[\lepoch^\prime-1]}{\sqnorm{\modeltll[\lepoch^\prime] - \modeltmavg[\lepoch^\prime]}}\bigg) \\
    &\leq \consthprime  \ec{\histl[\lepoch-1]}{\sqnorm{\modeltl - \modeltlavg}} + \constkprime + \constjprime \ec{\histl[\lepoch-1]}{\sqnorm{\nlossg(\modelt)}} + \constiprime \sum_{\lepoch^\prime=1}^{\lepoch-1} \frac{1}{\vert \honest \vert} \sumhc \ec{\histl[\lepoch^\prime-1]}{\sqnorm{\modeltll[\lepoch^\prime] - \modeltmavg[\lepoch^\prime]}},
\end{align*}
where the latter is by \cref{lemma:model_progress}, and we have
\begin{align*}
    \consthprime &\define \consth \define \frac{4\dimension}{\nseeds} 4 \lipschitz^2 \\
    \constiprime & \define \frac{4\dimension}{\nseeds} 4 \lipschitz^2 \constg \leq \consti \define \frac{4\dimension}{\nseeds} 4 \lipschitz^2 5 \cdot 32\lr^2 \left(\nlepochs \hlipschitz^2 + \frac{1}{\vert \honest \vert} \frac{\dimension}{\nseeds} \lipschitz^2\right) \\
    \constjprime & \define \frac{4\dimension}{\nseeds} 4 + \frac{4\dimension}{\nseeds} 4 \lipschitz^2 \conste \leq \constj \define \frac{4\dimension}{\nseeds} 4 + \frac{4\dimension}{\nseeds} 4 \lipschitz^2 5\cdot 32 \lr^2 \nlepochs \left(\nlepochs + \frac{\dimension}{\vert \honest \vert \nseeds} \right) \\
    \constkprime & \define \frac{4\dimension}{\nseeds} 4 \graddiv + \frac{4\dimension}{\nseeds} \gradvar +  \frac{\lipschitz^2 \mu^2 \dimension^2}{2\nseeds} + \frac{4\dimension}{\nseeds} 4 \lipschitz^2 \constf \\
    &\leq \frac{\dimension}{2\nseeds} \left(32 \graddiv + 8 \gradvar +  \lipschitz^2 \mu^2 \dimension\right) + \frac{4\dimension}{\nseeds} 4 \lipschitz^2\left(5\nlepochs \lr^2 \frac{\dimension}{\vert \honest \vert\nseeds} \left(32 \graddiv + 8 \gradvar +  \lipschitz^2 \mu^2 \dimension\right) + 5 \lr^2 24\nlepochs^2 \approxerr\right) \\
    &\leq \constk \define \left(\frac{\dimension}{2\nseeds} + \frac{80\dimension^2}{\nseeds^2 \vert \honest \vert} \lipschitz^2 \nlepochs \lr^2 \right) \left(32 \graddiv + 8 \gradvar +  \lipschitz^2 \mu^2 \dimension\right) + \frac{4\dimension}{\nseeds} 4 \lipschitz^2 5 \lr^2 24\nlepochs^2 \approxerr.
\end{align*}
This concludes the proof.
\end{proof}

\begin{proof}[Proof of \cref{lemma:graddiv}]

We start with stating an intermediate lemma proven by \citet{wang2025new}.
\begin{lemma}[Extracted from Lemma B.1, \citep{wang2025new}] The following holds for the divergence of the local gradients:
\begin{align*}
    \sqnorm{\frac{1}{\vert \honest \vert} \sumhcj \nlossj(\modeltlj) - \nlossi(\modeltl)} \leq 3 \frac{\hlipschitz^2}{\vert \honest \vert} \sumhcj \sqnorm{\modeltlavg - \modeltlj} + 3\lipschitz + 3\graddiv \sqnorm{\modeltlavg - \modeltl}.
\end{align*}
\end{lemma}

Following similar lines as in the proof of \citep[Lemma B.2]{wang2025new}, for some local iteration $\lepochtmp \in [\lepoch]$, we have
\begin{align*}
    &\e{\sqnorm{\perturbtm \projitl - \frac{1}{\vert \honest \vert} \sumhcj \perturbtm \projjtm}} \\
    &= \mathbb{E}\Bigg[ \bigg\Vert  \perturbtm \projitm - \nlossi(\modeltm) + \nlossi(\modeltm) \\
    &- \frac{1}{\vert \honest \vert} \sumhcj \nlossj(\modeltmj) + \frac{1}{\vert \honest \vert} \sumhcj \nlossj(\modeltmj) - \frac{1}{\vert \honest \vert} \sumhcj \perturbtm \projjtm \bigg\Vert \Bigg] \\
    &\leq 2 \e{\sqnorm{\nlossi(\modeltm) - \frac{1}{\vert \honest \vert} \sumhcj \nlossj(\modeltmj)}} \\
    &+ 2 \e{\sqnorm{\perturbtm \projitm - \nlossi(\modeltm) + \frac{1}{\vert \honest \vert} \sumhcj \nlossj(\modeltmj) - \frac{1}{\vert \honest \vert} \sumhcj \perturbtm \projjtm}} \\
    &\overset{(a)}{\leq} 2  \e{\sqnorm{\nlossi(\modeltm) - \frac{1}{\vert \honest \vert} \sumhcj \nlossj(\modeltmj) }} \\
    &+ 2 \e{\sqnorm{\perturbtm \projitm - \nlossi(\modeltm)}} \\
    &\leq 2 \e{3 \frac{\hlipschitz^2}{\vert \honest \vert} \sumhcj \sqnorm{\modeltmavg[\lepochtmp] - \modeltmj} + 3\lipschitz + 3\graddiv \sqnorm{\modeltmavg[\lepochtmp] - \modeltm}} \\
    &+ 2 \e{\sqnorm{ \left(\perturbtm \projitm - \nlossi(\modeltm) \right)}}, \numberthis \label{eq:graddiv_lipschitz} \\
\end{align*}
where $(a)$ is since $\frac{1}{\vert \honest \vert} \sumhc \sqnorm{\mathbf{x}_i - \frac{1}{\vert \honest \vert} \sumhcj \mathbf{x}_\clienttmp} = \frac{1}{\vert \honest \vert} \sumhcj \sqnorm{\mathbf{x}_\clienttmp} - \sqnorm{\frac{1}{\vert \honest \vert} \sumhcj \mathbf{x}_\clienttmp} \leq \frac{1}{\vert \honest \vert} \sumhc \sqnorm{\mathbf{x}_i}$ for $\mathbf{x}_i \in \mathbb{R}^d, \forall i$, where we set $\mathbf{x}_i = \perturbtm \projitm - \nlossi(\modeltm)$.

Summing over all benign clients, we obtain
\begin{align*}
    &\sumletmp \frac{1}{\vert \honest \vert} \sumhc \e{\sqnorm{\perturbtm \projitl - \frac{1}{\vert \honest \vert} \sumhcj \perturbtm \projjtm}} \\
    &\overset{(a)}{\leq} 2 \sum_{\lepochtmp=1}^{\lepoch} \e{\left(3 \frac{\hlipschitz^2}{\vert \honest \vert} + 3\frac{\graddiv}{\vert \honest \vert}\right) \sumhc \sqnorm{\modeltmavg[\lepochtmp] - \modeltm} + 3\lipschitz} \\
    &+ 2 \sumletmp \frac{1}{\vert \honest \vert} \sumhc \bigg( \consth[\lepochtmp]  \ec{\histl[\lepochtmp-1]}{\sqnorm{\modeltm - \modeltmavg[\lepochtmp]}} + \constk[\lepochtmp] + \constj[\lepochtmp] \ec{\histl[\lepochtmp-1]}{\sqnorm{\nlossg(\modelt)}} + \consti[\lepochtmp] \sum_{\lepoch^\prime=1}^{\lepochtmp-1} \sumhc \ec{\histl[\lepoch^\prime-1]}{\sqnorm{\modeltll[\lepoch^\prime] - \modeltmavg[\lepoch^\prime]}}\bigg) \\
    &\leq 2 \sumletmp \frac{1}{\vert \honest \vert} \sumhc \bigg( \left(3 \hlipschitz^2 + 3\graddiv + \consth[\lepochtmp]\right)   \ec{\histl[\lepochtmp-1]}{\sqnorm{\modeltm - \modeltmavg[\lepochtmp]}} + \constk[\lepochtmp] + 3\lipschitz \\
    &+\constj[\lepochtmp] \ec{\histl[\lepochtmp-1]}{\sqnorm{\nlossg(\modelt)}} + \consti[\lepochtmp] \sum_{\lepoch^\prime=1}^{\lepochtmp-1} \sumhc \ec{\histl[\lepoch^\prime-1]}{\sqnorm{\modeltll[\lepoch^\prime] - \modeltmavg[\lepoch^\prime]}}\bigg) \\
    &\leq 2 \sumletmp \frac{1}{\vert \honest \vert} \sumhc \bigg( \left(3 \hlipschitz^2 + 3\graddiv + \consth[\lepochtmp]\right)   \ec{\histl[\lepochtmp-1]}{\sqnorm{\modeltm - \modeltmavg[\lepochtmp]}} + \constk[\lepochtmp] + 3\lipschitz \\
    &+\constj[\lepochtmp] \ec{\histl[\lepochtmp-1]}{\sqnorm{\nlossg(\modelt)}} + \consti[\lepochtmp] \lepoch (\lepoch-1) \frac{1}{\vert \honest \vert} \sumhc \ec{\histl[\lepochtmp-1]}{\sqnorm{\modeltll[\lepochtmp] - \modeltmavg[\lepochtmp]}}\bigg) \\
    &\leq \sum_{\lepochtmp=1}^{\lepoch} \frac{1}{\vert \honest \vert} \sumhc \constaprime \e{ \sqnorm{\modeltmavg[\lepochtmp] - \modeltm}} + \sum_{\lepochtmp=1}^{\lepoch} (\constb + \constc) + \sum_{\lepochtmp=1}^{\lepoch} \constd \ec{\histl[\lepochtmp-1]}{\sqnorm{\nlossg(\modelt)}},
\end{align*}
where $(a)$ is due to \cref{lemma:zovar} and $(b)$ is since $\sum_{\lepochtmp=1}^\lepoch \sum_{\lepoch^\prime=1}^{\lepochtmp} x_m \leq \frac{\lepoch (\lepoch-1)}{2} \sum_{\lepochtmp} x_m$. Thereby,
\begin{align*}
    \constaprime[\lepoch] &\define 2 (3 \hlipschitz^2 + 3\graddiv + \consth[\lepochtmp] + \consti[\lepochtmp] \lepoch (\lepoch-1)) \leq \consta \define 6 \hlipschitz^2 + 6\graddiv + 2\consth[\lepochtmp] + 2\consti[\lepochtmp] \nlepochs^2  \\
    \constb &\define 2 \constk[\lepochtmp] \\
    \constc &\define 6 \lipschitz \\
    \constd &\define 2 \constj[\lepochtmp].
\end{align*}
This concludes the proof.
\end{proof}

\begin{proof}[Proof of \cref{lemma:model_divergence}]

From \cref{lemma:graddiv}, we have
\begin{align*}
    &\sumle \frac{1}{\vert \honest \vert} \sumhc \e{\sqnorm{\modeltl-\modeltlavg}} = \sumle \frac{1}{\vert \honest \vert} \sumhc \e{\sqnorm{\lr \sum_{\lepochtmp=1}^{\lepoch} \left(\perturbtm \projitl - \frac{1}{\vert \honest \vert} \sumhcj \perturbtm \projjtm \right)}} \\
    &\leq \lr^2 \sumle \lepoch \sum_{\lepochtmp=1}^{\lepoch} \frac{1}{\vert \honest \vert} \sumhc \e{\sqnorm{\perturbtm \projitl - \frac{1}{\vert \honest \vert} \sumhcj \perturbtm \projjtm}} \\
    &= \lr^2 \sumle \lepoch \sum_{\lepochtmp=1}^{\lepoch} \sumhc \frac{1}{\vert \honest \vert}  \consta \e{ \sqnorm{\modeltmavg[\lepochtmp] - \modeltm}} + \lr^2 \sumle \lepoch \sum_{\lepochtmp=1}^\lepoch \constb + \lr^2 \sumle \lepoch \sumletmp \constc + \lr^2 \sumle \lepoch \sum_{\lepochtmp=1}^\lepoch \constd \ec{\histl[\lepochtmp-1]}{\sqnorm{\nlossg(\modelt)}} \\
    &\leq \lr^2 \sumle \frac{1}{\vert \honest \vert} \sumhc \nlepochs^2 \consta[\nlepochs] \e{ \sqnorm{\modeltmavg[\lepoch] - \modeltl}} + \lr^2 \sumle \lepoch \sum_{\lepochtmp=1}^\lepoch \constb + \lr^2 \sumle \lepoch \sumletmp \constc + \lr^2 \sumle \lepoch \sum_{\lepochtmp=1}^\lepoch \constd \ec{\histl[\lepochtmp-1]}{\sqnorm{\nlossg(\modelt)}}.
\end{align*}
We rewrite the equation as
\begin{align*}
\left(1-\lr^2 \nlepochs^2 \consta[\nlepochs]\right) \sumle \frac{1}{\vert \honest \vert} \sumhc \e{\sqnorm{\modeltl-\modeltlavg}} \leq \lr^2 \sumle \lepoch \sum_{\lepochtmp=1}^\lepoch \constb + \lr^2 \sumle \lepoch \sumletmp \constc + \lr^2 \sumle \lepoch \sum_{\lepochtmp=1}^\lepoch \constd \ec{\histl[\lepochtmp-1]}{\sqnorm{\nlossg(\modelt)}}
\end{align*}
and choose the learning rate small enough so that $\left(1-\lr^2 \nlepochs^2 \consta[\nlepochs]\right) \geq \frac{1}{2}$. Letting $\consta[\nlepochs] = 6 \hlipschitz^2 + 6\graddiv + \frac{32\dimension}{\nseeds} \lipschitz^2 + \frac{160 \cdot 32\dimension}{\nseeds} \lipschitz^2 \lr^2 \left(\nlepochs \hlipschitz^2 + \frac{1}{\vert \honest \vert} \frac{\dimension}{\nseeds} \lipschitz^2\right) \nlepochs^2$, we require that $\lr \leq \sqrt{6 \frac{\hlipschitz^2}{\nlepochs^2} + 6\frac{\graddiv}{\nlepochs^2} + \frac{32\dimension}{\nseeds \nlepochs^2} \lipschitz^2 + \frac{160 \cdot 32\dimension}{\nseeds} \lipschitz^2 \lr^2 \left(\nlepochs \hlipschitz^2 + \frac{1}{\vert \honest \vert} \frac{\dimension}{\nseeds} \lipschitz^2\right)}$. Since $\frac{160 \cdot 32\dimension}{\nseeds} \lipschitz^2 \lr^2 \left(\nlepochs \hlipschitz^2 + \frac{1}{\vert \honest \vert} \frac{\dimension}{\nseeds} \lipschitz^2\right)\geq 0$, it suffices to let $\lr \leq \sqrt{6 \frac{\hlipschitz^2}{\nlepochs^2} + 6\frac{\graddiv}{\nlepochs^2} + \frac{32\dimension}{\nseeds \nlepochs^2} \lipschitz^2}$. 
Hence, we obtain
\begin{align*}
\sumle \frac{1}{\vert \honest \vert} \sumhc \e{\sqnorm{\modeltl-\modeltlavg}} \leq \constmprime + \constlprime \ec{\histl[\lepochtmp-1]}{\sqnorm{\nlossg(\modelt)}},
\end{align*}
where 
\begin{align*}
    \constlprime &\define 2\lr^2 \sumle \lepoch \sum_{\lepochtmp=1}^\lepoch \constd \leq 2\lr^2 \nlepochs^3 \constd \leq \constl \define 4 \lr^2 \nlepochs^3 \constj[\lepochtmp], \\ %
    \constmprime &\define 2\lr^2 \sumle \lepoch \sum_{\lepochtmp=1}^\lepoch \constb + 2\lr^2 \sumle \lepoch \sumletmp \constc \leq \constm \define 4\lr^2 \nlepochs^3 \constk[\lepochtmp] + 12\lr^2 \nlepochs^3 \lipschitz. %
\end{align*}

\end{proof}

\subsubsection{Proof of \cref{thm:convergence_non_convex} for $\pscale=0$} \label{subsubsec:proof_mu_0}
\begin{proof}
For the proof of the projected gradient variance, we rely on the following intermediate lemma.
\begin{lemma} \label{lem:est_var_proj} The second moment of the projected gradient according to \cref{def:zero_order_estimate} for $\pscale=0$ is bounded as
\begin{align*}
    \ec{\mathbf{z}}{\sqnorm{\perturbvec \projtmp}} \leq \dimension \sqnorm{\nabla \loss(\model, \globaldata)}.
\end{align*}
\begin{proof}[Proof of \cref{lem:est_var_proj}]
\begin{align*}
\ec{\mathbf{z}}{\sqnorm{\perturbvec \projtmp}} &= \ec{\mathbf{z}}{\sqnorm{\dimension \perturbvec \innerprod{\nabla \loss(\model, \globaldata)}{\perturbvec}}} \\
&= \ec{\mathbf{z}}{\sqnorm{\dimension \perturbvec \perturbvec^T \nabla \loss(\model, \globaldata)}} \\
&= \ec{\mathbf{z}}{\dimension^2 \nabla \loss(\model, \globaldata)^T \perturbvec \perturbvec^T \perturbvec \perturbvec^T \nabla \loss(\model, \globaldata)} \\
&= \dimension^2 \nabla \loss(\model, \globaldata)^T \ec{\mathbf{z}}{\perturbvec \perturbvec^T} \nabla \loss(\model, \globaldata) \\
&= \dimension \nabla \loss(\model, \globaldata)^T \nabla \loss(\model, \globaldata) \\
&= \dimension \sqnorm{\nabla \loss(\model, \globaldata)},
\end{align*}
where the penultimate step is by \citep[Lemma 7.3]{gao2018information}, which states that $\ec{\mathbf{z}}{\perturbvec \perturbvec^T} = \frac{1}{\dimension} \mathbf{I}$, for $\mathbf{I}$ being the identity matrix.
\end{proof}
\end{lemma}
Accordingly, the bound in \cref{lem:est_var} for the zero-order estimate is an upper bound to the result of \cref{lem:est_var_proj} when choosing $\pscale=0$ in \cref{lem:est_var}. Further, we observe that \cref{prop:bias} still holds for $\pscale=0$, due to a non-zero bias in the gradient projection case. Since those are the only two intermediate results where the case of gradient projection differs from the zero-order estimate, the result established in \cref{thm:convergence_non_convex} holds for the gradient projection case when choosing $\pscale=0$.
\end{proof}

\subsection{Proof of \cref{thm:convergence_non_convex_lipschitz}}

\begin{proof}
We assume for all that follows that the objective $\loss$ exhibits a $\glipschitz$-Lipschitz behavior. Similar to the proof of \cref{thm:convergence_non_convex}, we first state necessary intermediate lemmas, which we prove in the sequel. All lemmas hold under \cref{ass:bounded_grad_div,ass:bounded_grad_var,ass:lipschitz_avg_grad,ass:lipschitz,ass:lipschitz_objective} and a robust aggregator according to \cref{def:robustness}.
\begin{lemma} \label{lemma:model_progress_lipschitz}
Let
\begin{align*}
    \conste &\define 5\cdot 16 \lr^2 \nlepochs^2 \\
    \constf &\define 5\nlepochs \lr^2 \frac{\glipub}{\vert \honest \vert\nseeds} + 5 \lr^2 16\nlepochs^2 \approxerr \\
    \constg &\define 5 \cdot 16\lr^2 \left(\nlepochs \hlipschitz^2\right)
\end{align*}

For a learning rate satisfying $\lr \leq \sqrt{\frac{1}{32 \lipschitz^2 \nlepochs^2}}$, we have the following upper bound on the averaged local model divergence:
\begin{align*}
    &\ec{\histl[\lepoch-1]}{\sqnorm{\modeltlavg-\modelt}} \leq \conste \ec{\histl[\lepoch-1]}{\sqnorm{\nlossg(\modelt)}} + \constf + \constg \sum_{\lepoch^\prime=1}^{\lepoch-1} \frac{1}{\vert \honest \vert} \sumhc \ec{\histl[\lepoch^\prime-1]}{\sqnorm{\modeltll[\lepoch^\prime] - \modeltmavg[\lepoch^\prime]}}
\end{align*}
\end{lemma}

\begin{lemma} \label{lemma:zovar_lipschitz}
The following holds for the gradient estimate variance
    \begin{align*}
    &\e{\sqnorm{\perturbtl \projitl - \nlossiu(\modeltl)}} \leq \frac{\glipub}{\nseeds}
\end{align*}
\end{lemma}

\begin{lemma} \label{lemma:graddiv_lipschitz}
Let
\begin{align*}
    \consta[\lepoch] &\define 6 \hlipschitz^2 + 6\graddiv \\
    \constb &\define 2 \frac{\glipub}{\nseeds} \\
    \constc &\define 6 \lipschitz
\end{align*}
Then
\begin{align*}
    &\sumletmp \frac{1}{\vert \honest \vert} \sumhc \e{\sqnorm{\perturbtm \projitl - \frac{1}{\vert \honest \vert} \sumhcj \perturbtm \projjtm}} \\
    &\leq \sum_{\lepochtmp=1}^{\lepoch} \frac{1}{\vert \honest \vert} \sumhc \consta \e{ \sqnorm{\modeltmavg[\lepochtmp] - \modeltm}} + \sum_{\lepochtmp=1}^\lepoch (\constb + \constc)
\end{align*}
\end{lemma}

\begin{lemma} \label{lemma:model_divergence_lipschitz}
Let
\begin{align*}
    \constm \define 4\lr^2 \nlepochs^3 \frac{\glipub}{\nseeds} + 12\lr^2 \nlepochs^3 \lipschitz. %
\end{align*}
For a learning rate that satisfies $\lr \leq \sqrt{\frac{1}{12 \nlepochs^2 (\hlipschitz^2 + \graddiv)}}$, we have
\begin{align*}
    \sumle \frac{1}{\vert \honest \vert} \sumhc \e{\sqnorm{\modeltl-\modeltlavg}} &\leq \constm
\end{align*}
\end{lemma}

To proof \cref{thm:convergence_non_convex_lipschitz}, we first follow the same lines as in the proof of \cref{thm:convergence_non_convex}, arriving at \cref{eq:main_equation}. We continue with bounding the individual terms.
\begin{align*}
    &\ec{\hist}{\sqnorm{\sumle \left( \sumhc \frac{1}{\vert\honest\vert} \nlossi(\modeltl) - \perturbtl \projitlavgh \right)}} \\
    &\leq 2\ec{\hist}{\sqnorm{\sumle \sumhc \frac{1}{\vert\honest\vert}\left(\nlossi(\modeltl) - \nlossiu(\modeltl)\right)}} + 2 \ec{\hist}{\sqnorm{\sumle \sumhc \frac{1}{\vert\honest\vert}\left( \nlossiu(\modeltl) - \perturbtl \projitl \right)}} \\
    &\overset{(a)}{\leq} 2 \frac{\nlepochs}{\vert\honest\vert} \sumhc \sumle \sqnorm{\left(\nlossi(\modeltl) - \nlossiu(\modeltl)\right)} + 2\frac{1}{\vert\honest\vert^2} \sumhc \ec{\hist}{\sqnorm{\sumle \left( \nlossiu(\modeltl) - \perturbtl \projitl \right)}} \\
    &\overset{(b)}{\leq} 2 \nlepochs^2 \approxerr + 2\frac{1}{\vert\honest\vert^2} \sumhc \sumle \ec{\randl}{\sqnorm{\nlossiu(\modeltl) - \perturbtl \projitl}} \\
    &\overset{(c)}{\leq} 2 \nlepochs^2 \approxerr + 2\frac{1}{\vert\honest\vert^2} \sumhc \sumle \frac{\glipub}{\nseeds} \\
    &\overset{(c)}{\leq} 2 \nlepochs^2 \approxerr + 2\frac{1}{\vert\honest\vert} \nlepochs \frac{\glipub}{\nseeds} \numberthis \label{eq:gradaggvar_lipschitz}\\
\end{align*}
where $(a)$ is due to the independence of $\sumle \left( \nlossiu(\modeltl) - \perturbtl \projitl \right)$ and $\sumle \left( \nlossju(\modeltlj) - \perturbtl \projjtl \right)$ for $\client \neq \clienttmp$. $(b)$ follows from \cref{prop:estexp} and \citep[Lemma 2]{wang2021novel}. $(c)$ is by the application of \cref{lemma:zovar_lipschitz}.

We continue with bounding the robustness term using as the main ingredient a double-sided application of an extended Johnson-Lindenstrauss Lemma as stated in \cref{lemma:modified_jl} and the application of \cref{lemma:graddiv_lipschitz}. We have
\begin{align*}
    &\e{\sqnorm{\sumle \perturbtl \left(\raggtl - \projitlavgh\right)}} \\
    &\leq \nlepochs \sumle \sqnorm{\perturbtl \left(\raggtl - \projitlavgh\right)} \\
    &\overset{(a)}{\leq} \nlepochs \sumle (1+\epsilon) \frac{\robustness}{\card{\honest}} \sumhc \e{\sqnorm{\projitl -\projitlavgh}} \\
    &\overset{(b)}{\leq} \nlepochs \sumle \frac{(1+\epsilon)}{(1-\epsilon)} \frac{\robustness}{\card{\honest}} \sumhc \e{\sqnorm{\perturbtl (\projitl -\projitlavgh)}} \\
    &\overset{(c)}{\leq} \nlepochs \frac{(1+\epsilon)}{(1-\epsilon)} \robustness \left(\sum_{\lepoch=1}^{\nlepochs} \frac{1}{\vert \honest \vert} \sumhc \consta[\lepoch] \e{ \sqnorm{\modeltmavg[\lepoch] - \modeltl}} + \sum_{\lepoch=1}^\nlepochs (\constb[\lepoch] + \constc[\lepoch])\right) \numberthis \label{eq:robustness}
\end{align*}
where $(a)$ and $(b)$ are by the application of \cref{lemma:modified_jl} for in total $\vert \honest \vert+1$ projections. By a union bound over \cref{lemma:modified_jl}, the distance preservation holds with probability $1-\delta$ for $\epsilon \geq \sqrt{\frac{64}{\nseeds} \log(\frac{2(\vert \honest \vert-1)}{ \delta})}$. We choose the smallest possible $\epsilon$. This must hold for each iteration, so the distance preservation holds w.p. at least $1-\nlepochs \delta$ for all local epochs. $(c)$ is by \cref{lemma:graddiv}.

To bound the local model divergence from the global model, by \cref{lemma:model_progress}, we have
\begin{align*}
    \sumle \ec{\histl[\lepoch-1]}{\sqnorm{\modeltlavg-\modelt}} &\leq \sumle \conste \ec{\histl[\lepoch-1]}{\sqnorm{\nlossg(\modelt)}} + \sumle \constf + \sumle \constg \sum_{\lepoch^\prime=1}^{\lepoch-1} \frac{1}{\vert \honest \vert} \sumhc \ec{\histl[\lepoch^\prime-1]}{\sqnorm{\modeltll[\lepoch^\prime] - \modeltmavg[\lepoch^\prime]}} \\
    &\leq \sumle \conste \ec{\histl[\lepoch-1]}{\sqnorm{\nlossg(\modelt)}} + \sumle \constf + \sumle \frac{\constg \nlepochs (\nlepochs-1)}{\vert \honest \vert} \sumhc \ec{\histl[\lepoch-1]}{\sqnorm{\modeltll[\lepoch] - \modeltmavg[\lepoch]}} \numberthis \label{eq:model_progress},
\end{align*}
where the latter follows since $\sum_{\lepoch=1}^\nlepochs \sum_{\lepoch^\prime=1}^{\lepoch-1} x_{\lepoch^\prime} \leq \sum_{\lepoch=1}^\nlepochs \sum_{\lepoch^\prime=1}^{\lepoch} x_{\lepoch^\prime} \leq \frac{\nlepochs (\nlepochs-1)}{2} \sum_{\lepoch} x_\lepoch$.

Plugging \eqref{eq:gradaggvar}, \eqref{eq:robustness}, and \eqref{eq:model_progress} into \eqref{eq:main_equation}, we obtain
\begin{align*}
    &\ec{\hist}{\sqnorm{\nlepochs \nlossh(\modelt)-\sumle \perturbtl \raggtl}} \nonumber \\
    &\leq 4 \nlepochs \lipschitz^2 \left(\sumle \conste \ec{\histl[\lepoch-1]}{\sqnorm{\nlossg(\modelt)}} + \sumle \constf + \sumle \frac{\constg \nlepochs (\nlepochs-1)}{\vert \honest \vert} \sumhc \ec{\histl[\lepoch-1]}{\sqnorm{\modeltll[\lepoch] - \modeltmavg[\lepoch]}}\right) \\
    &+ 4 \left(\nlepochs \frac{(1+\epsilon)}{(1-\epsilon)} \robustness \left(\sum_{\lepoch=1}^{\nlepochs} \frac{1}{\vert \honest \vert} \sumhc \consta[\lepoch] \e{ \sqnorm{\modeltmavg[\lepoch] - \modeltl}} + \sum_{\lepoch=1}^\nlepochs (\constb[\lepoch] + \constc[\lepoch])\right)\right) \nonumber \\
    &+ 4\nlepochs \sumle  \frac{\hlipschitz^2}{\vert \honest \vert} \sumhc \ec{\hist}{\sqnorm{\modeltl-\modeltlavg}} + 4 \left(2 \nlepochs^2 \approxerr + 2\frac{1}{\vert\honest\vert} \nlepochs \frac{\glipub}{\nseeds}\right) \\
    &\leq \constnprime \ec{\histl[\lepoch-1]}{\sqnorm{\nlossg(\modelt)}} + \constoprime \sum_{\lepoch=1}^{\nlepochs} \frac{1}{\vert \honest \vert} \sumhc \e{ \sqnorm{\modeltmavg[\lepoch] - \modeltl}} + \constpprime \\
    &\leq \constnprime \ec{\histl[\lepoch-1]}{\sqnorm{\nlossg(\modelt)}} + \constoprime \constm + \constpprime, \numberthis \label{eq:intermediate_bound_lipschitz}
\end{align*}
where
\begin{align*}
    \constnprime &\define 4 \nlepochs \lipschitz^2 \sumle \conste \leq \constn \define 4 \nlepochs^4 \lipschitz^2 5\cdot 16 \lr^2 \\
    \constoprime &\define 4 \nlepochs \lipschitz^2 \constg \nlepochs (\nlepochs-1) + 4 \nlepochs \frac{(1+\epsilon)}{(1-\epsilon)} \robustness \consta[\nlepochs] + 4\nlepochs \hlipschitz^2 \\
    &\leq 4 \nlepochs \lipschitz^2 5 \cdot 16\lr^2 \left(\nlepochs \hlipschitz^2\right) \nlepochs (\nlepochs-1) + 4 \nlepochs \robustness \frac{(1+\epsilon)}{(1-\epsilon)} 6(\hlipschitz^2+\graddiv) + 4\nlepochs \hlipschitz^2 \\
    &\leq \consto \define 20 \cdot 16 \nlepochs^4 \lipschitz^2 \lr^2 \hlipschitz^2 + 4 \nlepochs \frac{(1+\epsilon)}{(1-\epsilon)} \robustness 6(\hlipschitz^2+\graddiv) + 4\nlepochs \hlipschitz^2 \\
    \constpprime &\define 4 \nlepochs \lipschitz^2 \sumle \constf + 4 \nlepochs \frac{(1+\epsilon)}{(1-\epsilon)} \robustness \sum_{\lepoch=1}^\nlepochs (\constb[\lepoch] + \constc[\lepoch]) + 4 \cdot 2 \nlepochs^2 \approxerr + 4 \cdot 2\frac{1}{\vert\honest\vert^2} \nlepochs \frac{\glipub}{\nseeds} \\
    &\leq 4 \nlepochs \lipschitz^2 \sumle \left(5\nlepochs \lr^2 \frac{\glipub}{\vert \honest \vert\nseeds} + 5 \lr^2 16\nlepochs^2 \approxerr\right) +  4 \nlepochs \frac{(1+\epsilon)}{(1-\epsilon)} \robustness \sumle (2\frac{\glipub}{\nseeds}+6\lipschitz) \\
    &+ 4 \cdot 2 \nlepochs^2 \approxerr + 4 \cdot 2\frac{1}{\vert\honest\vert^2} \nlepochs \frac{\glipub}{\nseeds} \\
    &\leq 4 \nlepochs^2 \lipschitz^2 \left(5\nlepochs \lr^2 \frac{\glipub}{\vert \honest \vert\nseeds} + 5 \lr^2 16\nlepochs^2 \approxerr\right) +  4 \nlepochs^2 \frac{(1+\epsilon)}{(1-\epsilon)} \robustness (2\frac{\glipub}{\nseeds}+6\lipschitz) \\
    &+ 4 \cdot 2 \nlepochs^2 \approxerr + 4 \cdot 2\frac{1}{\vert\honest\vert^2} \nlepochs \frac{\glipub}{\nseeds} \\
    &\leq \constp \define 4\nlepochs \frac{\glipub}{\nseeds} \left(5 \nlepochs^2 \lipschitz^2 \lr^2 \frac{1}{\vert \honest \vert} + 2 \nlepochs \frac{(1+\epsilon)}{(1-\epsilon)} \robustness + \frac{8}{\vert \honest \vert^2} \right) + 4\nlepochs^2 \lipschitz \left(5\lr^2 16 \nlepochs^2 \lipschitz^2 \pscale + 6 \robustness \frac{(1+\epsilon)}{(1-\epsilon)} + 2\pscale\right) \\
\end{align*}

By taking the expectation over \eqref{eq:base_equation} and replacing $\e{\sqnorm{\nlepochs \nlossh(\modelt)-\sumle \perturbtl \raggtl}}$ by \eqref{eq:intermediate_bound}, we can write
\begin{align*}
    &\e{\lossh(\modelt[\gepoch+1])} - \e{\lossh(\modelt)} \\
    &\leq -\lr/(2) \nlepochs \e{\sqnorm{\nlossh(\modelt)}} + \left(\frac{\lr^2 \lipschitz}{2}-\frac{\lr}{2\nlepochs}\right) \e{\sqnorm{\sumle \raggtl \perturbtl}} + \lr/(2\nlepochs) \e{\sqnorm{\nlepochs \nlossh(\modelt)-\sumle \raggtl \perturbtl}} \\
    &\overset{(a)}{\leq} -\lr/(2) \nlepochs \e{\sqnorm{\nlossh(\modelt)}} +  \lr/(2\nlepochs) \left(\constn \ec{\histl[\lepoch-1]}{\sqnorm{\nlossg(\modelt)}} + \consto \constm + \constp \right) \\
    &\overset{(b)}{\leq} -\frac{\lr \nlepochs}{4} \e{\sqnorm{\nlossh(\modelt)}} +  \lr/(2\nlepochs) \left(\consto \constm + \constp \right),
\end{align*}
where $(a)$ holds when $\lr \leq \frac{1}{\nlepochs \lipschitz}$ and $(b)$ assumes that $\lr/(2\nlepochs) \constn = \lr/(2\nlepochs) 4 \nlepochs^2 \lipschitz^2 5\cdot 16 \lr^2 \nlepochs^2 \leq \frac{\lr \nlepochs}{4}$. The learning rate must hence satisfy $\lr^2 \leq \frac{1}{8\cdot \nlepochs^2 \lipschitz^2 5\cdot 16}$, and consequently $\lr \leq \frac{1}{26 \nlepochs \lipschitz}$.

Reordering and telescoping over $\gepoch$, we obtain
\begin{align*}
    \frac{1}{\ngepochs} \sum_{\gepoch=1}^{\ngepochs} \e{\sqnorm{\nlossh(\modelt)}} \leq \frac{4(\e{\lossh(\modelt[1])} - \e{\lossh(\modelt[\ngepochs+1])})}{\ngepochs \lr \nlepochs} + \lr/(2\nlepochs) \left(\consto \constm + \constp \right) \\
\end{align*}
with probability $1-\delta \nlepochs \ngepochs$ by a union bound argument over all global iterations $\ngepochs$. Noting that $\lossh(\modelt[\ngepochs+1]) \geq \optlossh$ by definition concludes the proof.

Since the learning rates must satisfy
$\lr \leq \frac{1}{6 \nlepochs \lipschitz} \leq \sqrt{\frac{1}{32 \lipschitz^2 \nlepochs^2}}$, $\lr \leq \frac{1}{4 \nlepochs \sqrt{\hlipschitz^2 + \graddiv}} \leq \sqrt{\frac{1}{12 \nlepochs^2 (\hlipschitz^2 + \graddiv)}}$ and $\lr \leq \frac{1}{26 \nlepochs \lipschitz}$ for all lemmas to hold, and hence $\lr \leq \min\left\{\frac{1}{26 \nlepochs \lipschitz}, \frac{1}{4 \nlepochs \sqrt{\hlipschitz^2 + \graddiv}}\right\}$, we have
\begin{align*}
&\lr/(2\nlepochs) \left(\consto \constm + \constp \right) \\
&= \frac{\lr}{2\nlepochs} \left(4\lr^2 \nlepochs^3 \frac{\glipub}{\nseeds} + 12\lr^2 \nlepochs^3 \lipschitz\right) \left(20 \cdot 16 \nlepochs^4 \lipschitz^2 \lr^2 \hlipschitz^2 + 4 \nlepochs \frac{(1+\epsilon)}{(1-\epsilon)} \robustness 6(\hlipschitz^2+\graddiv) + 4\nlepochs \hlipschitz^2\right) \\
&+ 4\nlepochs \frac{\glipub}{\nseeds} \frac{\lr}{2\nlepochs} \left(5 \nlepochs^2 \lipschitz^2 \lr^2 \frac{1}{\vert \honest \vert} + 2 \nlepochs \frac{(1+\epsilon)}{(1-\epsilon)} \robustness + \frac{8}{\vert \honest \vert^2} \right) + 4\nlepochs^2 \lipschitz \frac{\lr}{2\nlepochs} \left(5\lr^2 16 \nlepochs^2 \lipschitz^2 \pscale + 6 \robustness \frac{(1+\epsilon)}{(1-\epsilon)} + 2\pscale\right) \\
&= 2\lr^3 \nlepochs^2 \left(\frac{\glipub}{\nseeds} + 3 \lipschitz\right) \left(20 \cdot 16 \nlepochs^4 \lipschitz^2 \lr^2 \hlipschitz^2 + 4 \nlepochs \frac{(1+\epsilon)}{(1-\epsilon)} \robustness 6(\hlipschitz^2+\graddiv) + 4\nlepochs \hlipschitz^2\right) \\
&+ 2 \frac{\glipub}{\nseeds} \lr \left(5 \nlepochs^2 \lipschitz^2 \lr^2 \frac{1}{\vert \honest \vert} + 2 \nlepochs \frac{(1+\epsilon)}{(1-\epsilon)} \robustness + \frac{8}{\vert \honest \vert^2} \right) + 4\nlepochs^2 \lipschitz \frac{\lr}{2\nlepochs} \left(5\lr^2 16 \nlepochs^2 \lipschitz^2 \pscale + 6 \robustness \frac{(1+\epsilon)}{(1-\epsilon)} + 2\pscale\right) \\
&= 2 \lr \nlepochs \left(\frac{\glipub}{\lipschitz^2 \nseeds} + 3 \frac{1}{\lipschitz}\right) \left(13 \nlepochs \hlipschitz^2 + 4 \frac{(1+\epsilon)}{(1-\epsilon)} \robustness 6(\hlipschitz^2+\graddiv) + 4 \hlipschitz^2\right) \\
&+ 2 \lr \frac{\glipub}{\nseeds} \left(\frac{1}{5\vert \honest \vert} + 2 \nlepochs \frac{(1+\epsilon)}{(1-\epsilon)} \robustness + \frac{8}{\vert \honest \vert^2} \right) + 2\nlepochs \lipschitz \lr \left(6 \robustness \frac{(1+\epsilon)}{(1-\epsilon)} + 6\pscale\right). \\
\end{align*}
We let $\probguarantee \define \delta \nlepochs \ngepochs$, and obtain $\epsilon \geq \sqrt{\frac{64}{\nseeds} \log(\frac{2(\vert \honest \vert-1)}{ \delta})} = \sqrt{\frac{64}{\nseeds} \log(\frac{2(\vert \honest \vert-1) \ngepochs \nlepochs}{\probguarantee})}$. Since it is required to satisfy $\epsilon < 1$, the proof holds for $\nseeds \geq 64 \log(\frac{2(\vert \honest \vert-1) \ngepochs \nlepochs}{\probguarantee})$. 
This concludes the proof.

\end{proof}

\begin{proof}[Proof of \cref{lemma:zovar_lipschitz}]
For the proof of the zero-order approximated gradient variance, we rely on the following intermediate lemma.
\begin{lemma}[Lemma 5.3, \citep{tang2020distributed}] \label{lem:est_var_lipschitz} Let $\loss$ be $\glipschitz$-Lipschitz. Then for any $\model \in \mathbb{R}^\dimension$, $\perturbvec$ and $\pscale > 0$, we have
\begin{align*}
    \ec{\mathbf{z}}{\sqnorm{\perturbvec^T \projtmp}} \leq \glipub
\end{align*}
\end{lemma}
We bound the zero-order approximated gradient variance as follows:
Since $\e{\sqnorm{Z-\e{Z}}} \leq \e{\sqnorm{Z}}$, and $\e{\perturbtlr \projitlr} = \nlossiu(\modeltl)$, we have
\begin{align*}
    &\e{\sqnorm{\perturbtl \projitl - \nlossiu(\modeltl)}} = \e{\sqnorm{\frac{1}{\nseeds} \sum_{\seed = 1}^\nseeds \perturbtlr \projitlr - \nlossiu(\modeltl)}} \\
    &\overset{(a)}{=} \frac{1}{\nseeds^2} \sum_{\seed = 1}^\nseeds \e{\sqnorm{\perturbtlr \projitlr - \nlossiu(\modeltl)}} \\
    &\leq \frac{1}{\nseeds^2} \sum_{\seed = 1}^\nseeds \e{\sqnorm{\perturbtlr \projitlr}} \overset{(b)}{\leq} \frac{\glipub}{\nseeds}
\end{align*}
where $(a)$ is due to the independence of $\perturbtlr \projitlr$ and $\perturbtlr[\seed^\prime] \projitlr[\seed^\prime]$ for $\seed \neq \seed^\prime$. $(b)$ is by \cref{lem:est_var_lipschitz}.
\end{proof}

\begin{proof}[Proof of \cref{lemma:model_progress_lipschitz}]
By definition, $\ec{\histl[\lepoch-1]}{\sqnorm{\modeltmavg[1]-\modelt}}=0$. From \eqref{eq:model_progress_lipschitz}, we have for $\lepoch \in \{2, \cdots, \nlepochs\}$, 
\begin{align*}
    &\ec{\histl[\lepoch-1]}{\sqnorm{\modeltlavg-\modelt}} \\
    &\leq (1+\frac{1}{\tradetmp}) \ec{\histl[\lepoch-1]}{\sqnorm{\modeltmavg[\lepoch-1] - \modelt}} + 8(1+\tradetmp) \eta^2 \ec{\histl[\lepoch-1]}{\sqnorm{\frac{1}{\vert \honest \vert} \sumhc \nlossiu(\modeltll[\lepoch-1]) - \frac{1}{\vert \honest \vert} \sumhc \nlossi(\modeltll[\lepoch-1])}}
    \\
    &+ 8(1+\tradetmp) \eta^2 \ec{\histl[\lepoch-1]}{\sqnorm{\nlossg(\modelt)}} \\
    &+ 8(1+\tradetmp) \eta^2 \ec{\histl[\lepoch-1]}{\sqnorm{\frac{1}{\vert \honest \vert} \sumhc \nlossi(\modeltll[\lepoch-1]) - \nlossg(\modeltmavg[\lepoch-1])}}
    + 8(1+\tradetmp) \eta^2 \ec{\histl[\lepoch-1]}{\sqnorm{\nlossg(\modeltmavg[\lepoch-1]) - \nlossg(\modelt)}}
    \\
    &+ 2\eta^2 \frac{1}{\vert \honest \vert^2} \sumhc \ec{\histl[\lepoch-1]}{\sqnorm{\perturbtl[\lepoch-1] \projitll[\lepoch-1] - \nlossiu(\modeltll[\lepoch-1])}} \\
    &\overset{(a)}{\leq} (1+\frac{1}{\tradetmp}) \ec{\histl[\lepoch-1]}{\sqnorm{\modeltmavg[\lepoch-1] - \modelt}} + 8(1+\tradetmp) \eta^2 \approxerr 
    + 8(1+\tradetmp) \eta^2 \ec{\histl[\lepoch-1]}{\sqnorm{\nlossg(\modelt)}} \\
    &+ 8(1+\tradetmp) \eta^2 \frac{\hlipschitz^2}{\vert \honest \vert} \sumhc \ec{\histl[\lepoch-1]}{\sqnorm{\modeltll[\lepoch-1] - \modeltmavg[\lepoch-1]}}
    + 8(1+\tradetmp) \eta^2 \lipschitz^2 \ec{\histl[\lepoch-1]}{\sqnorm{\modeltmavg[\lepoch-1] - \modelt}}
    \\
    &+ 2\eta^2 \frac{1}{\vert \honest \vert^2} \sumhc \frac{\glipub}{\nseeds} \\
    &= \left((1+\frac{1}{\tradetmp}) + 8(1+\tradetmp) \eta^2 \lipschitz^2 \right) \ec{\histl[\lepoch-1]}{\sqnorm{\modeltmavg[\lepoch-1] - \modelt}} + 8(1+\tradetmp) \eta^2 \approxerr 
    \\
    &+ \left(8(1+\tradetmp) \eta^2\right) \ec{\histl[\lepoch-1]}{\sqnorm{\nlossg(\modelt)}} \\
    &+ \left(8(1+\tradetmp) \eta^2 \frac{\hlipschitz^2}{\vert \honest \vert} \right)  \sumhc \ec{\histl[\lepoch-1]}{\sqnorm{\modeltll[\lepoch-1] - \modeltmavg[\lepoch-1]}} + 2\eta^2 \frac{1}{\vert \honest \vert} \frac{\glipub}{\nseeds},
\end{align*}
where $(a)$ follows from \cref{ass:lipschitz}, \cref{ass:lipschitz_avg_grad} and an intermediate step in the proof of \cref{lemma:zovar}.

To ensure the bound holds uniformly for all $\lepoch \in [\nlepochs]$, we now choose $\tradetmp=2\nlepochs-1$ and the learning rate small enough so that $\left((1+\frac{1}{\tradetmp}) + 8(1+\tradetmp) \eta^2 \lipschitz^2 \right) \leq 1 + \frac{1}{\nlepochs-1}$, i.e., that $8(1+\tradetmp) \eta^2 \lipschitz^2 \leq \frac{1}{2\nlepochs}$. Hence, the learning rate is required to satisfy $\lr \leq \sqrt{\frac{1}{32 \lipschitz^2 \nlepochs^2}}$
With this choice of the learning rate, we have
\begin{align*}
    &\ec{\histl[\lepoch-1]}{\sqnorm{\modeltlavg-\modelt}} \\
    &\leq (1+\frac{1}{\nlepochs-1}) \ec{\histl[\lepoch-1]}{\sqnorm{\modeltmavg[\lepoch-1] - \modelt}} + 8(1+\tradetmp) \eta^2 \approxerr 
    + \left(8(1+\tradetmp) \eta^2\right) \ec{\histl[\lepoch-1]}{\sqnorm{\nlossg(\modelt)}} \\
    &+ \left(8(1+\tradetmp) \eta^2 \frac{\hlipschitz^2}{\vert \honest \vert} \right)  \sumhc \ec{\histl[\lepoch-1]}{\sqnorm{\modeltll[\lepoch-1] - \modeltmavg[\lepoch-1]}} + 2\eta^2 \frac{1}{\vert \honest \vert} \frac{\glipub}{\nseeds} \\
    &\overset{(d)}{\leq} \consteprime \ec{\histl[\lepoch-1]}{\sqnorm{\nlossg(\modelt)}} + \constfprime + \constgprime \sum_{\lepoch^\prime=1}^{\lepoch-1} \sumhc \ec{\histl[\lepoch^\prime-1]}{\sqnorm{\modeltll[\lepoch^\prime] - \modeltmavg[\lepoch^\prime]}},
\end{align*}
where $(e)$ is by the recursive application of $(d)$ and the fact that $(1+\frac{1}{\nlepochs})^\lepoch \leq (1+\frac{1}{\lepoch})^\lepoch \leq e \leq 5$ for all $\lepoch \in [\nlepochs]$. 
This concludes the proof. The constants are given as
\begin{align*}
    \consteprime &\define 5(\lepoch-1)  \left(8(1+\tradetmp) \eta^2\right) \leq \conste \define 5\cdot 16 \lr^2 \nlepochs^2 \\
    \constfprime &\define 5(\lepoch-1) \left(2\eta^2 \frac{1}{\vert \honest \vert} \frac{\glipub}{\nseeds} + 8(1+\tradetmp) \eta^2 \approxerr\right) \\
    &\leq \constf \define 5\nlepochs \lr^2 \frac{\glipub}{\vert \honest \vert\nseeds} + 5 \lr^2 16\nlepochs^2 \approxerr \\
    \constgprime &\define 5\left(8(1+\tradetmp) \eta^2 \hlipschitz^2\right) \leq \constg \define 5 \cdot 16\lr^2 \left(\nlepochs \hlipschitz^2\right)
\end{align*}
\end{proof}

\begin{proof}[Proof of \cref{lemma:graddiv_lipschitz}]

Using \eqref{eq:graddiv_lipschitz} , we obtain
\begin{align*}
    &\sumletmp \frac{1}{\vert \honest \vert} \sumhc \e{\sqnorm{\perturbtm \projitl - \frac{1}{\vert \honest \vert} \sumhcj \perturbtm \projjtm}} \\
    &\overset{(a)}{\leq} 2 \sum_{\lepochtmp=1}^{\lepoch} \e{\left(3 \frac{\hlipschitz^2}{\vert \honest \vert} + 3\frac{\graddiv}{\vert \honest \vert}\right) \sumhc \sqnorm{\modeltmavg[\lepochtmp] - \modeltm} + 3\lipschitz} \\
    &+ 2 \sumletmp \frac{1}{\vert \honest \vert} \sumhc \bigg(\frac{\glipub}{\nseeds}\bigg) \\
    &\leq \sum_{\lepochtmp=1}^{\lepoch} \frac{1}{\vert \honest \vert} \sumhc \consta \e{ \sqnorm{\modeltmavg[\lepochtmp] - \modeltm}} + \sum_{\lepochtmp=1}^{\lepoch} (\constb + \constc)
\end{align*}
where $(a)$ is due to \cref{lemma:zovar_lipschitz}. Thereby,
\begin{align*}
    \consta[\lepoch] &\define 6 \hlipschitz^2 + 6\graddiv \\
    \constb &\define 2 \frac{\glipub}{\nseeds} \\
    \constc &\define 6 \lipschitz
\end{align*}
\end{proof}

\begin{proof}[Proof of \cref{lemma:model_divergence_lipschitz}]

From \cref{lemma:graddiv_lipschitz}, we have
\begin{align*}
    &\sumle \frac{1}{\vert \honest \vert} \sumhc \e{\sqnorm{\modeltl-\modeltlavg}} = \sumle \frac{1}{\vert \honest \vert} \sumhc \e{\sqnorm{\lr \sum_{\lepochtmp=1}^{\lepoch} \left(\perturbtm \projitl - \frac{1}{\vert \honest \vert} \sumhcj \perturbtm \projjtm \right)}} \\
    &\leq \lr^2 \sumle \lepoch \sum_{\lepochtmp=1}^{\lepoch} \frac{1}{\vert \honest \vert} \sumhc \e{\sqnorm{\perturbtm \projitl - \frac{1}{\vert \honest \vert} \sumhcj \perturbtm \projjtm}} \\
    &= \lr^2 \sumle \lepoch \sum_{\lepochtmp=1}^{\lepoch} \sumhc \frac{1}{\vert \honest \vert}  \consta \e{ \sqnorm{\modeltmavg[\lepochtmp] - \modeltm}} + \lr^2 \sumle \lepoch \sum_{\lepochtmp=1}^\lepoch \constb + \lr^2 \sumle \lepoch \sumletmp \constc \\
    &\leq \lr^2 \sumle \frac{1}{\vert \honest \vert} \sumhc \nlepochs^2 \consta[\nlepochs] \e{ \sqnorm{\modeltmavg[\lepoch] - \modeltl}} + \lr^2put \sumle \lepoch \sum_{\lepochtmp=1}^\lepoch \constb + \lr^2 \sumle \lepoch \sumletmp \constc 
\end{align*}
We rewrite the expression as
\begin{align*}
\left(1-\lr^2 \nlepochs^2 \consta[\nlepochs]\right) \sumle \frac{1}{\vert \honest \vert} \sumhc \e{\sqnorm{\modeltl-\modeltlavg}} \leq \lr^2 \sumle \lepoch \sum_{\lepochtmp=1}^\lepoch \constb + \lr^2 \sumle \lepoch \sumletmp \constc
\end{align*}
and choose the learning rate small enough so that $\left(1-\lr^2 \nlepochs^2 \consta[\nlepochs]\right) = \left(1-6\lr^2 \nlepochs^2 (\hlipschitz^2 + \graddiv) \right) \geq \frac{1}{2}$. Hence, we obtain
\begin{align*}
\sumle \frac{1}{\vert \honest \vert} \sumhc \e{\sqnorm{\modeltl-\modeltlavg}} \leq \constmprime,
\end{align*}
where
\begin{align*}
    \constmprime &\define 2\lr^2 \sumle \lepoch \sum_{\lepochtmp=1}^\lepoch \constb + 2\lr^2 \sumle \lepoch \sumletmp \constc \leq \constm \define 4\lr^2 \nlepochs^3 \frac{\glipub}{\nseeds} + 12\lr^2 \nlepochs^3 \lipschitz. %
\end{align*}
The learning rate must satisfy $\lr \leq \sqrt{\frac{1}{12 \nlepochs^2 (\hlipschitz^2 + \graddiv)}}$. This concludes the proof.
\end{proof}

\end{document}